\pdfoutput=1
\documentclass[11pt]{article}

\usepackage[]{acl}

\usepackage{times}
\usepackage{latexsym}

\usepackage[T1]{fontenc}
\usepackage[utf8]{inputenc}

\usepackage{microtype}

\usepackage{inconsolata}

\usepackage{graphicx}

\usepackage{hyperref}       % hyperlinks
\usepackage{url}            % simple URL typesetting
\usepackage{booktabs}       % professional-quality tables
\usepackage{amsfonts}       % blackboard math symbols
\usepackage{nicefrac}       % compact symbols for 1/2, etc.
\usepackage{microtype}      % microtypography
\usepackage{xcolor}         % colors
\usepackage{colortbl}

\usepackage{subcaption}
\usepackage{amsmath}
\usepackage{verbatim}
\usepackage{algorithm}
\usepackage{algpseudocode}
\usepackage{amsmath}
\usepackage{amssymb}
\usepackage{multirow}
\usepackage{makecell}
\usepackage{booktabs}
\usepackage{stfloats}
\usepackage[misc]{ifsym}
\usepackage{CJKutf8}
\usepackage{color}
\usepackage{bbding}
\usepackage{wrapfig}
\usepackage{enumitem}
\usepackage{CJKutf8}
\usepackage{tcolorbox}
\tcbuselibrary{breakable}

\definecolor{c1}{RGB}{233,220,185}
\definecolor{c2}{RGB}{197,217,192}
\definecolor{c3}{RGB}{201,213,239}
\definecolor{c4}{RGB}{234,180,179}

\newcommand*\samethanks[1][\value{footnote}]{\footnotemark[#1]}

\title{\textsc{SocialEval}: Evaluating Social Intelligence of Large Language Models}

\author{
    Jinfeng Zhou\textsuperscript{\rm 1}\thanks{Equal contribution.} \quad
    Yuxuan Chen\textsuperscript{\rm 1}\samethanks{} \quad
    Yihan Shi\textsuperscript{\rm 2} \quad
    Xuanming Zhang\textsuperscript{\rm 3} \quad
    Leqi Lei\textsuperscript{\rm 1} \\
    \textbf{Yi Feng\textsuperscript{\rm 4} \quad
    Zexuan Xiong\textsuperscript{\rm 1} \quad
    Miao Yan\textsuperscript{\rm 5} \quad
    Xunzhi Wang\textsuperscript{\rm 6} \quad
    Yaru Cao\textsuperscript{\rm 7} \quad
    Jianing Yin\textsuperscript{\rm 8}} \\
    \textbf{Shuai Wang\textsuperscript{\rm 9} \quad
    Quanyu Dai\textsuperscript{\rm 9} \quad
    Zhenhua Dong\textsuperscript{\rm 9} \quad
    Hongning Wang\textsuperscript{\rm 1} \quad
    Minlie Huang\textsuperscript{\rm 1}} \\
    \textsuperscript{\rm 1}The CoAI Group, DCST, Tsinghua University \quad 
    \textsuperscript{\rm 2}Harvard University \\
    \textsuperscript{\rm 3}University of Wisconsin–Madison \quad
    \textsuperscript{\rm 4}Beijing Jiaotong University \\
    \textsuperscript{\rm 5}Peking University \quad
    \textsuperscript{\rm 6}Nankai University \quad
    \textsuperscript{\rm 7}Northwest Minzu University \\
    \textsuperscript{\rm 8}University of Pennsylvania \quad
    \textsuperscript{\rm 9}Huawei Noah' Ark Lab \\
    \texttt{zjf23@mails.tsinghua.edu.cn \quad \{hw-ai,aihuang\}@tsinghua.edu.cn} \\
}

\begin{document}
\maketitle
\begin{abstract}

LLMs exhibit promising Social Intelligence (SI) in modeling human behavior, raising the need to evaluate LLMs' SI and their discrepancy with humans. SI equips humans with interpersonal abilities to behave wisely in navigating social interactions to achieve social goals. This presents an operational evaluation paradigm: outcome-oriented goal achievement evaluation and process-oriented interpersonal ability evaluation, which existing work fails to address. To this end, we propose \textsc{SocialEval}, a script-based bilingual SI benchmark, integrating outcome- and process-oriented evaluation by manually crafting narrative scripts. Each script is structured as a world tree that contains plot lines driven by interpersonal ability, providing a comprehensive view of how LLMs navigate social interactions. Experiments show that LLMs fall behind humans on both SI evaluations, exhibit prosociality, and prefer more positive social behaviors, even if they lead to goal failure. Analysis of LLMs' formed representation space and neuronal activations reveals that LLMs have developed ability-specific functional partitions akin to the human brain.\footnote{Data: \url{https://github.com/thu-coai/SocialEval}}

\end{abstract}

\section{Introduction}

Social Intelligence (SI) involves understanding and managing human behaviors to act wisely in social interactions \cite{si_definition}. It is essential for maintaining interpersonal relationships by building trust \cite{emotional_intelligence}, resolving conflicts \cite{emotional_conflict}, promoting collaboration \cite{collective_intelligence}, thus effectively navigating complex social dynamics \cite{carnegie2024win}. 
Driven by the fast development of large language models (LLMs, \citealt{llama,llama2}), more and more research has reported the emergence of SI in LLMs through society simulations \cite{ai_town}, where social dynamics among interacting LLMs are observed and show parallels to human behaviors \cite{simulate_human_trust_behavior}. 
This sparks the community's strong interest in using LLMs to study social science \cite{llm_prosocial_irrationality}, train people to handle interpersonal situations \cite{lin2024imbue}, and many more. 
%As LLMs become active participants in human social ecosystems, 
However, a fundamental question whose answer measures the validity of existing research in this line has not received enough attention: \textit{\textbf{how can we thoroughly evaluate an LLM's SI and recognize its discrepancy from humans'?}}

\begin{figure}[t]
    \centering
    \includegraphics[width=\columnwidth]{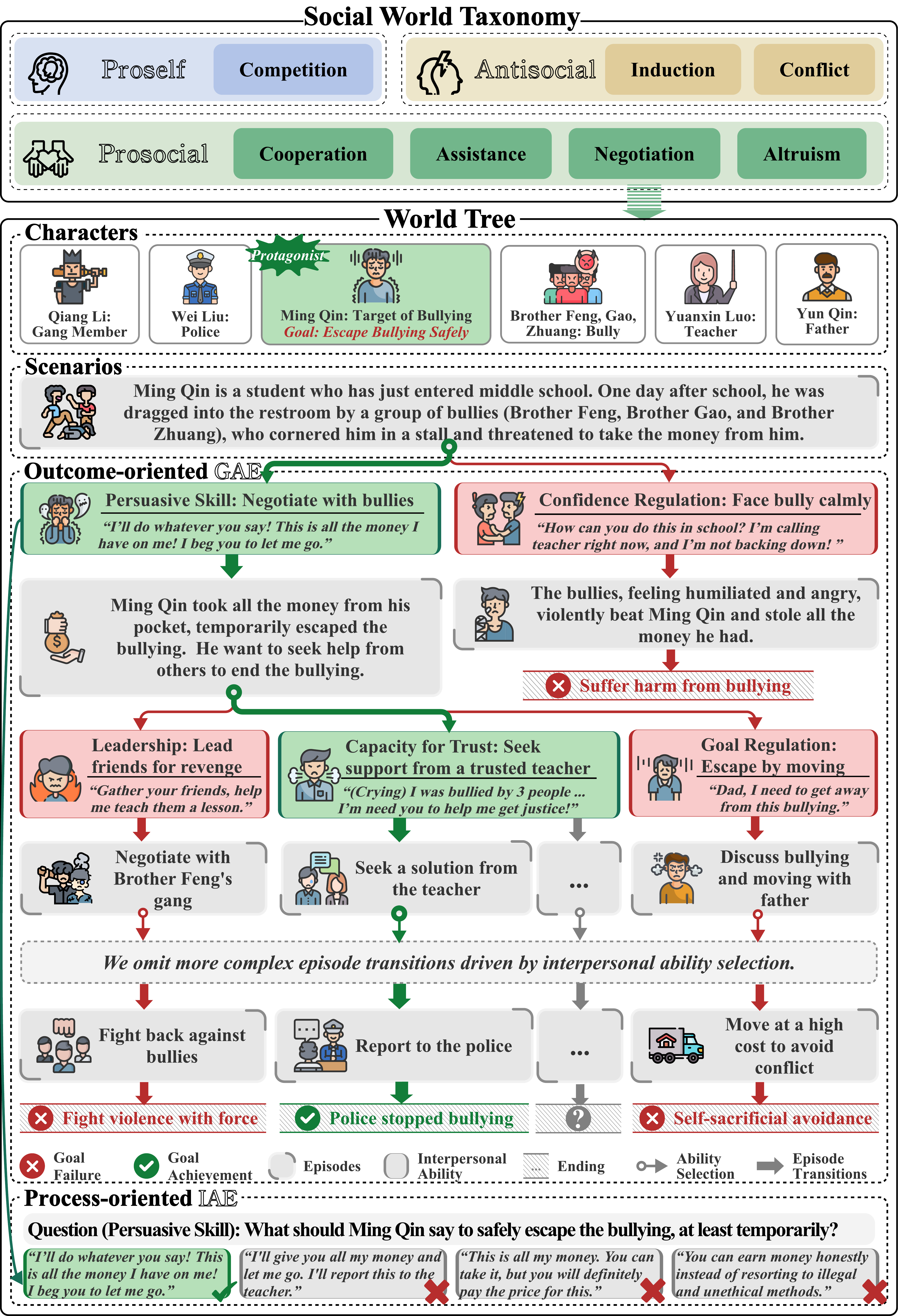}
    \caption{\textsc{SocialEval} framework and a case of world tree-based narrative script performing outcome-oriented goal achievement evaluation and process-oriented interpersonal ability evaluation ($\mathbb{GAE}$ and $\mathbb{IAE}$).}
    \label{socialeval_framework}
    \vspace{-3mm}
\end{figure}

Social psychology theories suggest that humans' SI manifests in a dynamic process, which is a continuous adaptation of interpersonal skills within evolving narrative-like social activities that can be naturally organized as ``scripts'' \cite{script_theory}. 
As shown in Figure \ref{socialeval_framework}, the protagonist's initial confrontation (\textit{Episode 1}) informs his subsequent decision to seek a solution from his teacher (\textit{Episode 2}), which then shapes his institutional response (\textit{Episode 3}). Each decision transits current episode to the next, forming an interdependent sequence that moves toward distinct social goals (e.g., \textit{escaping bullying}) while realized via interpersonal skills (e.g., \textit{persuasive skill}) \cite{social_cognitive_theory,dramaturgical_theory}. This process naturally presents an operational evaluation paradigm for both humans and machines: 1) \textit{\textbf{Outcome-oriented evaluation} of goal achievement within social activities} and 2) \textit{\textbf{Process-oriented evaluation} of interpersonal abilities towards goal pursuit}. 

Existing related works \cite{sotopia,stss,agentsense} still fail to meet this paradigm due to two serious issues, hindering a holistic SI evaluation: 1) Limited on single-episode social dynamics, ignoring how SI is reflected in sequentially dependent episodes; 2) Focus only on terminal goal achievement, lacking fine-grained interactive process design for evaluating goal pursuit.

Drawing on social psychology theories \cite{script_theory,dramaturgical_theory}, we propose \textsc{SocialEval}, a script-based bilingual benchmark designed to evaluate and inspect LLMs' SI. It integrates process- and outcome-oriented evaluations by manually crafted narrative scripts, providing a comprehensive view of how LLMs navigate through social interactions. 
As shown in Figure \ref{socialeval_framework}, each script constitutes a social world, presented by multiple plot lines with interconnected episode sequences. At each critical juncture in a plot's progression, we craft options exhibiting interpersonal abilities that lead to distinct episode transitions, and thus subsequent diverging endings. The plot lines intertwine via episode transitions to form a tree-like social world, so we call it the ``world tree''.

With \textsc{SocialEval}, we thoroughly evaluate and inspect LLMs’ SI via:
\textbf{1) Comprehensive social world constructions}. Drawing on interdependence theory \cite{interdependence_theory}, we compile a rich set of prototypical narrative-like social activities to define our social world taxonomy (\S \ref{sec:social_world_taxonomy}), which includes 3 categories across 7 sub-categories (Figure \ref{socialeval_framework}).
\textbf{2) Outcome-oriented evaluation}. We manually craft 153 world trees according to the taxonomy (\S \ref{sec:socialeval_collection}), each featuring plot lines marked by goal achievement indicators, presented via dialogues among a protagonist and other characters. The tested LLM plays the protagonist and selects options, which are presented as candidate utterances and exhibit interpersonal abilities, to transit the episode towards achieving social goals (\S \ref{sec:gae_and_iae_results}).
This explicit navigation process supports investigating possible behavioral discrepancies between LLMs and humans (\S \ref{sec:behavior_level_analysis}).
\textbf{3) Process-oriented evaluation}. To test whether LLM can correctly use the abilities in the options for goal pursuit , we manually craft probing questions and plausible yet misleading options reflecting wrong abilities, forming multiple-choice questions (\S \ref{sec:gae_and_iae_results}).
Notably, this setup support delves deeper into LLMs' representation spaces and neuronal activations, examining whether LLMs' interpersonal abilities are stored in specific functional groups, akin to the human brain (\S \ref{sec:representation_neron_level_analysis}).
Each option forms a test sample. 153 world trees produce 2,493 test samples, covering 5 interpersonal ability aspects and 32 specific abilities.

To the best of our knowledge, \textsc{SocialEval} is the first script-based bilingual SI benchmark. Upon it, we perform outcome-oriented goal achievement evaluation and process-oriented interpersonal ability evaluation of both LLMs and humans. Results reveal that LLMs fall behind humans in both SI evaluations, while both of them exhibit cross-linguistic differences in SI. We find that existing LLMs show strong preferences over prosocial and positive behaviors, even when such behaviors may ultimately lead to goal failure. By inspecting LLMs' formed representations and neuronal activations regarding interpersonal abilities, we unveil that with growing model size, LLMs gradually develop ability-specific functional partitions, similar to those found in the human brain \cite{feature_lobes}.

\section{Preliminaries}

\subsection{Social World Taxonomy}
\label{sec:social_world_taxonomy}

SI enables individuals to behave wisely in social interactions by considering how their own and others’ behaviors would influence immediate and future outcomes for all involved. The outcomes are often reflected in individual social goals that shape one’s interpersonal orientations, essentially the ways of interacting with others, and ultimately guide social behaviors \cite{interdependence_theory}. Along this theory, the orientations are divided by outcome transformations, which reflect the flow of interests between oneself and others, classifying social worlds into \textbf{prosocial}, \textbf{proself}, and \textbf{antisocial}. Based on this interest flow to extend this theory, we refine these orientations by defining outcome as a set constructed by self-interest and altruism, i.e., \texttt{outcome = (self-interest, altruism)}:
\begin{itemize}[leftmargin=*,topsep=3pt, itemsep=0pt, partopsep=0pt]
\setlength\itemsep{-1mm}

\item \textbf{Self-interest}: the desire of pursuing personal gains, with values [1, 0, -1] being the strong pursuit of self-interest, balancing personal and others' interests, sacrificing personal interests.

\item \textbf{Altruism}: the desire of prioritizing others' interests, with values [1, 0, -1] being strong prioritization of others' interests, balancing personal and others' interests, neglecting others' interests.
\end{itemize}
Via Cartesian product, we obtain 9 orientations: \textbf{cooperation} (1,1), \textbf{negotiation} (1,0), \textbf{competition} (1,-1), \textbf{assistance} (0,1), chitchat (0,0), \textbf{induction} (0,-1), \textbf{altruism} (-1,1), withdrawal (-1,0), \textbf{conflict} (-1,-1). As chitchat lacks clear social goals and withdrawal is rare, we exclude them, leaving 7 orientations in our taxonomy. Detailed explanations are provided in Appendix \ref{sec:social_world_taxonomy_appendix}.

\subsection{Interpersonal Ability Inventory}

Interpersonal ability enables individuals to navigate social dynamics for goal pursuit.
We adopt the BESSI (Behavioral, Emotional, and Social Skills Inventory), a psychological framework \cite{bessi} that integrates comprehensive interpersonal abilities, as our inventory for SI evaluation. The framework includes 5 aspects covering 32 specific abilities: 
\textbf{(1) Social Engagement} (5 abilities): actively engaging with others.
\textbf{(2) Cooperation} (5 abilities): maintaining positive social relationships.
\textbf{(3) Self-Management} (12 abilities): effectively pursuing social goals and completing social tasks.
\textbf{(4) Emotional Resilience} (5 abilities): regulating emotions and moods.
\textbf{(5) Innovation} (5 abilities): engaging with novel ideas and experiences.
Details on the 32 abilities are provided in Appendix \ref{sec:interpersonal_ability_details}.

\subsection{Evaluation Task Formulation}

We examine LLMs' SI by investigating their navigation in the world trees to achieve desired social goals.
Each world tree can be formalized as a goal-conditioned Markov Decision Process \cite{mdp} with the 4-tuple $(\mathcal{S},\mathcal{A},\mathcal{T},\mathcal{R})$, where $\mathcal{S}$ is the state space, $\mathcal{A}$ is the action space, $\mathcal{T}: \mathcal{S} \times \mathcal{A} \rightarrow \mathcal{S}$ is the transition function, and $\mathcal{R}:\mathcal{S} \times \mathcal{A} \rightarrow \mathbb{R} \in\{0,1\}$ is the reward function.

\paragraph{Task 1: Outcome-oriented Goal Achievement Evaluation}

This task is defined as implementing a function $F_{\mathcal{T},\mathcal{R}}:\mathcal{S} \times \mathcal{A} \rightarrow \mathcal{S} \times \mathbb{R}$, which maps a given state and action (i.e., $s_{t}$, $a_{t}$) to the subsequent state and a goal achievement reward (i.e., $s_{t+1}$, $r_{t+1}$). 
Here, the state $s_{t}$ is the current episode $\mathcal{E}_{t}$ in which the player is located in the world tree, i.e., $s_{t}=\mathcal{E}_{t}$. 
The action $a_{t}$ is the protagonist's next utterance $u_{t}$ from action space $\mathcal{A}_{t}$ that links the subsequent episode $\mathcal{E}_{t+1}$ and reflects specific interpersonal abilities for the episode transition, i.e., $a_{t} \in \mathcal{A}_{t}=\{u_{t,1},…,u_{t,m}\}$. 
Thus, we define interpersonal ability-driven episode transition as:
\begin{equation}
s_{t+1}=\mathcal{E}_{t+1}=F_{\mathcal{T}}(s_{t},a_{t}).
\end{equation}
After that, we assess LLMs' goal achievement by:
\begin{equation}
r_{t+1}=F_{\mathcal{R}}(s_{t+1}),
\end{equation}
where $r_{t+1}=1$ indicates the social goal has been achieved; otherwise, $r_{t+1}=0$.

\paragraph{Task 2: Process-oriented Interpersonal Ability Evaluation}

For each candidate utterance $u_{t,i}$ from action space $\mathcal{A}_t=\{u_{t,1},…,u_{t,m}\}$ of state $s_{t}$, $i \in [1,m]$, which showcases specific interpersonal abilities, we construct tailored questions $\mathcal{Q}_{t,i}$ to probe these abilities, along with several incorrect but misleading distractor utterances $u_{t,i}^{d}=\{u_{t,i,1}^{d},…,u_{t,i,k}^d\}$. Thus, this task is defined as a function $F_{\mathcal{R}}:\mathcal{S} \times \mathcal{Q} \times \mathcal{U}^{k+1} \rightarrow \mathcal{U} \times \mathbb{R}$ that maps from the given state $s_{t}$, question $\mathcal{Q}_{t,i}$, and set of utterance options $\mathcal{U}_{t,i}=\{u_{t,i}\} \cup u_{t,i}^{d}$ to the selected utterance $u$, we assess LLMs' interpersonal ability:
\begin{equation}
u, r=F_{\mathcal{R}}(s_{t},\mathcal{Q}_{t,i},\mathcal{U}_{t,i}),
\end{equation}
where $r=1$ denotes selected utterance $u$ correctly reflects the intended abilities, otherwise, $r=0$.

\section{\textsc{SocialEval} Collection}
\label{sec:socialeval_collection}

\subsection{World Tree Construction}
 
We hire screenwriters to manually craft world trees, and the components of a world tree are as follows.

\paragraph{Characters} involved in each world tree are divided into a protagonist and several supporting characters, with the protagonist serving as the central figure who drives plots' progression via interactions with others. Each character is set with manually crafted information: \textbf{1) Public profile} is the details known to others, e.g., identity, experiences, and social relationships. \textbf{2) Private profile} contains information known only to the character. \textbf{3) Social goals} represent the outcomes each character strives to achieve, shaping the orientation of the social world and influencing their decisions while navigating in the world tree. Social goals are also %Additionally, characters can only access each other's public profiles within the social world, while all other information remains 
invisible to other characters.

\paragraph{Scenarios} serve as the root of world trees for plot progression, providing background of the social worlds. We adopt freely created interactive videos from platforms like \href{https://www.bilibili.com/}{BiliBili} and \href{https://www.youtube.com/}{YouTube} as references for crafting scenarios tailored to social worlds with specific orientations.

\paragraph{Episodes} are composed of dialogue interactions between characters. In a world tree, multiple interconnected episodes often involve transitions in social situations, changes in social relationships, and shifts in dialogue topics, shaping plots that closely mirror real-world social dynamics.

\paragraph{Episode Transitions} occur at critical junctures in the developing plot, driving the progression of episodes. At each transition, multiple candidate utterances reflecting distinct interpersonal abilities are crafted for the protagonist to choose, each creating distinct subsequent plot lines. 32 interpersonal abilities are involved in these transitions. To evaluate them, we craft tailored questions to probe the abilities embodied in the candidate utterances. The questions follow the psychological definitions of the specific abilities and are consistent with the current episode. We also introduce distracting utterances, which are carefully crafted to be plausible yet incorrect, avoiding easy dismissal. 
These distractors, alongside the correct utterance, form the choice set for each question. Each candidate utterance produces a unique test case, comprising the current episode, question, and choice set for process-oriented interpersonal ability evaluation.

\paragraph{Plot Ending} is the final outcome of a plot line. For each plot line in a world tree, we manually craft a reasonable and logical ending that follows the sequence of episode transitions. These plot endings are annotated to indicate whether the protagonist successfully achieves their social goals, used for outcome-oriented goal achievement evaluation.

\begin{table*}[t]
\centering
\resizebox{\textwidth}{!}{
\begin{tabular}{c|c|c|c|c|c|c|c|c|c}
\toprule
    \multicolumn{2}{c|}{\makecell[c]{Orientations}} & \# World Trees & \# Avg. Characters & \# Avg. Plot Lines & \# Avg. Episodes & \# Avg. Cand.Uttr. & \# Avg. Suc.End. & \# Avg. Interactions & \# Ability Samples \\
\midrule
    \multirow{5}{*}{\makecell[c]{Prosocial}}
    & Cooperation & 25 & 6.00 & 10.20 & 7.02 & 2.10 & 1.36 & 110.04 & 447 \\
    & Negotiation & 29 & 6.28 & 11.48 & 6.41 & 2.29 & 1.72 & 112.72 & 569 \\
    & Assistance  & 26 & 5.69 & 8.38 & 6.09 & 2.18 & 1.35 & 92.65 & 373 \\
    & Altruism    & 26 & 7.00 & 16.13 & 7.18 & 2.01 & 1.35 & 119.96 & 454 \\
    & \cellcolor{c1}Overall     & \cellcolor{c1}106 & \cellcolor{c1}6.25 & \cellcolor{c1}11.56 & \cellcolor{c1}6.74 & \cellcolor{c1}2.19 & \cellcolor{c1}1.45 & \cellcolor{c1}108.94 & \cellcolor{c1}1,843 \\
\midrule
    \multirow{2}{*}{\makecell[c]{Proself}} 
    & Competition & 27 & 5.59 & 8.26 & 6.01 & 2.13 & 1.41 & 94.96 & 384 \\
    & \cellcolor{c2}Overall     & \cellcolor{c2}27 & \cellcolor{c2}5.59 & \cellcolor{c2}8.26 & \cellcolor{c2}6.01 & \cellcolor{c2}2.13 & \cellcolor{c2}1.41 & \cellcolor{c2}94.96 & \cellcolor{c2}384 \\
\midrule
    \multirow{3}{*}{\makecell[c]{Antisocial}} 
    & Induction   & 10 & 3.60 & 8.00 & 5.20 & 2.20 & 1.10 & 60.50 & 134 \\
    & Conflict    & 10 & 4.00 & 8.00 & 5.40 & 2.10 & 1.30 & 77.60 & 132 \\
    & \cellcolor{c3}Overall     & \cellcolor{c3}20 & \cellcolor{c3}3.80 & \cellcolor{c3}8.00 & \cellcolor{c3}5.30 & \cellcolor{c3}2.13 & \cellcolor{c3}1.20 & \cellcolor{c3}69.05 & \cellcolor{c3}266 \\
\midrule
    \multicolumn{2}{c|}{\cellcolor{c4} Overall}     & \cellcolor{c4} 153 & \cellcolor{c4}5.81 & \cellcolor{c4}9.46 & \cellcolor{c4}6.50 & \cellcolor{c4}2.17 & \cellcolor{c4}1.41 & \cellcolor{c4}101.27 & \cellcolor{c4}2,493 \\
\bottomrule
\end{tabular}}
% \vspace{-3mm}
\caption{Statistics of our \textsc{SocialEval}. \textit{Cand.Uttr.} is the candidate utterances. \textit{Suc.End.} is the successful ending.}
\label{socialeval_data_statistic}
% \vspace{-3mm}
\end{table*}

\subsection{Quality Control of \textsc{SocialEval}}

We employ a dedicated team of quality inspectors who are trained in inspection guidelines as follows: 
\textbf{1) Orientation feasibility}: whether the character profiles and scenario settings match the intended orientation of a social world.
\textbf{2) Plot reasonableness and coherence}: whether the progression of plots driven by character dialogues, episode transitions, and plot endings are logical and coherent throughout.
\textbf{3) Annotation accuracy}: whether the interpersonal abilities annotated to the candidate utterances are accurate and that the tailored questions accurately reflect the definitions of corresponding abilities.
\textbf{4) Task difficulty}: whether the distracting utterances can be easily identified. %We construct plausible yet incorrect options to ensure difficulty.

Upon this, our quality control pipeline involves:
\begin{itemize}[leftmargin=*,topsep=3pt, itemsep=0pt, partopsep=0pt]
\setlength\itemsep{-1mm}

\item \textbf{Inspector Training}: All inspectors are required to complete a training session that includes pilot inspections of 2 world trees. We offer feedback to help them calibrate the inspection criteria.

\item \textbf{Three-Stage Inspection}: In the first stage, an annotator thoroughly reviews the world tree upon the guidelines, identifying and correcting any content that fails to meet quality standards. In the second stage, another annotator compares the revised tree with the original, repeating the inspection process and discussing any discrepancies with the first annotator. If disagreements persist, the process moves to the third stage, where a third annotator joins the discussion to resolve differences and decide on the final modifications.

\item \textbf{Cross-Inspection and Full Check}: Each world tree is initially reviewed by the first inspector and then re-examined by a different inspector. Similarly, the trees passing examinations by two inspectors are finally reviewed by a third annotator. The assignment of inspectors to the world trees is fully random, and the final average agreement rate on all world trees is 95\%.

\end{itemize}

\subsection{Translation \& Statistics of \textsc{SocialEval}}

\paragraph{Translation}

The world trees we collected were initially crafted in Chinese. We used GPT-4o to translate them into English. To ensure the translations remain faithful to the original content, we employ professionals specializing in multi-lingual translation to review them. They are also required to evaluate the translated content with respect to the inspection guidelines. %Each orientation is checked five world trees, and 
The average acceptance rate of the translations reaches 97\%. The translation prompt is provided in Appendix \ref{sec:translation_prompt_appendix}.

\paragraph{Statistics}

\begin{figure}[t]
    \centering
    \includegraphics[width=\columnwidth]{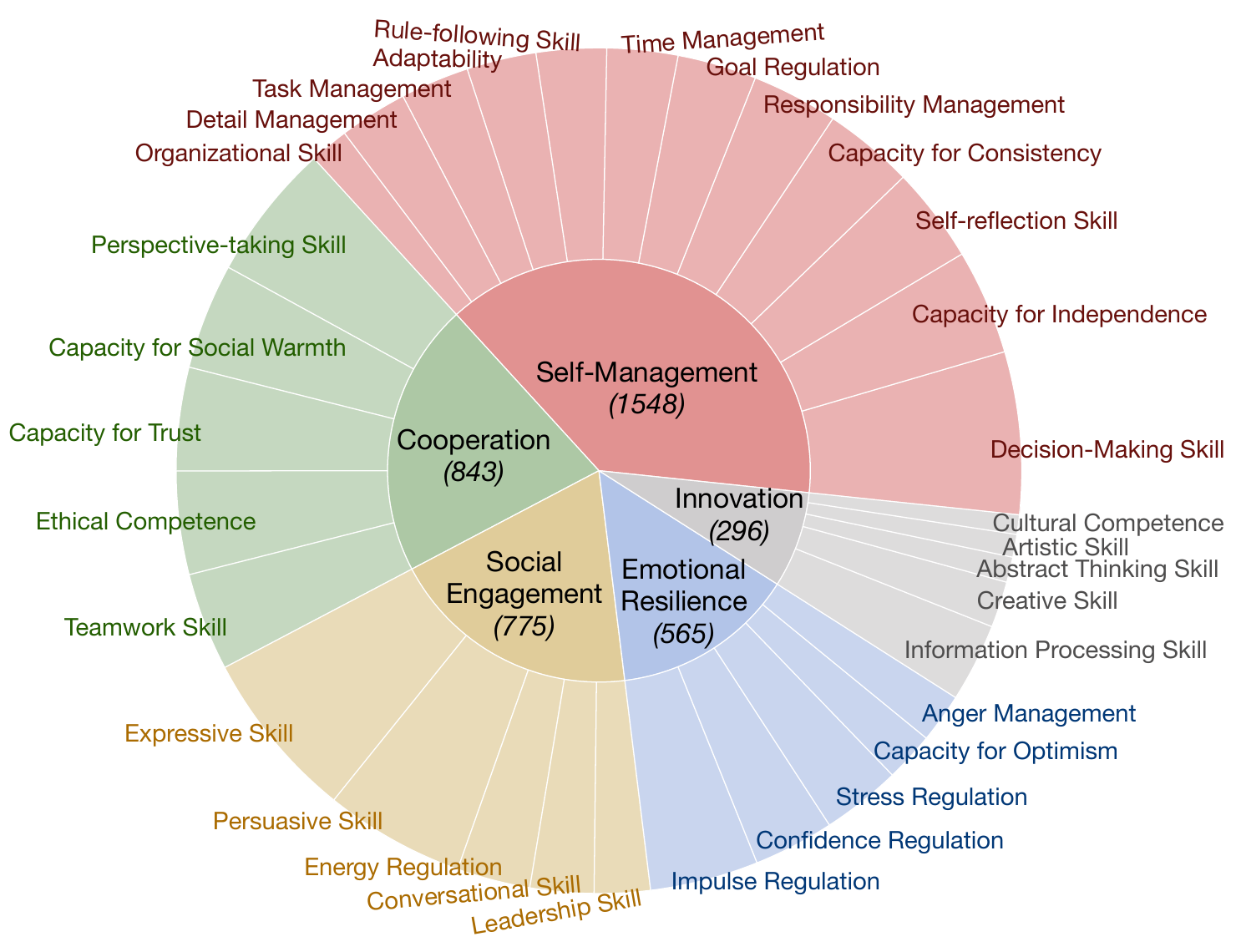}
    \caption{Distributions of interpersonal abilities in \textsc{SocialEval}, with 5 aspects and 32 specific abilities.}
    \label{interpersonal_ability_statistic}
    % \vspace{-3mm}
\end{figure}

The detailed statistics of \textsc{SocialEval} is reported in Table \ref{socialeval_data_statistic}. We crafted 153 world trees distributed in 7 orientations of social worlds. The antisocial world, Induction and Conflict, each contains only 10 trees. This is because these worlds, which harm others' interests, lack sufficient reference scripts from public sources, thus increasing construction difficulty and cost. On average, each world tree has 101 dialogue interactions between characters and contains 6.5 episodes, resulting in 5.5 episode transitions. At each transition, there are 2.17 candidate utterances, finally producing 9.46 plot lines. These show the complex dynamics of our world trees. Based on these utterances, we crafted 2,493 samples for interpersonal ability evaluation, distributed across 5 aspects and 32 abilities (Figure \ref{interpersonal_ability_statistic}). Note that a single candidate utterance may reflect multiple abilities, thus the crafted questions are also designed to assess these abilities.

\begin{table*}[t]
\centering
\resizebox{\textwidth}{!}{
\begin{tabular}{l c c c c c c c c c c c}
\toprule
    \multirow{3}{*}{\makecell[c]{Models}} 
    & \multicolumn{5}{c}{\makecell[c]{Prosocial}}
    & \multicolumn{2}{c}{\makecell[c]{Proself}} 
    & \multicolumn{3}{c}{\makecell[c]{Antisocial}} 
    & \multirow{1}{*}{\makecell[c]{\textbf{Overall}}} \\
    \cmidrule(lr){2-6} \cmidrule(lr){7-8} \cmidrule(lr){9-11}  \cmidrule(lr){12-12}
    & \texttt{Cooperation} & \texttt{Negotiation} & \texttt{Assistant} & \texttt{Altruism} & \cellcolor{c1}\texttt{Avg.} & \texttt{Competition} & \cellcolor{c2}\texttt{Avg.} & \texttt{Induction} & \texttt{Conflict} & \cellcolor{c3}\texttt{Avg.} & \cellcolor{c4}\texttt{Avg.} \\
    & \texttt{zh}/\texttt{en} & \texttt{zh}/\texttt{en} & \texttt{zh}/\texttt{en} & \texttt{zh}/\texttt{en} & \cellcolor{c1}\texttt{zh}/\texttt{en} & \texttt{zh}/\texttt{en} & \cellcolor{c2}\texttt{zh}/\texttt{en} & \texttt{zh}/\texttt{en} & \texttt{zh}/\texttt{en} & \cellcolor{c3}\texttt{zh}/\texttt{en} & \cellcolor{c4}\texttt{zh}/\texttt{en} \\ 
\midrule
    Human (best) & 100.00/100.00 & 100.00/100.00 & 100.00/100.00 & 100.00/100.00 & \cellcolor{c1}100.00/100.00 & 100.00/100.00 & \cellcolor{c2}100.00/100.00 & 100.00/100.00 & 100.00/100.00 & \cellcolor{c3}100.00/100.00 & \cellcolor{c4}100.00/100.00 \\
    Human (average) & 60.00/60.00 & 60.00/55.00 & 70.00/55.00 & 70.00/70.00 & \cellcolor{c1}64.91/59.86 & 55.00/40.00 & \cellcolor{c2}55.00/40.00 & 65.00/60.00 & 37.50/40.00 & \cellcolor{c3}51.25/50.00 & \cellcolor{c4}61.84/55.16 \\
\midrule
    \multicolumn{12}{c}{\makecell[c]{\textit{Closed-sourced LLMs}}} \\
\midrule
    % GPT-4o-mini & 54.16/53.25 & 50.45/47.14 & 47.35/43.65 & 47.41/45.97 & \cellcolor{c1}49.82/47.44 & 24.41/21.97 & \cellcolor{c2}24.41/21.97 & 19.56/11.41 & 23.46/17.48 & \cellcolor{c3}21.41/14.29 & \cellcolor{c4}41.75/38.77 \\
    Claude-3-sonnet & 55.10/52.41 & 52.14/48.32 & 48.41/45.32 & 48.62/44.65 & \cellcolor{c1}51.06/47.65 & 25.41/24.10 & \cellcolor{c2}25.41/24.10 & 22.42/21.56 & 26.44/23.10 & \cellcolor{c3}24.32/22.29 & \cellcolor{c4}43.16/40.30 \\
    GLM-4 & 56.14/52.31 & 51.42/46.73 & 47.81/42.63 & 48.24/43.71 & \cellcolor{c1}50.87/46.30 & 28.64/24.18 & \cellcolor{c2}28.64/24.18 & 20.05/16.64 & 25.90/17.14 & \cellcolor{c3}22.82/16.88 & \cellcolor{c4}43.41/38.69 \\
    % o1-mini & 55.41/54.87 & 52.41/51.11 & 51.18/50.12 & 50.11/48.16 & \cellcolor{c1}52.25/51.03 & 27.49/24.19 & \cellcolor{c2}27.49/24.19 & 20.10/28.21 & 25.15/20.09 & \cellcolor{c3}22.49/24.36 & \cellcolor{c4}44.13/42.93 \\
    GPT-4 & 55.29/55.16 & 52.41/51.15 & 49.85/46.41 & 51.45/49.35 & \cellcolor{c1}52.23/50.49 & 29.44/27.12 & \cellcolor{c2}29.44/27.12 & 19.53/15.16 & 26.85/19.74 & \cellcolor{c3}23.19/17.33 & \cellcolor{c4}44.52/42.19 \\
    GPT-4o & 56.65/55.98 & 53.14/51.56 & 51.16/50.54 & 50.41/48.41 & \cellcolor{c1}52.81/51.58 & 27.56/25.46 & \cellcolor{c2}27.56/25.46 & 21.41/17.46 & 25.16/17.65 & \cellcolor{c3}23.19/17.55 & \cellcolor{c4}44.62/42.69 \\
    Claude-3-opus & 58.12/57.45 & 53.07/50.49 & 54.12/53.11 & 50.98/48.66 & \cellcolor{c1}54.01/52.33 & 31.49/29.75 & \cellcolor{c2}31.49/29.75 & \textbf{27.46}/\textbf{25.31} & \textbf{32.10}/\textbf{30.17} & \cellcolor{c3}\textbf{29.66}/\textbf{27.61} & \cellcolor{c4}46.96/45.23 \\
    o1 & 57.41/55.65 & \textbf{54.12}/52.19 & 53.65/52.88 & 51.98/50.10 & \cellcolor{c1}54.26/52.66 & \textbf{32.45}/\textbf{30.98} & \cellcolor{c2}\textbf{32.45}/\textbf{30.98} & 25.41/23.12 & 29.89/28.46 & \cellcolor{c3}27.53/25.65 & \cellcolor{c4}47.04/45.43 \\
\midrule
    \multicolumn{12}{c}{\makecell[c]{\textit{Open-sourced LLMs}}} \\
\midrule
    % Baichuan2-7B & 37.44/34.12 & 35.13/36.41 & 34.02/32.62 & 33.43/30.09 & \cellcolor{c1}34.99/33.39 & 19.41/15.41 & \cellcolor{c2}19.41/15.41 & 11.95/10.11 & 14.11/9.69 & \cellcolor{c3}12.97/9.91 & \cellcolor{c4}29.47/27.26 \\
    % Yi-1.5-9B & 41.78/38.89 & 37.45/32.68 & 35.47/31.06 & 32.41/29.86 & \cellcolor{c1}36.75/33.06 & 20.74/17.11 & \cellcolor{c2}20.74/17.11 & 13.21/9.75 & 14.49/12.16 & \cellcolor{c3}13.82/10.89 & \cellcolor{c4}31.04/27.45 \\
    % Baichuan2-13B & 38.41/35.14 & 34.98/32.46 & 36.74/34.11 & 38.14/34.71 & \cellcolor{c1}37.00/34.05 & 21.87/16.17 & \cellcolor{c2}21.87/16.17 & 13.47/12.18 & 16.74/12.88 & \cellcolor{c3}15.02/12.51 & \cellcolor{c4}31.56/28.18 \\
    Mistral-7B-v0.3 & 40.79/40.46 & 35.74/29.06 & 38.47/31.24 & 33.74/30.52 & \cellcolor{c1}37.11/32.64 & 22.46/17.37 & \cellcolor{c2}22.46/17.37 & 12.49/8.97 & 15.65/11.57 & \cellcolor{c3}13.99/10.2 & \cellcolor{c4}31.62/27.12 \\
    Llama-3.1-8B & 41.11/39.45 & 34.65/28.44 & 40.57/37.66 & 35.74/30.26 & \cellcolor{c1}37.89/33.74 & 21.16/18.65 & \cellcolor{c2}21.16/18.65 & 11.07/10.41 & 15.16/11.24 & \cellcolor{c3}13.01/10.80 & \cellcolor{c4}31.81/28.20 \\
    % Phi-3.5-mini-instruct & 40.56/39.44 & 37.40/34.16 & 38.46/36.41 & 35.47/34.12 & \cellcolor{c1}37.93/35.95 & 23.07/20.11 & \cellcolor{c2}23.07/20.11 & 13.85/10.74 & 14.34/12.24 & \cellcolor{c3}14.08/11.45 & \cellcolor{c4}32.31/30.07 \\
    % Yi-1.5-34B & 43.41/39.14 & 41.44/38.43 & 37.41/35.68 & 37.43/35.11 & \cellcolor{c1}39.93/37.11 & 20.46/15.41 & \cellcolor{c2}20.46/15.41 & 15.41/10.99 & 16.14/15.72 & \cellcolor{c3}15.76/13.23 & \cellcolor{c4}33.45/30.27 \\
    Qwen-2.5-7B & 45.12/42.32 & 37.40/31.70 & 41.09/38.42 & 36.71/32.51 & \cellcolor{c1}39.96/36.05 & 23.06/18.61 & \cellcolor{c2}23.06/18.61 & 14.42/9.41 & 16.55/13.32 & \cellcolor{c3}15.43/11.26 & \cellcolor{c4}33.89/29.85 \\
    % Phi-3.5-moe-instruct & 42.97/41.12 & 38.66/34.55 & 40.49/38.13 & 37.96/37.11 & \cellcolor{c1}39.95/37.61 & 24.10/21.20 & \cellcolor{c2}24.10/21.20 & 14.18/10.49 & 14.74/12.85 & \cellcolor{c3}14.45/11.61 & \cellcolor{c4}33.95/31.44 \\
    Mistral-8*7B-v0.1 & 42.16/43.11 & 41.62/40.10 & 39.46/34.41 & 37.56/34.46 & \cellcolor{c1}40.22/38.03 & 24.16/16.45 & \cellcolor{c2}24.16/16.45 & 15.46/12.32 & 16.41/12.47 & \cellcolor{c3}15.91/12.39 & \cellcolor{c4}34.33/30.99 \\
    GLM-4-9B & 44.98/43.14 & 38.41/32.10 & 43.56/37.66 & 37.46/34.29 & \cellcolor{c1}40.99/36.60 & 22.64/17.65 & \cellcolor{c2}22.64/17.65 & 15.74/9.22 & 14.65/12.14 & \cellcolor{c3}15.22/10.60 & \cellcolor{c4}34.51/29.99 \\
    % Phi-4 & 44.58/42.87 & 39.49/35.79 & 42.17/39.87 & 37.54/36.19 & \cellcolor{c1}40.87/38.56 & 23.97/22.12 & \cellcolor{c2}23.97/22.12 & 16.88/12.11 & 15.94/13.48 & \cellcolor{c3}16.43/12.76 & \cellcolor{c4}34.81/32.41 \\
    Mistral-8*22B-v0.1 & 43.74/43.13 & 45.74/42.65 & 42.41/37.41 & 39.41/34.61 & \cellcolor{c1}42.90/39.51 & 26.41/17.12 & \cellcolor{c2}26.41/17.12 & 15.41/11.41 & 16.52/12.31 & \cellcolor{c3}15.94/11.84 & \cellcolor{c4}36.60/32.07 \\
    Llama-3.1-70B & 47.08/42.32 & 41.36/39.45 & 43.57/41.67 & 38.96/37.42 & \cellcolor{c1}42.65/39.81 & 25.61/23.37 & \cellcolor{c2}25.61/23.37 & 15.07/11.56 & 19.05/13.76 & \cellcolor{c3}17.12/12.96 & \cellcolor{c4}36.94/34.15 \\
    Qwen-2.5-14B & 51.65/49.52 & 38.65/34.98 & 46.48/42.12 & 39.45/37.41 & \cellcolor{c1}43.83/40.76 & 24.31/21.06 & \cellcolor{c2}24.31/21.06 & 16.74/12.38 & 20.12/16.82 & \cellcolor{c3}18.34/14.48 & \cellcolor{c4}37.18/33.97 \\
    Llama-3.3-70B & 49.25/43.83 & 45.26/44.16 & 43.29/43.19 & 40.19/34.23 & \cellcolor{c1}44.38/41.92 & 25.62/22.58 & \cellcolor{c2}25.62/22.58 & 16.19/13.47 & 16.42/13.25 & \cellcolor{c3}17.87/14.03 & \cellcolor{c4}38.14/35.09 \\  
    Qwen-2.5-32B & 52.43/50.98 & 42.65/37.96 & 47.48/41.99 & 40.32/38.46 & \cellcolor{c1}45.57/42.14 & 26.20/21.56 & \cellcolor{c2}26.20/21.56 & 17.44/14.29 & 21.07/16.75 & \cellcolor{c3}19.16/15.46 & \cellcolor{c4}38.83/35.15 \\
    Qwen-2.5-72B & 54.96/51.26 & 45.46/41.65 & 48.24/44.63 & 41.65/40.45 & \cellcolor{c1}47.45/44.35 & 26.62/23.41 & \cellcolor{c2}26.62/23.41 & 18.40/15.12 & 22.10/17.12 & \cellcolor{c3}20.15/16.07 & \cellcolor{c4}40.34/37.10 \\
    DeepSeek-V3  & 57.08/56.43 & 53.74/52.61 & 52.87/51.32 & 51.67/49.88 & \cellcolor{c1}53.42/51.94 & 30.12/28.46 & \cellcolor{c2}30.12/28.46 & 23.17/20.88 & 28.31/25.47 & \cellcolor{c3}25.68/23.51 & \cellcolor{c4}46.37/44.82  \\
    DeepSeek-R1  & \textbf{57.83}/\textbf{57.12} & 54.05/\textbf{52.74} & \textbf{54.28}/\textbf{53.16} & \textbf{52.41}/\textbf{50.97} & \cellcolor{c1}\textbf{54.33}/\textbf{52.81} & 32.17/30.84 & \cellcolor{c2}32.17/30.84 & 26.85/24.63 & 30.45/28.92 & \cellcolor{c3}28.04/26.37 & \cellcolor{c4}\textbf{47.11}/\textbf{45.68}  \\
\bottomrule
\end{tabular}}
% \vspace{-3mm}
\caption{Results of goal achievement evaluation ($\mathbb{GAE}$) task. The score is the average goal achievement ratio (\%).}
\label{goal_evaluation}
\vspace{-1.8mm}
\end{table*}

\begin{table}[t]
\centering
\setlength{\tabcolsep}{1.4mm}
\resizebox{\columnwidth}{!}{
\begin{tabular}{l c c c c c c}
\toprule
    \multirow{2}{*}{\makecell[c]{Models}} 
    & \multicolumn{1}{c}{\makecell[c]{Social \\ Engagement}}
    & \multicolumn{1}{c}{\makecell[c]{Cooperation}} 
    & \multicolumn{1}{c}{\makecell[c]{Self- \\ Management}} 
    & \multicolumn{1}{c}{\makecell[c]{Emotional \\ Resilience}} 
    & \multicolumn{1}{c}{\makecell[c]{Innovation}} 
    & \multicolumn{1}{c}{\makecell[c]{\textbf{Overall}}} \\
    % \cmidrule(lr){2-3} \cmidrule(lr){4-5} \cmidrule(lr){6-7}  \cmidrule(lr){8-9} \cmidrule(lr){10-11} \cmidrule(lr){12-13}
    & \texttt{zh}/\texttt{en} & \texttt{zh}/\texttt{en} & \texttt{zh}/\texttt{en} & \texttt{zh}/\texttt{en} & \texttt{zh}/\texttt{en} & \texttt{zh}/\texttt{en} \\ 
\midrule
    Human (best)  & 84.85/85.71 & 89.56/92.86 & 86.32/80.56 & 81.76/86.67 & 81.48/85.71 & \cellcolor{c4}85.73/\cellcolor{c4}85.32  \\
    Human (average)  & 79.51/82.16 & 82.65/84.57 & 80.46/74.53 & 78.61/79.06 & 76.84/79.91 & \cellcolor{c4}80.22/\cellcolor{c4}79.08  \\
\midrule
    \multicolumn{7}{c}{\makecell[c]{\textit{Closed-sourced LLMs}}} \\
\midrule
    % GPT-4o-mini  & 71.57/68.47 & 77.85/71.62 & 69.86/65.17 & 72.80/73.12 & 72.62/68.14 & \cellcolor{c4}72.46/\cellcolor{c4}68.46  \\
    GLM-4  & 73.49/70.58 & 80.44/73.25 & 72.12/65.55 & 75.74/72.15 & 72.46/67.84 & \cellcolor{c4}74.65/\cellcolor{c4}69.19  \\
    % o1-mini  & 73.12/70.98 & 79.12/76.09 & 73.20/70.31 & 75.48/72.14 & 73.12/70.95 & \cellcolor{c4}74.74/\cellcolor{c4}71.95  \\
    GPT-4o  & 74.39/73.28 & 80.65/76.04 & 72.49/69.12 & 75.63/74.68 & 70.23/67.59 & \cellcolor{c4}74.83/\cellcolor{c4}72.01  \\
    Claude-3-sonnet  & 74.87/73.68 & 79.86/77.42 & 73.49/71.86 & 74.97/71.85 & 72.79/69.14 & \cellcolor{c4}75.24/\cellcolor{c4}73.16  \\
    GPT-4  & 73.59/71.68 & 80.56/76.86 & 73.98/70.07 & 77.11/76.33 & 74.11/69.71 & \cellcolor{c4}75.73/\cellcolor{c4}72.64  \\
    o1  & 76.24/74.67 & 82.44/80.76 & 75.12/73.11 & \textbf{78.10}/76.57 & 74.99/73.12 & \cellcolor{c4}77.27/\cellcolor{c4}\textbf{75.49}  \\
    Claude-3-opus  & \textbf{77.49}/\textbf{76.48} & \textbf{84.46}/\textbf{82.46} & 74.61/72.48 & 76.94/72.37 & 75.16/72.13 & \cellcolor{c4}77.57/\cellcolor{c4}75.27  \\
\midrule
    \multicolumn{7}{c}{\makecell[c]{\textit{Open-sourced LLMs}}} \\
\midrule
    % Baichuan2-7B  & 41.67/40.59 & 51.47/50.15 & 50.64/49.54 & 59.76/57.48 & 42.41/41.41 & \cellcolor{c4}49.82/\cellcolor{c4}48.52  \\
    % Baichuan2-13B  & 53.28/50.16 & 54.74/52.40 & 52.30/51.10 & 61.32/59.48 & 49.02/45.18 & \cellcolor{c4}54.02/\cellcolor{c4}51.94  \\
    % Yi-1.5-9B  & 57.56/54.16 & 61.14/58.14 & 58.46/54.31 & 57.49/54.19 & 45.97/42.65 & \cellcolor{c4}57.81/\cellcolor{c4}54.22  \\
    Mistral-7B-v0.3  & 59.42/55.23 & 64.15/62.11 & 58.46/52.55 & 55.41/52.46 & 49.88/47.68 & \cellcolor{c4}58.78/\cellcolor{c4}54.68  \\
    Mistral-8*7B-v0.1  & 62.12/57.25 & 65.16/63.14 & 60.22/58.75 & 57.56/55.41 & 52.16/50.11 & \cellcolor{c4}60.65/\cellcolor{c4}58.29  \\
    Llama-3.1-8B  & 61.04/59.79 & 62.63/57.61 & 59.47/55.02 & 63.42/59.15 & 56.82/51.72 & \cellcolor{c4}60.78/\cellcolor{c4}56.79  \\
    % Yi-1.5-34B  & 61.46/57.49 & 65.34/61.46 & 59.97/57.92 & 61.56/58.54 & 51.11/50.73 & \cellcolor{c4}60.95/\cellcolor{c4}58.15  \\
    % Phi-3.5-mini-instruct  & 62.41/60.47 & 65.14/64.19 & 58.52/55.95 & 61.74/57.96 & 57.89/54.74 & \cellcolor{c4}61.04/\cellcolor{c4}58.71  \\
    % Phi-3.5-moe-instruct  & 64.85/61.09 & 66.74/65.69 & 59.18/56.47 & 62.41/58.65 & 58.44/55.14 & \cellcolor{c4}62.22/\cellcolor{c4}59.47  \\
    Mistral-8*22B-v0.1  & 62.08/58.45 & 66.47/64.15 & 63.41/61.74 & 58.45/55.98 & 60.35/57.86 & \cellcolor{c4}62.89/\cellcolor{c4}60.54  \\
    % Phi-4  & 65.18/63.45 & 68.43/64.12 & 60.17/57.68 & 64.27/60.48 & 60.11/56.09 & \cellcolor{c4}63.40/\cellcolor{c4}60.38  \\
    GLM-4-9B  & 64.60/61.42 & 69.48/64.47 & 61.94/59.98 & 63.91/60.71 & 67.05/62.37 & \cellcolor{c4}64.66/\cellcolor{c4}61.46  \\
    Qwen-2.5-7B  & 64.58/62.10 & 74.47/71.22 & 61.20/58.46 & 65.61/62.12 & 57.14/52.25 & \cellcolor{c4}64.93/\cellcolor{c4}61.87  \\
    Qwen-2.5-14B  & 67.48/64.16 & 74.85/72.45 & 66.19/63.19 & 67.75/63.71 & 65.32/61.02 & \cellcolor{c4}68.40/\cellcolor{c4}65.22  \\
    Qwen-2.5-32B  & 67.64/64.36 & 75.12/72.61 & 67.59/63.67 & 69.97/64.32 & 66.79/62.20 & \cellcolor{c4}69.45/\cellcolor{c4}65.65  \\
    Qwen-2.5-72B  & 68.93/64.55 & 75.69/73.46 & 68.28/64.59 & 71.24/65.17 & 68.95/64.41 & \cellcolor{c4}70.41/\cellcolor{c4}66.51  \\
    Llama-3.1-70B  & 66.81/63.15 & 77.53/73.56 & 69.25/65.20 & 74.56/69.48 & 67.63/63.45 & \cellcolor{c4}71.15/\cellcolor{c4}67.04  \\
    Llama-3.3-70B  & 68.11/63.98 & 78.12/74.42 & 69.74/65.44 & 75.41/68.96 & 66.48/64.42 & \cellcolor{c4}71.75/\cellcolor{c4}67.46  \\
    DeepSeek-V3 & 75.62/74.18 & 81.37/78.49 & 74.93/72.41 & 77.58/76.31 & 74.88/72.89 & \cellcolor{c4}76.51/\cellcolor{c4}73.81 \\
    DeepSeek-R1 & 77.28/76.34 & 83.52/81.87 & \textbf{75.24}/\textbf{73.38} & 78.06/\textbf{76.62} & \textbf{75.19}/\textbf{73.22} & \cellcolor{c4}\textbf{77.62}/\cellcolor{c4}75.44 \\
\bottomrule
\end{tabular}}
% \vspace{-3mm}
\caption{Results (\%) of interpersonal ability evaluation ($\mathbb{IAE}$). The score is average ability selection accuracy.}
\label{ability_evaluation}
% \vspace{-4mm}
\end{table}

\begin{table}[t]
\centering
\resizebox{\columnwidth}{!}{
\begin{tabular}{l c c c}
\toprule
    \multirow{2}{*}{\makecell[c]{Models}} 
    & \multicolumn{1}{c}{\makecell[c]{Similarity Ratio}}
    & \multicolumn{1}{c}{\makecell[c]{Selection Ratio}} 
    & \multicolumn{1}{c}{\makecell[c]{\texttt{kappa}}} \\
    & \texttt{zh}/\texttt{en} & \texttt{zh}/\texttt{en} & \texttt{zh}/\texttt{en}  \\ 
\midrule
    GPT-4o                 & \textbf{0.75}/0.64 & \textbf{0.87}/\textbf{0.85} & 0.67/0.53 \\
\midrule
    Qwen-2.5-72B   & 0.72/\textbf{0.72} & 0.84/\textbf{0.85} & 0.60/0.48 \\
    Llama-3.1-70B & 0.72/0.66 & 0.85/0.82 & 0.67/0.59 \\
    Average       & 0.72/0.69 & 0.85/0.84 & 0.64/0.54 \\
\midrule
    Qwen-2.5-7B    & 0.57/0.43 & 0.76/0.72 & 0.59/0.37 \\
    Llama-3.1-8B  & 0.58/0.56 & 0.72/0.74 & 0.62/0.60 \\
    Average        & 0.58/0.50 & 0.74/0.73 & 0.61/0.49 \\ 
\midrule   
    Average-Overall        & 0.67/0.60 & 0.81/0.80 & 0.63/0.51 \\
\midrule
\end{tabular}}
% % \vspace{-3mm}
\caption{The semantic similarity ratio between the utterances generated by LLMs and the candidate utterances used for episode transitions, as well as the ratio of similar candidate utterances selected by LLMs in $\mathbb{GAE}$ task.}
\label{understanding_vs_generation}
% \vspace{-3.1mm}
\end{table}

\section{Experiments}

We evaluate 19 LLMs: \textbf{(1) Closed-source}: GPT series (4, 4o, o1, \citealt{o1}), Claude-3 series (sonnet, opus, \citealt{claude}), GLM-4 \cite{chatglm}. \textbf{(2) Open-source}: Llama-Instruct series (3.1-8B, 3.1-70B, 3.3-70B, \citealt{llama3modelcard}), Qwen-2.5-Instruct series (7B, 14B, 32B, 72B, \citealt{qwen2_5}), Mistral-Instruct series (7B-v0.3, 8*7B-v0.1, 8*22B-v0.1, \citealt{mistral}), GLM-4-9B-Chat, DeepSeek series (v3, r1, \citealt{deepseek_v3,deepseek_r1}). More evaluated LLMs are in Appendix \ref{sec:more_llm_results_appendix}.

We use Chain-of-Thought to prompt LLMs to perform outcome-oriented goal achievement evaluation ($\mathbb{GAE}$) and process-oriented interpersonal ability evaluation ($\mathbb{IAE}$) tasks (\citealt{cot}, prompts are in Appendix \ref{sec:evaluation_prompts_appendix}). The metrics are \textit{goal achievement ratio} and \textit{ability selection accuracy}. To avoid position bias, we randomly shuffle the options' order for each sample 3 times and determine the LLM's final choice by majority vote. We establish human baselines by hiring 20 graduate students, native in Chinese and English, to respectively complete tasks in two languages, each with 14 samples for $\mathbb{GAE}$ and 160 samples for $\mathbb{IAE}$. We obtain \textbf{Human (best)} and \textbf{Human (average)} baselines by taking the best human result for each task and averaging the results of all participants.

\subsection{Results and Findings}
\label{sec:gae_and_iae_results}

We report the results of LLMs and humans on two tasks of SI evaluation in Tables \ref{goal_evaluation} and \ref{ability_evaluation}. Detailed results on 32 specific abilities are in Appendix \ref{sec:32_ability_results}.

\paragraph{LLMs' SI performance.}

\textbf{First}, open-source LLMs have slightly surpassed closed-source models in both tasks, e.g., DeepSeek-R1 outperforms o1 by 0.1\%/0.6\% ($\mathbb{GAE}$) and Claude-3-opus by 0.1\%/0.2\% ($\mathbb{IAE}$) at the Overall level (\texttt{zh/en}).
\textbf{Second}, open-source LLMs exhibit a positive correlation between performance and size in both tasks, showing that LLMs' SI strengthens as the parameters grow.
\textbf{Third}, machine SI still lags behind: all LLMs' performance is distinctly lower than that of humans', with the smallest gap being 23.8\%/17.2\% and 3.2\%/4.6\% (\texttt{zh/en}, \textit{Human-average vs. best LLMs}) in $\mathbb{GAE}$ and $\mathbb{IAE}$ tasks at the Overall level.

\paragraph{LLMs exhibit more distinct prosociality than humans.}
\textbf{First}, humans perform better in prosocial worlds than in proself and antisocial worlds by 15.3\%/33.2\% and 21.0\%/16.5\% (\texttt{zh/en}) at the Average level, whereas LLMs (e.g., DeepSeek-R1) show a more significant gap, with 40.8\%/41.6\% and 48.4\%/50.0\%.
\textbf{Second}, humans show comparable performance in antisocial worlds (i.e., Induction) as in prosocial worlds. This can be attributed to humans’ ability to flexibly follow various profiles when achieving social goals, as reflected in feedback from our participants. However, LLMs fail to match this human trait.
\textbf{Third}, both humans and LLMs show exceptional Cooperation ability in $\mathbb{IAE}$ task, with even Claude-3-opus outperforming humans in the Chinese evaluation. This also shows LLMs prefer prosociality, aligning with that \citet{llm_cooperation} highlights LLMs spontaneously display strong cooperative behaviors.

\paragraph{Both humans and LLMs show significant cross-lingual differences in SI.}

First, we conduct a Kolmogorov-Smirnov test \cite{kolmogorov_smirnov} on the Chinese and English results of LLMs, finding that both $\mathbb{GAE}$ and $\mathbb{IAE}$ results do not follow a normal distribution ($p$\textless 0.05). Thus, we then use the Wilcoxon Signed-Rank test \cite{wilcoxon} to examine the differences caused by language, yielding $p$\textless 0.001 on $\mathbb{GAE}$ and $\mathbb{IAE}$. This shows that LLMs exhibit significant cross-lingual differences in SI, consistent with those observed in human results.

\subsection{Behavior-Level Analysis of LLMs' SI}
\label{sec:behavior_level_analysis}

\begin{figure}[t]
    \centering
    \includegraphics[width=\columnwidth]{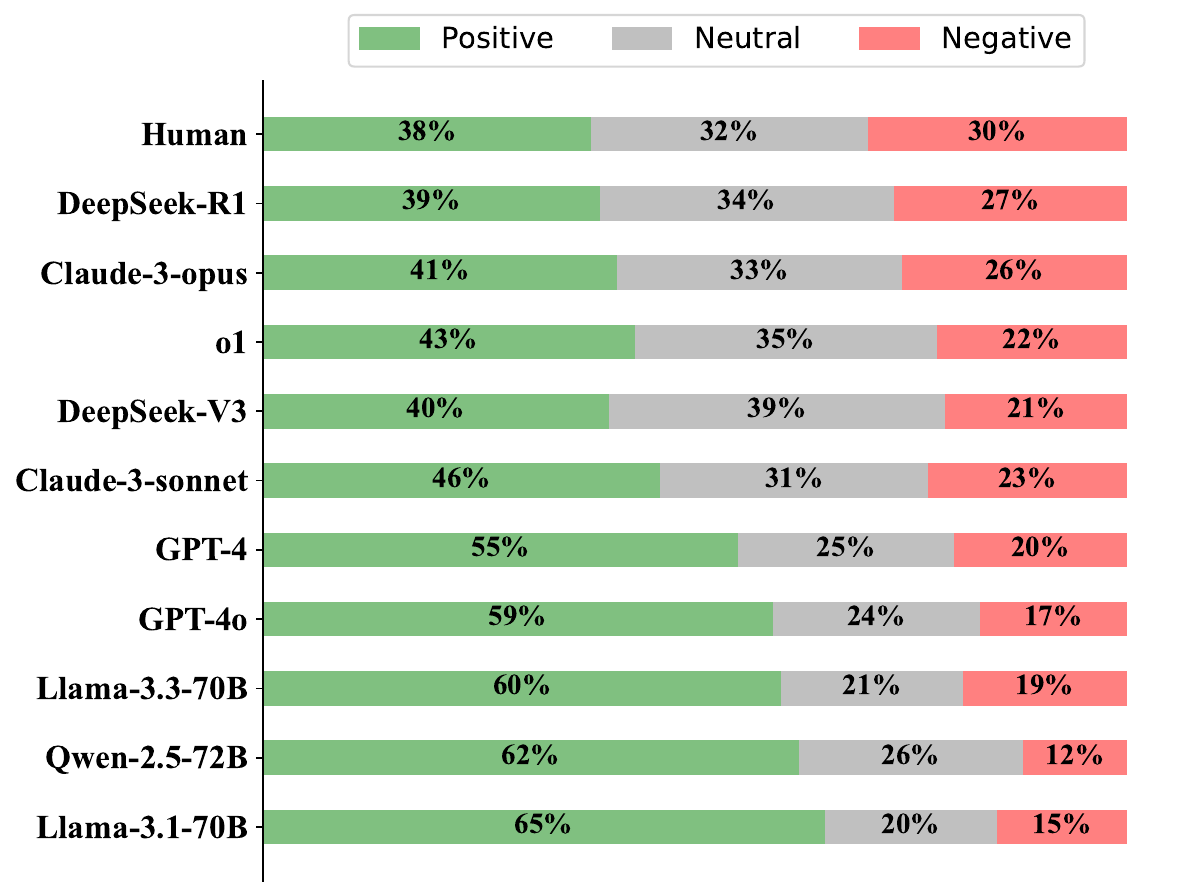}
    % \vspace{-3mm}
    \caption{Behavioral distribution of selections made by LLMs and humans at the same episode transitions.}
    \label{overall_behvior_distribution}
     % \vspace{-1mm}
\end{figure}

\begin{figure}[t]
    \centering
    \includegraphics[width=\columnwidth]{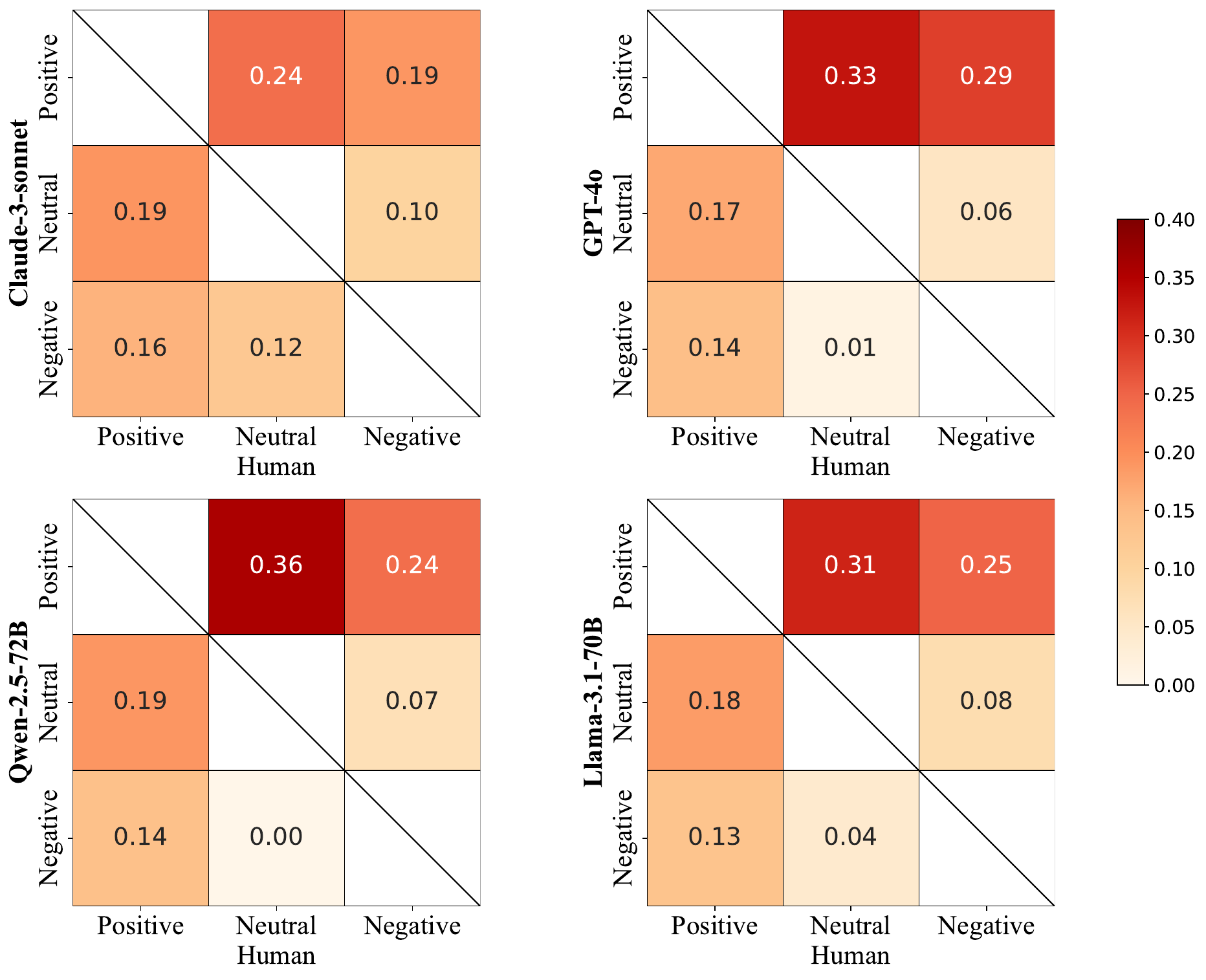}
    % \vspace{-7mm}
    \caption{Distribution of behavior combinations shown by LLMs and humans at the same episode transition.}
    \label{compared_behavior_distribution}
     % \vspace{-3mm}
\end{figure}

\begin{figure}[t]
    \centering
    \begin{subfigure}[b]{0.43\textwidth}
        \centering
        \includegraphics[width=.93\textwidth]{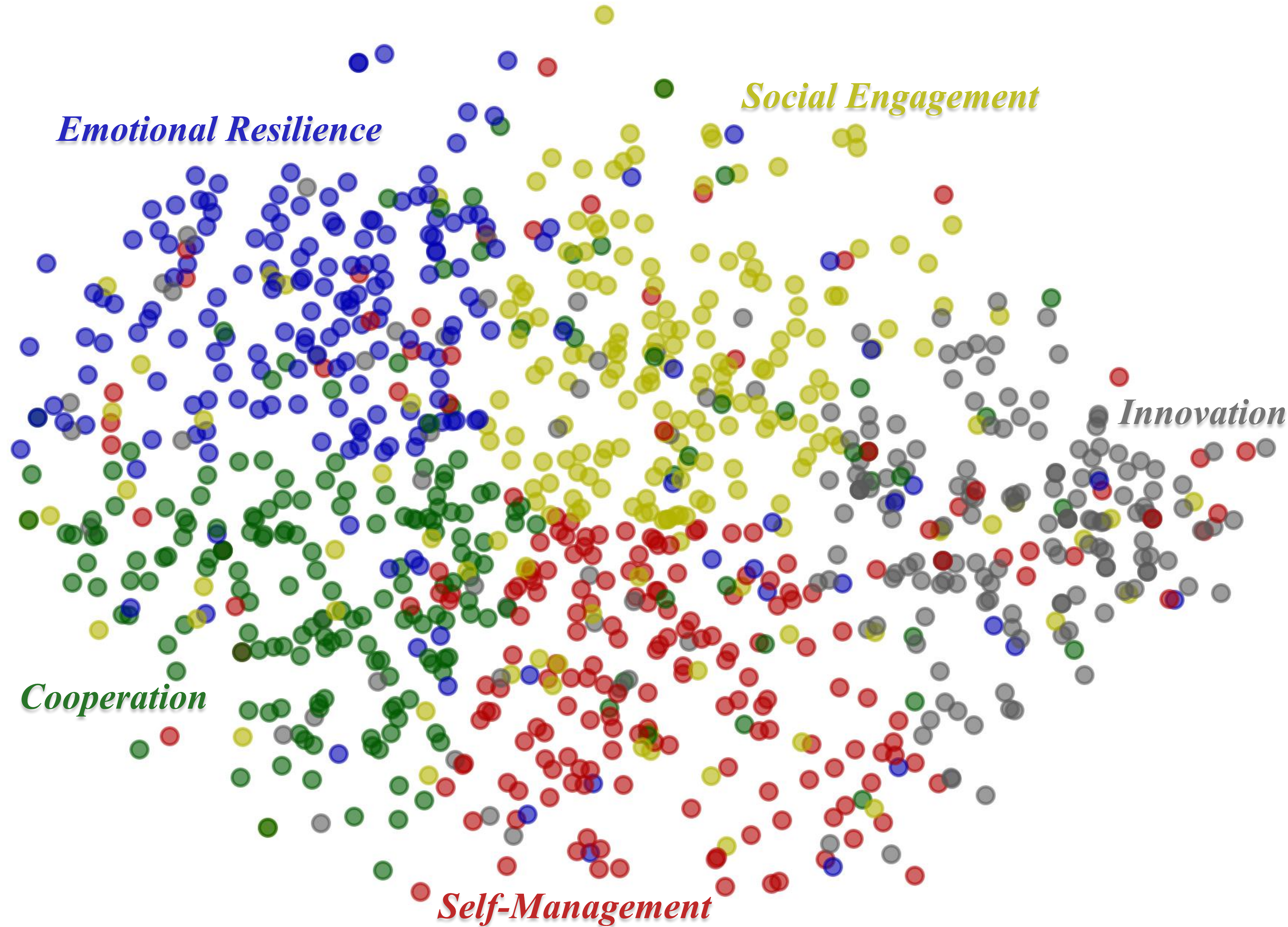}
        \caption{Clusters of interpersonal abilities in Llama-3.1-8B.}
        \label{llama_3.1_8b_cluster}
    \end{subfigure}
    \hfill
    \begin{subfigure}[b]{0.43\textwidth}
        \centering
        \includegraphics[width=.93\textwidth]{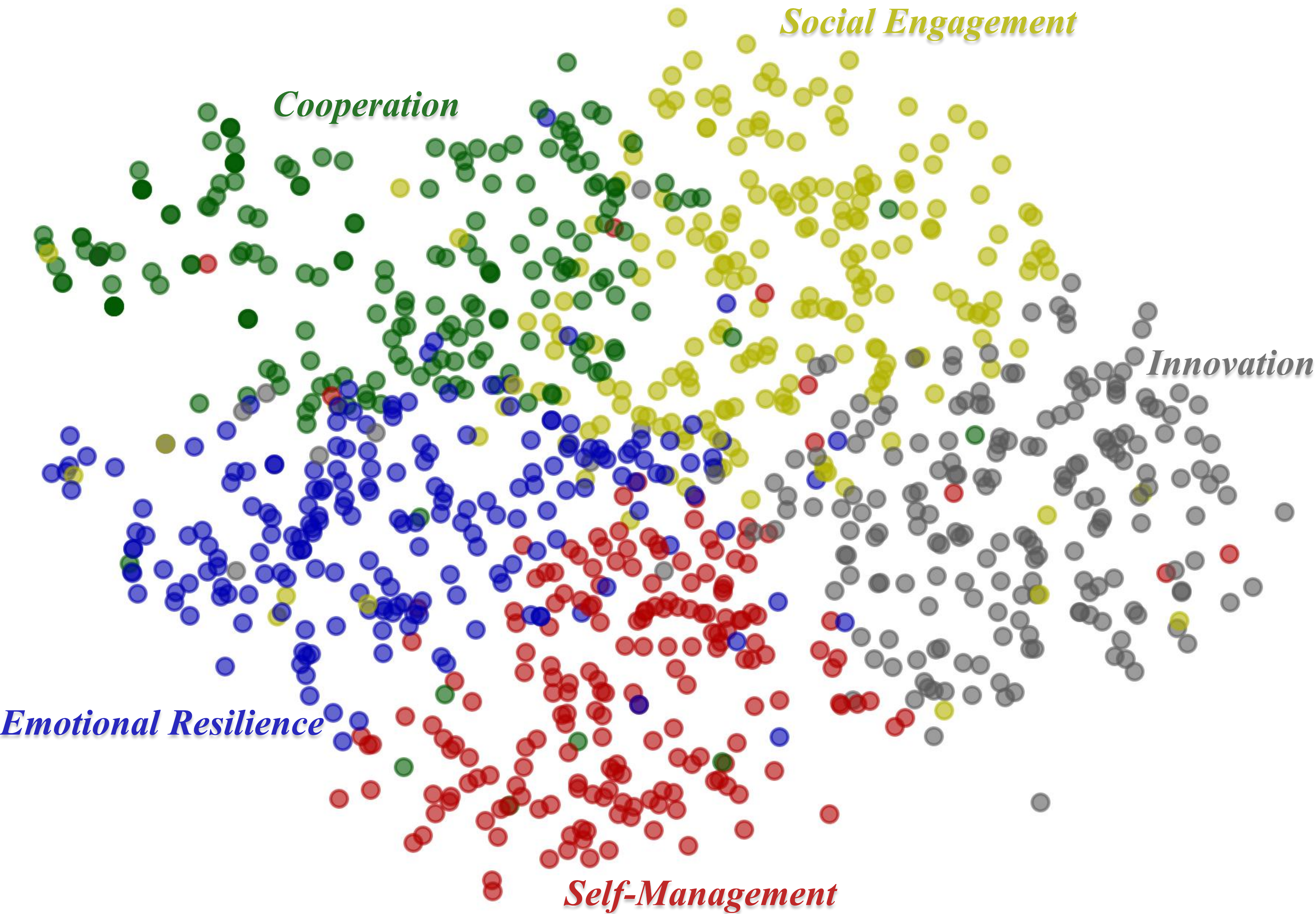}
        \caption{Clusters of interpersonal abilities in Llama-3.1-70B.}
        \label{llama_3.1_70b_cluster}
    \end{subfigure}
    \caption{Cluster distributions of interpersonal abilities in the representation space of Llama-3.1-8B \& 70B.}
    % \vspace{-3mm}
    \label{interpersonal_ability_cluster}
\end{figure}

Social behavior, as the external manifestation of SI, reflects how well an individual navigates complex social dynamics \cite{social_behavior}. We analyze behavioral discrepancies between LLMs and humans by examining their selection of candidate utterances during episode transitions in the $\mathbb{GAE}$ task.

\paragraph{LLMs prefer to exhibit more positive behaviors, even when they lead to goal failure, while humans are more flexible in adjusting their behaviors to reach goals.}

Here, we manually annotate the behavioral polarity of each candidate utterance in \textsc{SocialEval}, i.e., whether the expressed feelings, thoughts, and actions are \textbf{positive}, \textbf{neutral}, or \textbf{negative}. We focus on world trees where LLMs fail, but humans succeed. To identify their behavioral discrepancies, we retain only the trees with a single successful ending, where LLMs and humans make different selections at the same episode transitions with varying behavioral polarities and endings. 
The behavioral distribution of 10 LLMs and humans are shown in Figure \ref{overall_behvior_distribution}. Results show that LLMs prefer positive behaviors to drive plot progression, while models whose behavior distribution aligns more closely with that of humans tend to perform better in $\mathbb{GAE}$ task, as supported by Table \ref{goal_evaluation}. Moreover, in Figure \ref{compared_behavior_distribution}, we visualize the distribution of different behavior combinations taken by LLMs and humans at the same episode transition (more results are in Appendix \ref{sec:more_analysis}). Results further indicate that, compared to humans, LLMs are more likely to select positive behaviors, even if they lead to goal failure. 
We also report the behavioral distribution of LLMs and humans from all world trees in $\mathbb{GAE}$ task, revealing similar findings (App. \ref{sec:more_analysis}).

\paragraph{Multiple-choice questions-based SI evaluation can effectively reflect LLMs' generation capability.}

We measure the difference between LLMs' behaviors, i.e., selecting the options (our work) vs. direct dialogue generation, in SI evaluation in the $\mathbb{GAE}$ task. Specifically, given the preceding plot at episode transitions, we prompt the LLMs to generate protagonist's next utterance (prompt is in Appendix \ref{sec:roleplay_prompt_appendix}). Each LLM produces 100 samples in both \texttt{zh} and \texttt{en}. We hire annotators to evaluate the semantic similarity between the generated and candidate utterances, with each sample being labeled by three annotators. In Table \ref{understanding_vs_generation}, LLMs exhibit a high semantic similarity ratio, significantly increasing with model size. This shows that our crafted candidate utterances align closely with what the LLMs would generate in the given context.
%, and thus validates our MCQ-based evaluation. 
Moreover, when semantically similar utterances exist in the choice set, the average ratio of LLMs selecting similar candidate utterances exceeds 80\%. %This shows that LLMs’ choices in MCQs are highly correlated with their context-based generation results.

\begin{figure*}[t]
    \centering
    \includegraphics[width=.95\textwidth]{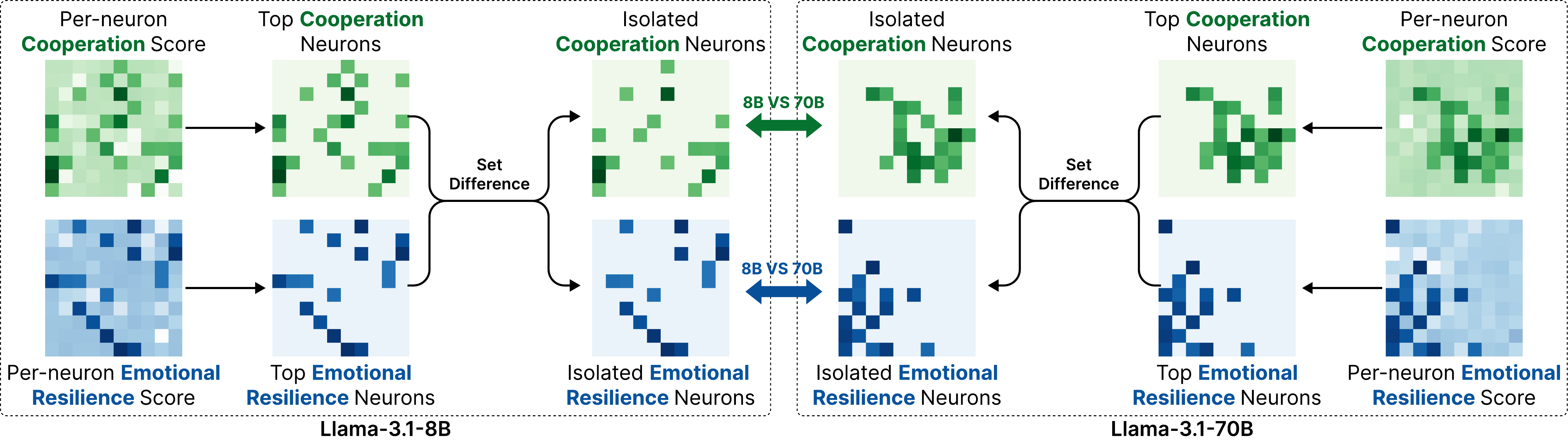}
    % \vspace{-3mm}
    \caption{Activated neurons of interpersonal abilities (Cooperation and Emotional Resilience) in Llama-3.1-8B \& 70B. We identify the top neurons for each ability by computing per-neuron importance scores, then isolate cooperation-critical neurons from emotional resilience neurons using set differences, and vice versa.}
    \label{neuron_importance_cooperation_emotion}
    % \vspace{-3.5mm}
\end{figure*}

\subsection{In-depth Analysis of LLMs’ SI at the Representation and Neuron Level}
\label{sec:representation_neron_level_analysis}

Neurons clustered in human brains form specific functional groups \cite{human_brain}. By analogy, we analyze whether similar features of interpersonal abilities can be found in LLMs. Here, we employ Llama-3.1-8B \& 70B as the backbone models, concatenating the question and the correct candidate utterance for 5 aspects of interpersonal abilities in the $\mathbb{IAE}$ as the input for the LLMs. We exclude samples that involve composite abilities.

\paragraph{LLMs' interpersonal abilities exhibit a clustered distribution in the representation space.}

We cluster and visualize the last input token's hidden state output by the LLM's last layer using t-SNE.
In Figure \ref{interpersonal_ability_cluster}, the 8B model partially distinguishes between 5 ability aspects, but there is still notable overlap between them. Yet, the 70B model more clearly separates these 5 aspects into distinct clusters. This shows as LLMs' size increases, interpersonal abilities tend to evolve into specific groups.

\paragraph{LLMs' neurons activated by specific interpersonal abilities form isolated regions.}

We use the Wanda score \cite{wanda_score} to identify activated neurons (details are in Appendix \ref{sec:wanda_score}). We first separately average all inputs' neuron activation matrices of each interpersonal ability. Following \citet{npti}, we then average the per-neuron importance scores in blocks of size 256×256 to reduce the high-dimensional neuron matrices into a lower dimension. Finally, we retain the top weights to define the neuron region for each interpersonal ability and isolate these regions using the set difference between the neuron regions of two abilities \cite{safety_neuron}. We show the isolated neuron regions for Cooperation and Emotional Resilience in Figure \ref{neuron_importance_cooperation_emotion}. More results are in App. \ref{sec:neuronal_activation_appendix}. In the 8B model, each interpersonal ability shows an isolated but sparse neuron region, while in the 70B model, these regions are more densely packed. This shows that as LLMs’ size increases, their SI evolves to form ability-specific partitions at the neuron level, as reported by ``lobes'' hypothesis of \citet{feature_lobes}.

\section{Related Work}

As LLMs' capabilities grow \cite{deepseek_r1}, researchers start to explore whether LLMs possess human-like SI \cite{characterglm,crisp}. A common method is to use LLMs for society simulations to model human behaviors \cite{ai_town}, which reveal LLMs' social dynamics  \cite{elicitron} and show parallels to human behaviors \cite{camel,simulate_human_trust_behavior}, e.g., achieving goals \cite{sotopia} and social norms evolution \cite{social_norm_evolution,social_norm_emergence}. This has led to reports of emergent SI in LLMs \cite{si_emergence} and sparked interest in using LLMs to tackle interpersonal situations \cite{improve_communication,lin2024imbue} and test social science theories \cite{simulate_human_collective_intelligence}, e.g., prosociality \cite{llm_prosocial_irrationality} and cooperation \cite{llm_cooperation}.

To evaluate LLMs' SI, there are three types of benchmarks. The first is inspired by clinical tests of SI for humans to assess theory-of-mind \cite{tomi,fauxpas,tombench}, 
% i.e., the ability to perceive and attribute mental states to oneself and others. 
The second evaluates LLMs' social understanding via social commonsense reasoning \cite{socialiqa,socailiq}. The third, closely related to our work \cite{sotopia,stss,agentsense}, asks LLMs to achieve predefined goals in an interactive situation via role-playing \cite{characterbench}. Yet, they are often limited to single-episode social dynamics and lack a fine-grained interactive process design for evaluating goal pursuit, thus hindering a holistic SI evaluation.

\section{Conclusions}

In this paper, we propose \textsc{SocialEval}, a script-based bilingual SI benchmark, integrating outcome-oriented goal achievement evaluation and process-oriented interpersonal ability evaluation via manually crafted narrative scripts. Each script forms a world tree with a well-designed plot evolution, providing a comprehensive view of how LLMs navigate social interactions. Experiments systemically analyze the similarities and discrepancies between humans' and LLMs' SI, from surface-level SI score to behavioral comparison during goal pursuit and to ability-specific functional partitions in LLMs, offering new insights for future work in LLMs' SI.

\section*{Limitations}
\label{sec:limitations}

We discuss the limitations of this work as follows.

\paragraph{Benchmark Scale}

In \textsc{SocialEval}, the average construction time for each world tree is approximately 12 hours, with an average cost of \$40. The entire process, from data construction to quality control, took about 4 months. The difficulty and cost of manual construction limit the scale of \textsc{SocialEval}, which includes only 153 world trees. Additionally, the antisocial worlds (i.e., Induction, Conflict), which involve harming others' interests, lack sufficient reference scripts from public sources, resulting in a smaller dataset for these worlds. Furthermore, each world tree is a coherent narrative script. At each episode transition, the options presented and the interpersonal abilities they exhibit are dynamically influenced by the preceding plot. Therefore, while we manually crafted 2,493 samples for interpersonal ability evaluation, the distribution across 32 interpersonal abilities is not balanced. There is an opportunity for future work to expand the benchmark scale, enabling a more detailed analysis of each social world and interpersonal ability.

\paragraph{Benchmark Language}

\textsc{SocialEval} supports bilingual evaluation, with our English data derived from Chinese originals through translation. Although we carefully designed prompts for translation using GPT-4o and achieved a 97\% acceptance rate through expert review, some cultural differences may still exist. Future efforts can build upon our construction methods to expand language coverage, better reflecting diverse cultural contexts.

\section*{Ethical Considerations}

In this work, we recruit a large number of human workers for benchmark construction, quality control, and human experiments, who are primarily college students. These workers are compensated fairly based on the market price. Our collected data do not contain any personal information. We are only responsible for publishing task information, and workers' privacy can be well preserved. We also declare that the antisocial worlds (i.e., Induction and Conflict) were constructed solely for research purposes, and they may contain sensitive and unethical content. We will release our data for research purposes. Our data is approved by the Institutional Review Boards, and we believe our work meets ACL's Code of Ethics.

\section*{Acknowledgements}

This work was supported by the National Key Research and Development Program of China (No. 2024YFC3606800).
This work was also supported by the National Science Foundation for Distinguished Young Scholars (with No. 62125604).

\bibliography{custom,socialeval}

\appendix

% \section{Appendix}
% \label{sec:appendix}

\newpage

\section{Preliminaries}

\begin{table*}[t]
\centering
\resizebox{\textwidth}{!}{
\begin{tabular}{c c c c l}
\toprule
    \multirow{2}{*}{\makecell[c]{Major \\ Orientations}} 
    & \multicolumn{2}{c}{\makecell[c]{Outcome}}
    & \multirow{2}{*}{\makecell[c]{Sub- \\ orientations}}
    & \multirow{2}{*}{\makecell[c]{Explanations}} \\
    \cmidrule(lr){2-3} 
    & \makecell[c]{Self-interest} & \makecell[c]{Altruism} &  & \\ 
\midrule
    \multirow{6}{*}{\makecell[c]{Prosocial}}
    & \makecell[c]{1} & \makecell[c]{1} & \makecell[c]{Cooperation} 
    & \makecell[l]{Individuals prioritize both their own and others' interests and tend to cooperate to achieve \\ mutually beneficial outcomes.} \\
    & \makecell[c]{1} & \makecell[c]{0} & \makecell[c]{Negotiation} 
    & \makecell[l]{Individuals prioritize their own interests and tend to negotiate to secure their best possible \\ outcomes.} \\
    & \makecell[c]{0} & \makecell[c]{1} & \makecell[c]{Assistance} 
    & \makecell[l]{Individuals value the interests of others and tend to offer assistance.} \\
    & \makecell[c]{-1} & \makecell[c]{1} & \makecell[c]{Altruism} 
    & \makecell[l]{Individuals sacrifice their own interests to support others, showing selflessness and altruism.} \\
\midrule
    \multirow{1}{*}{\makecell[c]{Proself}}
    & \makecell[c]{1} & \makecell[c]{-1} & \makecell[c]{Competition} 
    & \makecell[l]{Individuals prioritize their own interests, harm others’ interests of, and tend to achieve \\ their goals through competition.} \\
\midrule
    \multirow{3}{*}{\makecell[c]{Antisocial}}
    & \makecell[c]{0} & \makecell[c]{-1} & \makecell[c]{Induction} 
    & \makecell[l]{Individuals tend to influence others' behavior in order to make them bear a loss of benefits.} \\
    & \makecell[c]{-1} & \makecell[c]{-1} & \makecell[c]{Conflict} 
    & \makecell[l]{Individuals disregard both their own and others' interests, displaying antagonistic or destructive \\ behavior.} \\
\bottomrule
\end{tabular}}
\caption{Social World Taxonomy. The social worlds are classified into 3 major categories and their corresponding 7 subcategories based on interpersonal orientations. The taxonomy does not include $\texttt{(self-interest, altruism)} \in [(0,0), (0,-1)]$, as the former lacks clear social goals and the latter is rare in the real world.}
\label{social_world_taxonomy}
\end{table*}

\subsection{Social World Taxonomy}
\label{sec:social_world_taxonomy_appendix}

Detailed explanations about our defined 3 major orientations and 7 sub-orientations of social worlds are presented in Table \ref{social_world_taxonomy}.

\subsection{Details of Interpersonal Abilities}
\label{sec:interpersonal_ability_details}
Detailed explanations about our defined 5 aspects of interpersonal ability and 32 specific interpersonal abilities are presented in Table \ref{tab:interpersonal_abilities}.

\begin{table*}[t]
\centering
\resizebox{\textwidth}{!}{
\begin{tabular}{llll}
\toprule
\makecell[l]{Aspects} & \makecell[l]{Abilities} & \makecell[l]{Definition} \\
\midrule
\multirow{12}{*}{\makecell[l]{Self Management}} 
&  Task Management & The ability to maintain focus and discipline to complete tasks within deadlines, balancing quality and efficiency. \\
&  Time Management & Effectively allocating time to various tasks and goals, balancing priorities and ensuring that time is used productively. \\
&  Detail Management & Maintaining a high level of thoroughness and attention to all aspects of work, ensuring that no important detail is overlooked. \\
&  Organizational Skill & The ability to systematically arrange and structure personal spaces, tools, and tasks to enhance efficiency and ease of access. \\
&  Responsibility Management & Ensuring that commitments, promises, and responsibilities are met with reliability and accountability. \\
&  Capacity for Consistency & The ability to sustain steady performance in regular, routine tasks, regardless of external distractions or boredom. \\
&  Goal Regulation & The process of defining specific, measurable, and realistic goals, as well as maintaining the motivation and effort required to achieve them. \\
&  Rule-following Skill & Adhering to established rules, norms, and guidelines, both in structured environments and in everyday life. \\
&  Decision-Making Skill & The ability to make informed, balanced, and thoughtful choices by considering all relevant factors and potential consequences. \\
& Adaptability & The willingness and ability to try new things, respond to challenges, and modify behavior or thought processes when situations change. \\
&  Capacity for Independence & The ability to make decisions, set priorities, and manage tasks without relying on others for guidance or support. \\
&  Self-Reflection Skill & Engaging in thoughtful reflection on one’s thoughts, actions, and emotions to better understand oneself and improve behavior. \\
\midrule

\multirow{5}{*}{\makecell[l]{Social Engagement}} 
&  Leadership Skill & The ability to assert oneself in group settings, clearly communicating ideas and guiding discussions or decisions effectively. \\
&  Persuasive Skill & The ability to present ideas, arguments, and information in a compelling and convincing manner, influencing others’ opinions and decisions. \\
&  Conversational Skill & Initiating and sustaining conversations, including the ability to engage others, ask questions, listen actively, and provide relevant responses. \\
&  Expressive Skill & Effectively conveying personal thoughts, feelings, and experiences to others in ways that are both understandable and emotionally resonant. \\
&  Energy Regulation & Managing one’s energy levels and emotions to maintain productive, positive social interactions, avoiding burnout or overstimulation. \\
\midrule

\multirow{5}{*}{\makecell[l]{Cooperation}} 
&  Teamwork Skill & Collaborating effectively with others towards shared goals, contributing individual strengths while considering the needs and contributions of others. \\
&  Capacity for Trust & The ability to place trust in others, understanding their capabilities and motives, and being willing to forgive and move forward after conflicts. \\
&  Perspective-Taking Skill & The ability to see and understand the world from another person’s viewpoint, considering their emotions, needs, and reasoning. \\
&  Capacity for Social Warmth & The ability to make others feel welcomed, valued, and comfortable, creating positive and supportive social environments. \\
&  Ethical Competence & Upholding moral and ethical standards, even in difficult or ambiguous situations, while considering the impact of one’s actions on others. \\
\midrule

\multirow{5}{*}{\makecell[l]{Emotional Resilience}} 
&  Stress Regulation & Managing one’s responses to stress, anxiety, and fear, including using strategies to reduce stress and maintain emotional stability. \\
&  Capacity for Optimism & Maintaining a positive outlook, even in challenging situations, and finding hope or opportunity in adversity. \\
&  Anger Management & Recognizing and controlling the impulse to react with anger or irritation, responding to situations in a calm and rational manner. \\
&  Confidence Regulation & Maintaining self-assurance and a positive self-image, even in the face of criticism, failure, or uncertainty. \\
&  Impulse Regulation & Controlling immediate desires or urges that may lead to negative or undesired outcomes, making thoughtful decisions rather than acting on instinct. \\
\midrule

\multirow{5}{*}{\makecell[l]{Innovation}} 
&  Abstract Thinking Skill & Engaging with ideas that are theoretical, conceptual, or not immediately practical, exploring complex patterns and connections beyond concrete facts. \\
&  Creative Skill & The ability to generate novel and original ideas, approaches, or solutions, thinking outside conventional frameworks. \\
&  Artistic Skill & The ability to create or appreciate art, whether visual, musical or literary, using imagination and creativity to express or experience beauty. \\
&  Cultural Competence & Understanding and appreciating diverse cultural norms, and perspectives, and adapting behaviors to respect and integrate cultural differences. \\
&  Information Processing Skill & The ability to absorb, interpret, and apply new information quickly and effectively, using this knowledge to solve problems or create new insights. \\
\bottomrule
\end{tabular}}
\caption{Definitions of interpersonal ability inventory used in \textsc{SocialEval}, which contains 5 aspects of interpersonal abilities across 32 specific abilities.}
\label{tab:interpersonal_abilities}
\end{table*}

\section{Prompt for Data Translation}
\label{sec:translation_prompt_appendix}
The translation prompt is shown in Table \ref{trans_prompt}, which translates \textsc{SocialEval} from Chinese into English.

% Table \ref{trans_prompt} presents the prompt template for translating our Chinese social interactive game data into English.

\begin{table*}[t]
\scriptsize
\centering
\resizebox{\textwidth}{!}{
\begin{tabular}{|m{\textwidth}|}
\hline
    You are an experienced translator who only uses English in translating all texts.\\
    \\
    
    [Task]
    
    Translate the given Chinese social interactive game data to English. You should strictly follow the below rules.
    
    1. Return translations in correct JSON format with all key-value pairs intact. With no loss of any information. Especially, everything in interactive plot should be translated.
    
    2. Use idiomatic and context-appropriate English, varying between formal and informal tones as needed.
    
    3. Present only translation results without additional explanations.
    
    4. Maintain consistency in names and titles throughout the text.
    
    5. Align the tone of dialogues with the character profiles to accurately reflect personality and mood.
    
    6. Identify and correctly translate proper nouns, including historical and geographical terms.
    
    7. Preserve the original text's order, meaning, tone, and emotion.
    
    8. Adapt the translation tone to match the context, using appropriate colloquialisms or formal language as dictated by the dialogue.
    
    9. Pay close attention to idiomatic expressions, translating their implied rather than literal meanings.
    
    10. Ensure pronoun references are clear and contextually appropriate, particularly in complex dialogues.\\
    \\
    
    [Chinese social interactive game data]:
    
    \{\{
    
    \{data\}
    
    \}\}\\
    \\
    
    [OUTPUT English Translation]:\\
\hline
\end{tabular}}
\caption{Prompt to translate \textsc{SocialEval} from Chinese into English. \texttt{\{data\}} is a placeholder.}
\label{trans_prompt}
\end{table*}

\section{Experiments}

\subsection{Evaluation Prompts for Two Tasks}
\label{sec:evaluation_prompts_appendix}

We present the evaluation prompts for two tasks, outcome-oriented goal achievement evaluation ($\mathbb{GAE}$) task and process-oriented interpersonal ability evaluation ($\mathbb{IAE}$) task, in Table \ref{gae_eval} and \ref{iae_eval}. Both prompts are provided in Chinese and English.

\begin{table*}[t]
\scriptsize
\centering
\resizebox{\textwidth}{!}{
\begin{tabular}{|m{\textwidth}|}
\hline
    \begin{CJK*}{UTF8}{gbsn}请你扮演\{character\_name\}，给定\{character\_name\}的信息，其中包含了\{character\_name\}的公开信息、隐私信息和在社交场景中要实现的社交目标，同时，给定在社交中其他角色的信息，请你基于给定的角色信息、\{character\_name\}的社交目标和\{character\_name\}和其他角色的对话上下文，从给定的选项中选择一个最有可能达成目标结局的\{character\_name\}的回复选项，并给出解释，输出json格式。\end{CJK*}\\
    
    \begin{CJK*}{UTF8}{gbsn}[输出示例]
    
    \{\{
    
    "explaination": "解释",
    
    "choice": "A"
    
    \}\}\end{CJK*}\\
    \\
    
    \begin{CJK*}{UTF8}{gbsn}[\{character\_name\}的信息]

    公开信息：\{public\}
    
    隐私信息：\{private\}
    
    社交目标：\{goal\}\end{CJK*}\\
    \\
    
    \begin{CJK*}{UTF8}{gbsn}[其他角色的信息]
    
    \{user\_profile\}\end{CJK*}\\
    \\
    
    \begin{CJK*}{UTF8}{gbsn}[对话上下文]
    
    \{dialogue\_context\}\end{CJK*}\\
    \\
    
    \begin{CJK*}{UTF8}{gbsn}[选项]
    
    \{choices\}\end{CJK*}\\
    \\
\hline
    Please act as \{character\_name\}. You are provided with information about \{character\_name\}, which includes \{character\_name\}'s public information, private information, and social goals to achieve in the social scenario. Additionally, information about other roles in the social interaction is provided. Based on the given role information, \{character\_name\}'s social goal and the dialogue context involving \{character\_name\} and other roles, choose the option that is most likely to achieve \{character\_name\}'s social goal from the given options with your explanation. Output the result in JSON format. \\
    \\
    
    [Example output]
    
    \{\{
    
    "explanation":"explanation to your answer",
    
    "choice": "A"
    
    \}\}\\
    \\
    
    [\{character\_name\}'s Information]
    
    Public Information: \{public\}
    
    Private Information: \{private\}
    
    Social Goal: \{goal\}\\
    \\
    
    [Other Roles' Information]
    
    \{user\_profile\}\\
    \\
    
    [Dialogue Context]
    
    \{dialogue\_context\}\\
    \\
    
    [Options]
    
    \{choices\}\\
    \\
\hline
\end{tabular}}
\caption{Evaluation prompt (\texttt{zh} \& \texttt{en}) for goal achievement evaluation ($\mathbb{GAE}$) task. \texttt{\{character\_name\}}, \texttt{\{public\}}, \texttt{\{private\}}, \texttt{\{goal\}}, \texttt{\{user\_profile\}}, \texttt{\{dialogue\_context\}} and \texttt{\{choices\}} are placeholders.}
\label{gae_eval}
\end{table*}

\begin{table*}[t]
\scriptsize
\centering
\resizebox{\textwidth}{!}{
\begin{tabular}{|m{\textwidth}|}
\hline
    \begin{CJK*}{UTF8}{gbsn}请你扮演\{character\_name\}，给定\{character\_name\}的信息，其中包含了\{character\_name\}的公开信息、隐私信息和在社交场景中要实现的社交目标，同时，给定在社交中其他角色的信息，请你基于给定的角色信息、\{character\_name\}和其他角色的对话上下文、和提问，从给定的选项中选择一个正确的答案并给出解释，输出json格式。\end{CJK*}\\
    \\
    
    [\begin{CJK*}{UTF8}{gbsn}输出示例\end{CJK*}]

    \{\{
    
    "explaination": "\begin{CJK*}{UTF8}{gbsn}解释\end{CJK*}",
    
    "choice": "A"
    
    \}\}\\
    \\
    
    [\{character\_name\}\begin{CJK*}{UTF8}{gbsn}的信息\end{CJK*}]

    \begin{CJK*}{UTF8}{gbsn}公开信息：\{public\}
    
    隐私信息：\{private\}
    
    社交目标：\{goal\}\end{CJK*}\\
    \\
    
    \begin{CJK*}{UTF8}{gbsn}[其他角色的信息]
    
    \{user\_profile\}\end{CJK*}\\
    \\
    
    \begin{CJK*}{UTF8}{gbsn}[对话上下文]
    
    \{dialogue\_context\}\end{CJK*}\\
    \\
    
    \begin{CJK*}{UTF8}{gbsn}[提问]
    
    \{question\}\end{CJK*}\\
    \\
    
    \begin{CJK*}{UTF8}{gbsn}[选项]
    
    \{choices\}\end{CJK*}\\
    \\
\hline
    Please act as \{character\_name\}. You are provided with information about \{character\_name\}, which includes \{character\_name\}'s public information, private information, and social goals to achieve in the social scenario. Additionally, information about other roles in the social interaction is provided. Based on the given role information, the dialogue context involving \{character\_name\} and other roles, and the question, choose the correct answer from the given options with your explanation. Output the result in JSON format. \\
    \\
    
    [Example output]
    
    \{\{
    
    "explanation":"explanation to your answer",
    
    "choice": "A"
    
    \}\}\\
    \\
    
    [\{character\_name\}'s Information]
    
    Public Information: \{public\}
    
    Private Information: \{private\}
    
    Social Goal: \{goal\}\\
    \\
    
    [Other Roles' Information]
    
    \{user\_profile\}\\
    \\
    
    [Dialogue Context]
    
    \{dialogue\_context\}\\
    \\
    
    [Question]
    
    \{question\}\\
    \\
    
    [Options]
    
    \{choices\}\\
    \\
\hline
\end{tabular}}
\caption{Evaluation prompt (\texttt{zh} \& \texttt{en}) for interpersonal ability evaluation ($\mathbb{IAE}$) task. \texttt{\{character\_name\}}, \texttt{\{public\}}, \texttt{\{private\}}, \texttt{\{goal\}}, \texttt{\{user\_profile\}}, \texttt{\{dialogue\_context\}}, \texttt{\{question\}} and \texttt{\{choices\}} are placeholders.}
\label{iae_eval}
\end{table*}

\subsection{More LLMs' Results on Two Tasks}
\label{sec:more_llm_results_appendix}

We also evaluate other 9 LLMs: \textbf{(1) Closed-source}: GPT series (4o-mini, o1-mini, \citealt{o1}). \textbf{(2) Open-source}: Baichuan2-Chat series (7B, 13B, \citealt{baichuan2}), Yi-1.5-Chat series (9B, 34B, \citealt{yi}), Phi series (3.5-mini, 3.5-moe, 4, \citealt{phi4}). Their results on the $\mathbb{GAE}$ and $\mathbb{IAE}$ task are shown in Tables \ref{goal_evaluation_appendix} and \ref{ability_evaluation_appendix}, respectively.

\begin{table*}[t]
\centering
\resizebox{\textwidth}{!}{
\begin{tabular}{l c c c c c c c c c c c}
\toprule
    \multirow{3}{*}{\makecell[c]{Models}} 
    & \multicolumn{5}{c}{\makecell[c]{Prosocial}}
    & \multicolumn{2}{c}{\makecell[c]{Proself}} 
    & \multicolumn{3}{c}{\makecell[c]{Antisocial}} 
    & \multirow{1}{*}{\makecell[c]{\textbf{Overall}}} \\
    \cmidrule(lr){2-6} \cmidrule(lr){7-8} \cmidrule(lr){9-11}  \cmidrule(lr){12-12}
    & \texttt{Cooperation} & \texttt{Negotiation} & \texttt{Assistant} & \texttt{Altruism} & \cellcolor{c1}\texttt{Avg.} & \texttt{Competition} & \cellcolor{c2}\texttt{Avg.} & \texttt{Induction} & \texttt{Conflict} & \cellcolor{c3}\texttt{Avg.} & \cellcolor{c4}\texttt{Avg.} \\
    & \texttt{zh}/\texttt{en} & \texttt{zh}/\texttt{en} & \texttt{zh}/\texttt{en} & \texttt{zh}/\texttt{en} & \cellcolor{c1}\texttt{zh}/\texttt{en} & \texttt{zh}/\texttt{en} & \cellcolor{c2}\texttt{zh}/\texttt{en} & \texttt{zh}/\texttt{en} & \texttt{zh}/\texttt{en} & \cellcolor{c3}\texttt{zh}/\texttt{en} & \cellcolor{c4}\texttt{zh}/\texttt{en} \\ 
\midrule
    Human (best) & 100.00/100.00 & 100.00/100.00 & 100.00/100.00 & 100.00/100.00 & \cellcolor{c1}100.00/100.00 & 100.00/100.00 & \cellcolor{c2}100.00/100.00 & 100.00/100.00 & 100.00/100.00 & \cellcolor{c3}100.00/100.00 & \cellcolor{c4}100.00/100.00 \\
    Human (average) & 60.00/60.00 & 60.00/55.00 & 70.00/55.00 & 70.00/70.00 & \cellcolor{c1}64.91/59.86 & 55.00/40.00 & \cellcolor{c2}55.00/40.00 & 65.00/60.00 & 37.50/40.00 & \cellcolor{c3}51.25/50.00 & \cellcolor{c4}61.84/55.16 \\
\midrule
    \multicolumn{12}{c}{\makecell[c]{\textit{Closed-sourced LLMs}}} \\
\midrule
    GPT-4o-mini & 54.16/53.25 & 50.45/47.14 & 47.35/43.65 & 47.41/45.97 & \cellcolor{c1}49.82/47.44 & 24.41/21.97 & \cellcolor{c2}24.41/21.97 & 19.56/11.41 & 23.46/17.48 & \cellcolor{c3}21.41/14.29 & \cellcolor{c4}41.75/38.77 \\
    % Claude-3-sonnet & 55.10/52.41 & 52.14/48.32 & 48.41/45.32 & 48.62/44.65 & \cellcolor{c1}51.06/47.65 & 25.41/24.10 & \cellcolor{c2}25.41/24.10 & 22.42/21.56 & 26.44/23.10 & \cellcolor{c3}24.32/22.29 & \cellcolor{c4}43.16/40.30 \\
    % GLM-4-0520 & 56.14/52.31 & 51.42/46.73 & 47.81/42.63 & 48.24/43.71 & \cellcolor{c1}50.87/46.30 & 28.64/24.18 & \cellcolor{c2}28.64/24.18 & 20.05/16.64 & 25.90/17.14 & \cellcolor{c3}22.82/16.88 & \cellcolor{c4}43.41/38.69 \\
    o1-mini & 55.41/54.87 & 52.41/51.11 & 51.18/50.12 & 50.11/48.16 & \cellcolor{c1}52.25/51.03 & 27.49/24.19 & \cellcolor{c2}27.49/24.19 & 20.10/28.21 & 25.15/20.09 & \cellcolor{c3}22.49/24.36 & \cellcolor{c4}44.13/42.93 \\
    % GPT-4 & 55.29/55.16 & 52.41/51.15 & 49.85/46.41 & 51.45/49.35 & \cellcolor{c1}52.23/50.49 & 29.44/27.12 & \cellcolor{c2}29.44/27.12 & 19.53/15.16 & 26.85/19.74 & \cellcolor{c3}23.19/17.33 & \cellcolor{c4}44.52/42.19 \\
    % GPT-4o & 56.65/55.98 & 53.14/51.56 & 51.16/50.54 & 50.41/48.41 & \cellcolor{c1}52.81/51.58 & 27.56/25.46 & \cellcolor{c2}27.56/25.46 & 21.41/17.46 & 25.16/17.65 & \cellcolor{c3}23.19/17.55 & \cellcolor{c4}44.62/42.69 \\
    % Claude-3-opus & 58.12/57.45 & 53.07/50.49 & 54.12/53.11 & 50.98/48.66 & \cellcolor{c1}54.01/52.33 & 31.49/29.75 & \cellcolor{c2}31.49/29.75 & 27.46/25.31 & 32.10/30.17 & \cellcolor{c3}29.66/27.61 & \cellcolor{c4}46.96/45.23 \\
    % o1 & 57.41/55.65 & 54.12/52.19 & 53.65/52.88 & 51.98/50.10 & \cellcolor{c1}54.26/52.66 & 32.45/30.98 & \cellcolor{c2}32.45/30.98 & 25.41/23.12 & 29.89/28.46 & \cellcolor{c3}27.53/25.65 & \cellcolor{c4}47.04/45.43 \\
\midrule
    \multicolumn{12}{c}{\makecell[c]{\textit{Open-sourced LLMs}}} \\
\midrule
    Baichuan2-7B & 37.44/34.12 & 35.13/36.41 & 34.02/32.62 & 33.43/30.09 & \cellcolor{c1}34.99/33.39 & 19.41/15.41 & \cellcolor{c2}19.41/15.41 & 11.95/10.11 & 14.11/9.69 & \cellcolor{c3}12.97/9.91 & \cellcolor{c4}29.47/27.26 \\
    Yi-1.5-9B & 41.78/38.89 & 37.45/32.68 & 35.47/31.06 & 32.41/29.86 & \cellcolor{c1}36.75/33.06 & 20.74/17.11 & \cellcolor{c2}20.74/17.11 & 13.21/9.75 & 14.49/12.16 & \cellcolor{c3}13.82/10.89 & \cellcolor{c4}31.04/27.45 \\
    Baichuan2-13B & 38.41/35.14 & 34.98/32.46 & 36.74/34.11 & 38.14/34.71 & \cellcolor{c1}37.00/34.05 & 21.87/16.17 & \cellcolor{c2}21.87/16.17 & 13.47/12.18 & 16.74/12.88 & \cellcolor{c3}15.02/12.51 & \cellcolor{c4}31.56/28.18 \\
    % Mistral-7B-v0.3 & 40.79/40.46 & 35.74/29.06 & 38.47/31.24 & 33.74/30.52 & \cellcolor{c1}37.11/32.64 & 22.46/17.37 & \cellcolor{c2}22.46/17.37 & 12.49/8.97 & 15.65/11.57 & \cellcolor{c3}13.99/10.2 & \cellcolor{c4}31.62/27.12 \\
    % LLaMA-3.1-8B & 41.11/39.45 & 34.65/28.44 & 40.57/37.66 & 35.74/30.26 & \cellcolor{c1}37.89/33.74 & 21.16/18.65 & \cellcolor{c2}21.16/18.65 & 11.07/10.41 & 15.16/11.24 & \cellcolor{c3}13.01/10.80 & \cellcolor{c4}31.81/28.20 \\
    Phi-3.5-mini-instruct & 40.56/39.44 & 37.40/34.16 & 38.46/36.41 & 35.47/34.12 & \cellcolor{c1}37.93/35.95 & 23.07/20.11 & \cellcolor{c2}23.07/20.11 & 13.85/10.74 & 14.34/12.24 & \cellcolor{c3}14.08/11.45 & \cellcolor{c4}32.31/30.07 \\
    Yi-1.5-34B & 43.41/39.14 & 41.44/38.43 & 37.41/35.68 & 37.43/35.11 & \cellcolor{c1}39.93/37.11 & 20.46/15.41 & \cellcolor{c2}20.46/15.41 & 15.41/10.99 & 16.14/15.72 & \cellcolor{c3}15.76/13.23 & \cellcolor{c4}33.45/30.27 \\
    % Qwen-2.5-7B & 45.12/42.32 & 37.40/31.70 & 41.09/38.42 & 36.71/32.51 & \cellcolor{c1}39.96/36.05 & 23.06/18.61 & \cellcolor{c2}23.06/18.61 & 14.42/9.41 & 16.55/13.32 & \cellcolor{c3}15.43/11.26 & \cellcolor{c4}33.89/29.85 \\
    Phi-3.5-moe-instruct & 42.97/41.12 & 38.66/34.55 & 40.49/38.13 & 37.96/37.11 & \cellcolor{c1}39.95/37.61 & 24.10/21.20 & \cellcolor{c2}24.10/21.20 & 14.18/10.49 & 14.74/12.85 & \cellcolor{c3}14.45/11.61 & \cellcolor{c4}33.95/31.44 \\
    % Mistral-8*7B-v0.1 & 42.16/43.11 & 41.62/40.10 & 39.46/34.41 & 37.56/34.46 & \cellcolor{c1}40.22/38.03 & 24.16/16.45 & \cellcolor{c2}24.16/16.45 & 15.46/12.32 & 16.41/12.47 & \cellcolor{c3}15.91/12.39 & \cellcolor{c4}34.33/30.99 \\
    % GLM-4-9B & 44.98/43.14 & 38.41/32.10 & 43.56/37.66 & 37.46/34.29 & \cellcolor{c1}40.99/36.60 & 22.64/17.65 & \cellcolor{c2}22.64/17.65 & 15.74/9.22 & 14.65/12.14 & \cellcolor{c3}15.22/10.60 & \cellcolor{c4}34.51/29.99 \\
    Phi-4 & 44.58/42.87 & 39.49/35.79 & 42.17/39.87 & 37.54/36.19 & \cellcolor{c1}40.87/38.56 & 23.97/22.12 & \cellcolor{c2}23.97/22.12 & 16.88/12.11 & 15.94/13.48 & \cellcolor{c3}16.43/12.76 & \cellcolor{c4}34.81/32.41 \\
\bottomrule
\end{tabular}}
\caption{Results of goal achievement evaluation ($\mathbb{GAE}$) task. The score is the average goal achievement ratio (\%).}
\label{goal_evaluation_appendix}
\end{table*}

\begin{table}[t]
\centering
\resizebox{\columnwidth}{!}{
\begin{tabular}{l c c c c c c}
\toprule
    \multirow{2}{*}{\makecell[c]{Models}} 
    & \multicolumn{1}{c}{\makecell[c]{Social \\ Engagement}}
    & \multicolumn{1}{c}{\makecell[c]{Cooperation}} 
    & \multicolumn{1}{c}{\makecell[c]{Self- \\ Management}} 
    & \multicolumn{1}{c}{\makecell[c]{Emotional \\ Resilience}} 
    & \multicolumn{1}{c}{\makecell[c]{Innovation}} 
    & \multicolumn{1}{c}{\makecell[c]{\textbf{Overall}}} \\
    % \cmidrule(lr){2-3} \cmidrule(lr){4-5} \cmidrule(lr){6-7}  \cmidrule(lr){8-9} \cmidrule(lr){10-11} \cmidrule(lr){12-13}
    & \texttt{zh}/\texttt{en} & \texttt{zh}/\texttt{en} & \texttt{zh}/\texttt{en} & \texttt{zh}/\texttt{en} & \texttt{zh}/\texttt{en} & \texttt{zh}/\texttt{en} \\ 
\midrule
    Human (best)  & 84.85/85.71 & 89.56/92.86 & 86.32/80.56 & 81.76/86.67 & 81.48/85.71 & \cellcolor{c4}85.73/\cellcolor{c4}85.32  \\
    Human (average)  & 79.51/82.16 & 82.65/84.57 & 80.46/74.53 & 78.61/79.06 & 76.84/79.91 & \cellcolor{c4}80.22/\cellcolor{c4}79.08  \\
\midrule
    \multicolumn{7}{c}{\makecell[c]{\textit{Closed-sourced LLMs}}} \\
\midrule
    GPT-4o-mini  & 71.57/68.47 & 77.85/71.62 & 69.86/65.17 & 72.80/73.12 & 72.62/68.14 & \cellcolor{c4}72.46/\cellcolor{c4}68.46  \\
    % GLM-4-0520  & 73.49/70.58 & 80.44/73.25 & 72.12/65.55 & 75.74/72.15 & 72.46/67.84 & \cellcolor{c4}74.65/\cellcolor{c4}69.19  \\
    o1-mini  & 73.12/70.98 & 79.12/76.09 & 73.20/70.31 & 75.48/72.14 & 73.12/70.95 & \cellcolor{c4}74.74/\cellcolor{c4}71.95  \\
    % GPT-4o  & 74.39/73.28 & 80.65/76.04 & 72.49/69.12 & 75.63/74.68 & 70.23/67.59 & \cellcolor{c4}74.83/\cellcolor{c4}72.01  \\
    % Claude-3-sonnet  & 74.87/73.68 & 79.86/77.42 & 73.49/71.86 & 74.97/71.85 & 72.79/69.14 & \cellcolor{c4}75.24/\cellcolor{c4}73.16  \\
    % GPT-4  & 73.59/71.68 & 80.56/76.86 & 73.98/70.07 & 77.11/76.33 & 74.11/69.71 & \cellcolor{c4}75.73/\cellcolor{c4}72.64  \\
    % o1  & 76.24/74.67 & 82.44/80.76 & 75.12/73.11 & 78.10/76.57 & 74.99/73.12 & \cellcolor{c4}77.27/\cellcolor{c4}75.49  \\
    % Claude-3-opus  & 77.49/76.48 & 84.46/82.46 & 74.61/72.48 & 76.94/72.37 & 75.16/72.13 & \cellcolor{c4}77.57/\cellcolor{c4}75.27  \\
\midrule
    \multicolumn{7}{c}{\makecell[c]{\textit{Open-sourced LLMs}}} \\
\midrule
    Baichuan2-7B  & 41.67/40.59 & 51.47/50.15 & 50.64/49.54 & 59.76/57.48 & 42.41/41.41 & \cellcolor{c4}49.82/\cellcolor{c4}48.52  \\
    Baichuan2-13B  & 53.28/50.16 & 54.74/52.40 & 52.30/51.10 & 61.32/59.48 & 49.02/45.18 & \cellcolor{c4}54.02/\cellcolor{c4}51.94  \\
    Yi-1.5-9B  & 57.56/54.16 & 61.14/58.14 & 58.46/54.31 & 57.49/54.19 & 45.97/42.65 & \cellcolor{c4}57.81/\cellcolor{c4}54.22  \\
    % Mistral-7B-v0.3  & 59.42/55.23 & 64.15/62.11 & 58.46/52.55 & 55.41/52.46 & 49.88/47.68 & \cellcolor{c4}58.78/\cellcolor{c4}54.68  \\
    % Mistral-8*7B-v0.1  & 62.12/57.25 & 65.16/63.14 & 60.22/58.75 & 57.56/55.41 & 52.16/50.11 & \cellcolor{c4}60.65/\cellcolor{c4}58.29  \\
    % Llama-3.1-8B  & 61.04/59.79 & 62.63/57.61 & 59.47/55.02 & 63.42/59.15 & 56.82/51.72 & \cellcolor{c4}60.78/\cellcolor{c4}56.79  \\
    Yi-1.5-34B  & 61.46/57.49 & 65.34/61.46 & 59.97/57.92 & 61.56/58.54 & 51.11/50.73 & \cellcolor{c4}60.95/\cellcolor{c4}58.15  \\
    Phi-3.5-mini-instruct  & 62.41/60.47 & 65.14/64.19 & 58.52/55.95 & 61.74/57.96 & 57.89/54.74 & \cellcolor{c4}61.04/\cellcolor{c4}58.71  \\
    Phi-3.5-moe-instruct  & 64.85/61.09 & 66.74/65.69 & 59.18/56.47 & 62.41/58.65 & 58.44/55.14 & \cellcolor{c4}62.22/\cellcolor{c4}59.47  \\
    % Mistral-8*22B-v0.1  & 62.08/58.45 & 66.47/64.15 & 63.41/61.74 & 58.45/55.98 & 60.35/57.86 & \cellcolor{c4}62.89/\cellcolor{c4}60.54  \\
    Phi-4  & 65.18/63.45 & 68.43/64.12 & 60.17/57.68 & 64.27/60.48 & 60.11/56.09 & \cellcolor{c4}63.40/\cellcolor{c4}60.38  \\
    % GLM-4-9B  & 64.60/61.42 & 69.48/64.47 & 61.94/59.98 & 63.91/60.71 & 67.05/62.37 & \cellcolor{c4}64.66/\cellcolor{c4}61.46  \\
    % Qwen-2.5-7B  & 64.58/62.10 & 74.47/71.22 & 61.20/58.46 & 65.61/62.12 & 57.14/52.25 & \cellcolor{c4}64.93/\cellcolor{c4}61.87  \\
    % Qwen-2.5-14B  & 67.48/64.16 & 74.85/72.45 & 66.19/63.19 & 67.75/63.71 & 65.32/61.02 & \cellcolor{c4}68.40/\cellcolor{c4}65.22  \\
    % Qwen-2.5-32B  & 67.64/64.36 & 75.12/72.61 & 67.59/63.67 & 69.97/64.32 & 66.79/62.20 & \cellcolor{c4}69.45/\cellcolor{c4}65.65  \\
    % Qwen-2.5-72B  & 68.93/64.55 & 75.69/73.46 & 68.28/64.59 & 71.24/65.17 & 68.95/64.41 & \cellcolor{c4}70.41/\cellcolor{c4}66.51  \\
    % Llama-3.1-70B  & 66.81/63.15 & 77.53/73.56 & 69.25/65.20 & 74.56/69.48 & 67.63/63.45 & \cellcolor{c4}71.15/\cellcolor{c4}67.04  \\
    % Llama-3.3-70B  & 68.11/63.98 & 78.12/74.42 & 69.74/65.44 & 75.41/68.96 & 66.48/64.42 & \cellcolor{c4}71.75/\cellcolor{c4}67.46  \\
    % DeepSeek-V3 & 75.62/74.18 & 81.37/78.49 & 74.93/72.41 & 77.58/76.31 & 74.88/72.89 & \cellcolor{c4}76.51/\cellcolor{c4}73.81 \\
    % DeepSeek-R1 & 77.28/76.34 & 83.52/81.87 & 75.24/73.38 & 78.06/76.62 & 75.19/73.22 & \cellcolor{c4}77.62/\cellcolor{c4}75.44 \\
\bottomrule
\end{tabular}}
\caption{Results (\%) of interpersonal ability evaluation ($\mathbb{IAE}$). The score is average ability selection accuracy.}
\label{ability_evaluation_appendix}
\end{table}

\begin{table*}[t]
\centering
\resizebox{\textwidth}{!}{
\begin{tabular}{l c c c c c c c c c c c c c}
\toprule
    \multirow{2}{*}{\makecell[c]{Models}} 
    & Adapt. & CFConsist. & CFIndep. & DeManage. & DMSkill. & GoalReg. & OrganSkill. & ResManage. & RuleFollow. & SelfReflect. & TaskManage. & TimeManage. & \cellcolor{c4}Overall \\

    & \texttt{zh}/\texttt{en} & \texttt{zh}/\texttt{en} & \texttt{zh}/\texttt{en} & \texttt{zh}/\texttt{en} & \texttt{zh}/\texttt{en} & \texttt{zh}/\texttt{en} & \texttt{zh}/\texttt{en} & \texttt{zh}/\texttt{en} & \texttt{zh}/\texttt{en} & \texttt{zh}/\texttt{en} & \texttt{zh}/\texttt{en} & \texttt{zh}/\texttt{en} & \cellcolor{c4}\texttt{zh}/\texttt{en} \\ 
\midrule
    Human (best) & 90.29/83.56 & 81.32/77.65 & 85.52/78.91 & 85.64/79.34 & 89.85/82.78 & 84.09/77.98 & 92.42/89.93 & 86.87/82.34 & 83.82/77.95 & 82.08/77.95 & 85.04/78.74 & 91.33/85.80 & \cellcolor{c4}86.32/80.56 \\
    Human (average) & 85.54/76.62 & 73.24/69.87 & 80.05/73.60 & 80.80/74.60 & 85.14/76.04 & 76.91/73.51 & 86.49/83.08 & 83.32/75.05 & 74.96/72.68 & 74.30/71.71 & 79.78/73.55 & 86.30/81.12 & \cellcolor{c4}80.46/74.53 \\
\midrule
    \multicolumn{14}{c}{\makecell[c]{\textit{Closed-sourced LLMs}}} \\
\midrule
    GPT-4 & 78.55/76.85 & 67.08/63.43 & 75.17/69.47 & 75.18/70.55 & 77.08/72.16 & 70.78/67.06 & 80.32/76.92 & 76.10/71.99 & 68.58/66.61 & 67.50/65.95 & 74.80/68.57 & 80.20/76.90 & \cellcolor{c4}73.98/70.07 \\
    GPT-4o & 75.73/74.06 & 66.68/64.46 & 71.95/66.39 & 73.43/67.60 & 75.41/73.75 & 70.00/66.24 & 81.20/75.11 & 74.53/69.90 & 69.95/65.90 & 67.45/65.89 & 70.32/66.27 & 78.86/74.87 & \cellcolor{c4}72.49/69.12 \\
    GPT-4o-mini & 75.49/69.67 & 65.47/59.25 & 68.48/64.84 & 70.19/65.02 & 72.21/69.16 & 67.70/62.18 & 77.68/70.48 & 70.27/67.57 & 66.74/61.02 & 66.00/59.55 & 67.90/64.68 & 75.99/69.90 & \cellcolor{c4}69.86/65.17 \\
    o1 & 79.14/76.93 & 66.11/\textbf{66.88} & 76.48/73.24 & 76.51/73.52 & 78.83/\textbf{75.88} & 74.77/72.36 & 79.45/\textbf{78.01} & \textbf{77.29}/73.52 & 72.48/71.40 & 67.33/\textbf{68.18} & 75.16/73.02 & 79.44/77.19 & \cellcolor{c4}75.12/73.11 \\
    o1-mini & 76.99/75.13 & 65.46/64.14 & 72.51/69.96 & 75.09/70.20 & 76.65/74.59 & 70.88/67.91 & 77.41/76.06 & 76.50/70.99 & 69.42/67.09 & 69.26/64.98 & 72.29/68.74 & 77.11/75.36 & \cellcolor{c4}73.20/70.31 \\
    GLM-4 & 75.91/68.52 & 68.78/60.98 & 70.50/67.02 & 73.12/67.02 & 74.79/67.12 & 69.31/64.48 & 78.08/69.81 & 73.76/67.06 & 69.27/62.16 & 69.09/61.13 & 69.59/64.71 & 76.34/69.34 & \cellcolor{c4}72.12/65.55 \\
    Claude-3-sonnet & 77.43/76.20 & 68.21/63.29 & 73.68/72.10 & 73.77/73.56 & 76.49/75.59 & 70.31/69.37 & 81.80/77.11 & 74.30/74.39 & 70.06/67.91 & 68.82/66.77 & 70.62/71.16 & 81.26/76.78 & \cellcolor{c4}73.49/71.86 \\
    Claude-3-opus & 78.22/74.76 & 68.89/66.23 & 74.59/\textbf{73.57} & 75.87/\textbf{73.82} & 77.27/74.75 & 70.91/72.34 & \textbf{82.53}/77.05 & 76.48/74.01 & 70.03/70.24 & 69.42/66.47 & 74.50/72.53 & 82.04/76.95 & \cellcolor{c4}74.61/72.48 \\
\midrule
    \multicolumn{14}{c}{\makecell[c]{\textit{Open-sourced LLMs}}} \\
\midrule
    Qwen-2.5-7B & 65.05/61.52 & 57.05/54.04 & 61.47/57.70 & 62.09/58.34 & 63.09/60.83 & 59.11/57.16 & 66.76/61.86 & 62.17/60.46 & 58.15/56.42 & 57.62/55.27 & 60.34/57.34 & 65.18/61.83 & \cellcolor{c4}61.20/58.46 \\
    Qwen-2.5-14B & 68.29/66.28 & 61.15/60.26 & 66.66/62.55 & 66.72/62.92 & 67.72/65.21 & 65.92/61.24 & 72.20/69.33 & 67.35/63.06 & 63.94/60.66 & 62.36/60.47 & 66.10/61.75 & 70.46/68.45 & \cellcolor{c4}66.19/63.19 \\
    Qwen-2.5-32B & 70.11/68.56 & 61.22/57.72 & 68.64/62.97 & 69.23/63.19 & 70.04/68.54 & 66.15/60.95 & 72.06/69.61 & 69.99/65.82 & 64.79/58.99 & 63.00/57.76 & 67.51/61.79 & 70.44/68.77 & \cellcolor{c4}67.59/63.67 \\
    Qwen2.5-72B & 72.30/67.11 & 64.56/59.86 & 65.96/65.12 & 66.36/65.14 & 72.29/66.58 & 65.73/62.65 & 74.59/69.45 & 70.16/66.22 & 65.34/62.62 & 64.80/60.69 & 65.89/64.82 & 72.98/67.75 & \cellcolor{c4}68.28/64.59 \\
    LLaMA-3.1-8B & 61.94/58.43 & 54.52/51.07 & 60.97/53.89 & 61.05/55.55 & 61.57/58.01 & 56.84/53.26 & 64.38/60.86 & 61.35/56.49 & 56.46/51.77 & 56.28/51.13 & 57.95/53.31 & 62.46/59.10 & \cellcolor{c4}59.47/55.02 \\
    LLaMa-3.1-70B & 74.73/68.62 & 61.69/59.68 & 70.85/65.41 & 71.59/65.54 & 73.56/68.30 & 64.81/64.35 & 75.20/70.68 & 73.25/66.01 & 63.69/61.33 & 61.76/59.94 & 66.15/65.14 & 75.02/69.90 & \cellcolor{c4}69.25/65.20 \\
    LLaMA-3.3-70B & 71.60/70.00 & 64.84/60.77 & 69.89/63.69 & 70.58/66.16 & 71.12/67.96 & 69.04/62.77 & 74.15/73.99 & 70.69/67.58 & 68.42/62.23 & 66.73/61.05 & 69.51/63.04 & 73.82/72.27 & \cellcolor{c4}69.74/65.44 \\
    GLM-4-9B & 65.21/62.99 & 56.06/56.97 & 62.19/60.26 & 63.92/60.41 & 64.97/61.10 & 59.31/57.86 & 67.62/66.22 & 64.45/60.75 & 58.85/57.74 & 57.09/57.28 & 60.48/58.31 & 65.71/64.72 & \cellcolor{c4}61.94/59.98 \\
    Mistral-7B-Instruct-v0.3 & 59.77/56.21 & 56.02/47.53 & 57.86/52.60 & 59.34/53.31 & 59.71/55.84 & 57.02/50.55 & 62.14/57.12 & 59.60/53.36 & 56.89/49.59 & 56.22/47.91 & 57.74/51.39 & 61.77/56.51 & \cellcolor{c4}58.46/52.55 \\
    Mistral-8*7B-Instruct-v0.1 & 63.43/60.40 & 56.95/55.16 & 59.14/59.16 & 59.96/59.66 & 61.51/60.12 & 58.42/58.09 & 67.35/66.65 & 61.40/59.98 & 58.33/56.12 & 58.32/55.37 & 58.68/59.06 & 64.75/61.12 & \cellcolor{c4}60.22/58.75 \\
    Mistral-8*22B-Instruct-v0.1 & 66.36/66.17 & 57.33/56.73 & 63.67/62.16 & 64.66/63.40 & 66.35/64.79 & 62.28/57.78 & 67.45/67.31 & 65.96/64.17 & 60.30/57.45 & 58.44/57.03 & 63.05/59.13 & 66.37/66.96 & \cellcolor{c4}63.41/61.74 \\
    Baichuan2-7B-Chat & 53.24/53.08 & 46.19/46.09 & 51.08/49.83 & 51.34/50.75 & 52.68/52.03 & 48.58/46.76 & 54.45/53.27 & 51.42/50.83 & 48.38/46.32 & 47.85/46.14 & 49.97/47.20 & 54.29/53.10 & \cellcolor{c4}50.64/49.54 \\
    Baichuan2-13B-Chat & 57.32/53.88 & 47.51/47.73 & 50.61/50.15 & 50.89/51.77 & 56.60/53.48 & 49.19/49.37 & 57.75/56.94 & 54.10/52.02 & 48.73/48.44 & 48.52/47.73 & 49.69/49.61 & 57.33/55.43 & \cellcolor{c4}52.30/51.10 \\
    Yi-1.5-9B-Chat & 63.65/57.34 & 54.10/50.24 & 58.36/54.40 & 58.48/54.44 & 60.10/56.37 & 55.72/52.77 & 64.31/58.92 & 59.37/55.27 & 55.38/51.12 & 54.65/50.99 & 58.11/53.90 & 64.03/58.47 & \cellcolor{c4}58.46/54.31 \\
    Yi-1.5-34B-Chat & 62.56/62.47 & 54.55/52.28 & 61.85/56.40 & 61.92/57.21 & 62.46/62.21 & 55.76/55.63 & 65.32/63.08 & 62.06/59.99 & 55.63/53.76 & 55.25/53.49 & 61.43/56.33 & 63.43/62.89 & \cellcolor{c4}59.97/57.92 \\
    Phi-3.5-mini-instruct & 62.53/58.87 & 52.88/49.55 & 57.32/56.92 & 59.24/57.10 & 62.21/58.52 & 55.95/54.83 & 63.48/60.04 & 59.80/57.16 & 55.36/54.13 & 54.88/51.00 & 56.78/55.89 & 63.18/59.38 & \cellcolor{c4}58.52/55.95 \\
    Phi-3.5-moe-instruct & 62.33/59.84 & 53.77/51.27 & 60.39/55.90 & 60.80/56.14 & 62.12/59.47 & 55.71/54.69 & 64.57/60.36 & 61.37/57.76 & 55.68/53.98 & 53.95/53.96 & 59.23/55.10 & 62.47/59.86 & \cellcolor{c4}59.18/56.47 \\
    Phi-4 & 63.68/60.79 & 54.61/52.25 & 59.70/58.76 & 61.58/59.43 & 63.57/59.72 & 57.81/55.22 & 65.27/62.21 & 62.21/59.63 & 56.17/54.65 & 56.02/52.92 & 58.28/57.37 & 64.70/62.04 & \cellcolor{c4}60.17/57.68 \\
    DeepSeek-V3 & 78.82/76.70 & 67.35/64.10 & 75.60/70.25 & 75.95/71.80 & 77.45/73.20 & 71.85/68.45 & 80.75/77.30 & 76.40/72.65 & 69.70/67.25 & 68.15/66.05 & 75.05/69.80 & 80.45/77.10 & \cellcolor{c4}74.93/72.41 \\
    DeepSeek-R1 & \textbf{79.25}/\textbf{77.40} & \textbf{68.95}/66.75 & \textbf{76.55}/73.35 & \textbf{76.65}/73.70 & \textbf{78.95}/75.85 & \textbf{74.85}/\textbf{72.50} & 82.35/77.95 & 76.60/\textbf{74.05} & \textbf{72.55}/\textbf{71.45} & \textbf{69.60}/67.15 & \textbf{75.25}/\textbf{73.55} & \textbf{82.15}/\textbf{77.25} & \cellcolor{c4}\textbf{75.24}/\textbf{73.38} \\
\midrule
\end{tabular}}
\caption{Results (\%) of self-management skill evaluation task. Abbreviations: Adapt. (Adaptability), CFConsist. (Capacity for Consistency), CFIndep. (Capacity for Independence), DeManage. (Detail Management), DMSkill. (Decision-Making Skill), GoalReg. (Goal Regulation), OrganSkill. (Organizational Skill), ResManage. (Responsibility Management), RuleFollow. (Rule-following Skill), SelfReflect. (Self-Reflection Skill), TaskManage. (Task Management), TimeManage. (Time Management).}
\label{self_management_evaluation}
\end{table*}

\begin{table}[t]
\centering
\resizebox{\columnwidth}{!}{
\begin{tabular}{l c c c c c c}
\toprule
    \multirow{2}{*}{\makecell[c]{Models}} 
    & CFSocialWormth. & CFTrust. & EthicalCom. & PersTaking. & TeamSkill. & \cellcolor{c4}Overall \\
    & \texttt{zh}/\texttt{en} & \texttt{zh}/\texttt{en} & \texttt{zh}/\texttt{en} & \texttt{zh}/\texttt{en} & \texttt{zh}/\texttt{en} & \cellcolor{c4}\texttt{zh}/\texttt{en} \\ 
\midrule
    Human (best) & 88.45/90.51 & 86.85/90.47 & 84.00/90.12 & 94.33/99.59 & 92.80/91.40 & \cellcolor{c4}89.56/92.86 \\
    Human (average) & 82.28/84.27 & 78.32/81.51 & 76.33/80.54 & 88.07/89.90 & 86.70/84.91 & \cellcolor{c4}82.65/84.57 \\
\midrule
    \multicolumn{7}{c}{\makecell[c]{\textit{Closed-sourced LLMs}}} \\
\midrule
    GPT-4 & 84.32/75.72 & 72.46/74.47 & 72.01/71.62 & 86.23/81.48 & 86.10/79.67 & \cellcolor{c4}80.56/76.86 \\
    GPT-4o & 81.61/73.61 & 78.79/73.19 & 76.66/71.60 & 82.93/82.26 & 82.58/77.64 & \cellcolor{c4}80.65/76.04 \\
    GPT-4o-mini & 77.05/69.81 & 76.19/68.51 & 75.58/66.53 & 81.04/78.49 & 78.39/72.62 & \cellcolor{c4}77.85/71.62 \\
    o1 & 81.86/81.22 & \textbf{81.83}/80.24 & \textbf{80.01}/\textbf{75.60} & 85.00/84.14 & 82.70/81.53 & \cellcolor{c4}82.44/80.76 \\
    o1-mini & 79.02/77.35 & 75.48/73.62 & 71.34/72.80 & 84.37/78.15 & 83.91/77.91 & \cellcolor{c4}79.12/76.09 \\
    GLM-4 & 79.78/74.59 & 76.92/71.77 & 75.62/68.03 & 85.19/76.27 & 83.29/74.64 & \cellcolor{c4}80.44/73.25 \\
    Claude-3-sonnet & 76.24/79.05 & 74.63/77.18 & 73.96/70.45 & 88.11/80.00 & 83.96/79.65 & \cellcolor{c4}79.86/77.42 \\
    Claude-3-opus & \textbf{84.01}/\textbf{83.36} & 79.96/\textbf{82.33} & 77.69/73.86 & \textbf{87.51}/\textbf{85.94} & \textbf{85.56}/\textbf{85.82} & \cellcolor{c4}\textbf{84.46}/\textbf{82.46} \\
\midrule
    \multicolumn{7}{c}{\makecell[c]{\textit{Open-sourced LLMs}}} \\
\midrule
    Qwen-2.5-7B & 74.97/72.87 & 72.87/69.88 & 69.57/64.74 & 78.55/74.07 & 75.07/73.69 & \cellcolor{c4}74.47/71.22 \\
    Qwen-2.5-14B & 73.46/72.79 & 72.84/70.17 & 72.64/69.95 & 79.56/75.43 & 74.22/72.94 & \cellcolor{c4}74.85/72.45 \\
    Qwen-2.5-32B & 76.18/71.35 & 71.10/68.82 & 69.08/65.21 & 79.12/78.59 & 78.98/77.39 & \cellcolor{c4}75.12/72.61 \\
    Qwen2.5-72B & 77.68/71.99 & 71.31/69.77 & 70.21/68.82 & 80.15/79.85 & 77.69/74.89 & \cellcolor{c4}75.69/73.46 \\
    LLaMA-3.1-8B & 63.79/56.15 & 59.07/55.54 & 57.04/53.21 & 67.36/61.58 & 64.41/60.44 & \cellcolor{c4}62.63/57.61 \\
    LLaMa-3.1-70B & 79.69/74.22 & 70.89/69.45 & 68.67/67.88 & 83.50/77.71 & 83.18/77.35 & \cellcolor{c4}77.53/73.56 \\
    LLaMA-3.3-70B & 77.97/73.53 & 77.85/70.79 & 77.09/70.06 & 78.98/80.33 & 78.44/75.53 & \cellcolor{c4}78.12/74.42 \\
    GLM-4-9B & 70.55/64.57 & 67.28/62.00 & 64.17/61.87 & 73.16/67.49 & 71.10/65.47 & \cellcolor{c4}69.48/64.47 \\
    Mistral-7B-Instruct-v0.3 & 63.43/61.87 & 61.64/59.57 & 61.42/57.49 & 67.72/65.51 & 65.46/65.16 & \cellcolor{c4}64.15/62.11 \\
    Mistral-8*7B-Instruct-v0.1 & 64.28/63.22 & 64.17/62.22 & 62.66/62.11 & 67.67/64.28 & 66.29/63.52 & \cellcolor{c4}65.16/63.14 \\
    Mistral-8*22B-Instruct-v0.1 & 67.12/64.12 & 66.10/61.00 & 60.66/58.84 & 68.90/69.08 & 68.88/66.20 & \cellcolor{c4}66.47/64.15 \\
    Baichuan2-7B-Chat & 51.76/50.36 & 49.03/47.25 & 48.50/46.81 & 53.95/53.83 & 53.38/51.36 & \cellcolor{c4}51.47/50.15 \\
    Baichuan2-13B-Chat & 51.41/52.79 & 50.92/50.60 & 49.64/50.34 & 60.34/54.54 & 59.90/53.05 & \cellcolor{c4}54.74/52.40 \\
    Yi-1.5-9B-Chat & 63.37/56.09 & 59.84/55.95 & 53.86/55.13 & 64.03/62.92 & 63.75/59.15 & \cellcolor{c4}61.14/58.14 \\
    Yi-1.5-34B-Chat & 66.66/62.30 & 60.03/59.41 & 59.89/54.30 & 70.32/65.31 & 68.29/64.87 & \cellcolor{c4}65.34/61.46 \\
    Phi-3.5-mini-instruct & 67.42/63.91 & 61.54/62.76 & 59.48/60.79 & 68.47/66.68 & 67.78/66.10 & \cellcolor{c4}65.14/64.19 \\
    Phi-3.5-moe-instruct & 64.31/64.77 & 64.27/63.83 & 62.28/59.40 & 71.72/69.93 & 69.71/69.35 & \cellcolor{c4}66.74/65.69 \\
    Phi-4 & 69.90/66.17 & 63.77/58.73 & 61.99/57.97 & 74.09/68.84 & 70.62/67.46 & \cellcolor{c4}68.43/64.12 \\
    DeepSeek-V3 & 82.15/75.81 & 79.20/74.53 & 77.34/72.16 & 83.90/81.40 & 83.12/77.28 & \cellcolor{c4}81.37/78.49 \\
    DeepSeek-R1 & 83.72/81.18 & 81.45/80.05 & 78.80/75.27 & 85.52/84.32 & 84.35/83.23 & \cellcolor{c4}83.52/81.87 \\
\midrule
\end{tabular}}
\caption{Results (\%) of cooperation evaluation task. Abbreviations: CFSocialWormth. (Capacity for Social Warmth), CFTrust. (Capacity for Trust), EthicalCom. (Ethical Competence), PersTaking. (Perspective-taking Skill), TeamSkill. (Teamwork Skill).}
\label{cooperation_evaluation}
\end{table}

\begin{table}[t]
\centering
\resizebox{\columnwidth}{!}{
\begin{tabular}{l c c c c c c}
\toprule
    \multirow{2}{*}{\makecell[c]{Models}} 
    & AngerManage. & ConfidenceReg. & CFOptimism. & ImpulseReg. & StressReg. & \cellcolor{c4}Overall \\
    & \texttt{zh}/\texttt{en} & \texttt{zh}/\texttt{en} & \texttt{zh}/\texttt{en} & \texttt{zh}/\texttt{en} & \texttt{zh}/\texttt{en} & \cellcolor{c4}\texttt{zh}/\texttt{en} \\ 
\midrule
    Human (best) & 79.43/88.94 & 76.57/77.14 & 75.87/76.58 & 87.84/93.39 & 83.84/92.04 & \cellcolor{c4}81.76/86.67 \\
    Human (average) & 78.96/74.55 & 73.61/74.43 & 71.29/73.08 & 83.16/85.75 & 81.80/81.14 & \cellcolor{c4}78.61/79.06 \\
\midrule
    \multicolumn{7}{c}{\makecell[c]{\textit{Closed-sourced LLMs}}} \\
\midrule
    GPT-4 & 76.22/73.99 & 76.20/69.59 & 74.05/69.30 & 79.17/81.98 & 77.68/81.27 & \cellcolor{c4}77.11/76.33 \\
    GPT-4o & 71.19/71.93 & 70.91/70.55 & 68.21/69.90 & 81.12/80.33 & 80.30/75.84 & \cellcolor{c4}75.63/74.68 \\
    GPT-4o-mini & 72.95/72.28 & 69.21/70.42 & 69.13/67.43 & 76.06/77.68 & 74.18/73.70 & \cellcolor{c4}72.80/73.12 \\
    o1 & \textbf{81.71}/\textbf{76.64} & 69.68/69.64 & 69.33/68.75 & 83.34/\textbf{82.50} & \textbf{82.70}/\textbf{80.32} & \cellcolor{c4}\textbf{78.10}/76.57 \\
    o1-mini & 76.80/73.86 & 70.27/68.30 & 69.24/66.69 & 79.97/74.92 & 77.70/74.56 & \cellcolor{c4}75.48/72.14 \\
    GLM-4 & 73.68/67.45 & 73.55/67.29 & 69.22/66.17 & 81.07/78.66 & 76.02/74.83 & \cellcolor{c4}75.74/72.15 \\
    Claude-3-sonnet & 75.31/71.09 & 71.90/71.00 & 66.56/69.07 & 80.21/74.40 & 75.94/71.43 & \cellcolor{c4}74.97/71.85 \\
    Claude-3-opus & 73.84/71.56 & 72.76/69.54 & 72.51/66.10 & \textbf{84.42}/77.54 & 75.64/72.58 & \cellcolor{c4}76.94/72.37 \\
\midrule
    \multicolumn{7}{c}{\makecell[c]{\textit{Open-sourced LLMs}}} \\
\midrule
    Qwen-2.5-7B & 64.06/61.58 & 61.82/59.57 & 60.32/58.19 & 69.42/65.02 & 68.51/63.52 & \cellcolor{c4}65.61/62.12 \\
    Qwen-2.5-14B & 67.51/63.39 & 66.55/60.73 & 61.00/60.42 & 70.68/66.25 & 69.32/65.51 & \cellcolor{c4}67.75/63.71 \\
    Qwen-2.5-32B & 71.13/61.66 & 65.67/61.34 & 62.38/59.27 & 73.54/68.10 & 73.46/66.97 & \cellcolor{c4}69.97/64.32 \\
    Qwen2.5-72B & 73.76/67.18 & 65.61/61.80 & 63.06/59.55 & 75.71/67.76 & 74.37/67.30 & \cellcolor{c4}71.24/65.17 \\
    LLaMA-3.1-8B & 61.22/57.67 & 60.37/57.62 & 58.24/54.13 & 67.14/63.31 & 66.01/59.06 & \cellcolor{c4}63.42/59.15 \\
    LLaMa-3.1-70B & 74.38/67.15 & 71.64/66.47 & 71.06/66.17 & 77.31/74.38 & 76.05/69.35 & \cellcolor{c4}74.56/69.48 \\
    LLaMA-3.3-70B & 73.26/70.38 & 71.59/67.00 & 71.23/62.81 & 81.27/71.27 & 75.22/70.74 & \cellcolor{c4}75.41/68.96 \\
    GLM-4-9B & 62.49/59.42 & 62.40/58.08 & 59.65/54.90 & 67.51/63.91 & 64.05/63.43 & \cellcolor{c4}63.91/60.71 \\
    Mistral-7B-Instruct-v0.3 & 55.31/53.78 & 52.47/49.17 & 49.10/48.94 & 58.63/54.88 & 57.98/53.86 & \cellcolor{c4}55.41/52.46 \\
    Mistral-8*7B-Instruct-v0.1 & 57.86/55.89 & 55.27/53.08 & 53.58/50.50 & 60.32/58.68 & 58.41/56.06 & \cellcolor{c4}57.56/55.41 \\
    Mistral-8*22B-Instruct-v0.1 & 57.33/56.61 & 55.33/52.27 & 54.72/51.15 & 61.68/59.08 & 60.22/58.12 & \cellcolor{c4}58.45/55.98 \\
    Baichuan2-7B-Chat & 58.92/58.53 & 58.55/53.64 & 55.18/53.24 & 63.00/60.31 & 59.93/59.50 & \cellcolor{c4}59.76/57.48 \\
    Baichuan2-13B-Chat & 58.23/60.38 & 58.21/59.13 & 56.37/51.25 & 66.88/61.75 & 61.88/61.29 & \cellcolor{c4}61.32/59.48 \\
    Yi-1.5-9B-Chat & 56.46/53.50 & 55.91/52.46 & 55.51/49.98 & 60.44/57.62 & 56.94/54.31 & \cellcolor{c4}57.49/54.19 \\
    Yi-1.5-34B-Chat & 60.49/56.44 & 58.52/55.19 & 56.01/53.50 & 66.35/62.17 & 62.22/61.42 & \cellcolor{c4}61.56/58.54 \\
    Phi-3.5-mini-instruct & 58.14/58.58 & 56.18/56.46 & 55.48/54.35 & 67.08/59.62 & 66.22/59.08 & \cellcolor{c4}61.74/57.96 \\
    Phi-3.5-moe-instruct & 62.00/56.88 & 60.10/55.81 & 58.25/54.21 & 65.55/62.05 & 63.30/60.76 & \cellcolor{c4}62.41/58.65 \\
    Phi-4 & 60.18/56.82 & 57.79/55.73 & 56.07/53.62 & 66.81/61.93 & 64.39/60.62 & \cellcolor{c4}64.27/60.48 \\
    DeepSeek-V3 & 78.15/73.02 & 76.44/70.76 & 74.50/70.12 & 81.18/80.23 & 79.83/77.56 & \cellcolor{c4}77.58/76.31 \\
    DeepSeek-R1 & 78.34/74.22 & \textbf{76.10}/\textbf{73.55} & \textbf{72.98}/\textbf{70.23} & 81.12/79.68 & 80.27/78.12 & \cellcolor{c4}78.06/\textbf{76.62} \\
\midrule
\end{tabular}}
\caption{Results (\%) of emotional resilience evaluation task. Abbreviations: AngerManage. (Anger Management), ConfidenceReg. (Confidence Regulation), CFOptimism. (Capacity for Optimism), ImpulseReg. (Impulse Regulation), StressReg. (Stress Regulation).}
\label{emotional_resilience_evaluation}
\end{table}

\begin{table}[t]
\centering
\resizebox{\columnwidth}{!}{
\begin{tabular}{l c c c c c c}
\toprule
    \multirow{2}{*}{\makecell[c]{Models}} 
    & ConverSkill. & EnergyReg. & ExpreSkill. & LeaderSkill. & PersuaSkill. & \cellcolor{c4}Overall \\
    & \texttt{zh}/\texttt{en} & \texttt{zh}/\texttt{en} & \texttt{zh}/\texttt{en} & \texttt{zh}/\texttt{en} & \texttt{zh}/\texttt{en} & \cellcolor{c4}\texttt{zh}/\texttt{en} \\ 
\midrule
    Human (best) & 88.41/87.95 & 85.92/82.53 & 76.99/81.70 & 90.52/92.62 & 89.95/88.47 & \cellcolor{c4}84.85/85.71 \\
    Human (average) & 79.70/82.42 & 77.93/79.20 & 75.51/75.32 & 84.78/90.85 & 83.00/88.42 & \cellcolor{c4}79.51/82.16 \\

\midrule
    \multicolumn{7}{c}{\makecell[c]{\textit{Closed-sourced LLMs}}} \\
\midrule
    GPT-4 & 72.89/71.44 & 71.04/71.28 & 70.54/68.64 & 78.02/74.98 & 77.18/74.37 & \cellcolor{c4}73.59/71.68 \\
    GPT-4o & 77.46/74.49 & 68.65/70.55 & 68.51/68.68 & 82.77/79.16 & 79.80/77.40 & \cellcolor{c4}74.39/73.28 \\
    GPT-4o-mini & 72.86/68.45 & 72.02/65.93 & 67.07/64.93 & 77.88/76.34 & 73.70/70.97 & \cellcolor{c4}71.57/68.47 \\
    o1 & 77.09/75.12 & 73.64/74.50 & \textbf{72.21}/70.60 & 80.52/78.17 & 80.40/78.09 & \cellcolor{c4}76.24/74.67 \\
    o1-mini & 73.80/72.02 & 69.46/70.68 & 68.45/67.06 & 81.30/76.43 & 77.14/73.25 & \cellcolor{c4}73.12/70.98 \\
    GLM-4 & 75.26/70.37 & 73.56/68.22 & 67.66/67.53 & 80.73/74.82 & 76.83/73.92 & \cellcolor{c4}73.49/70.58 \\
    Claude-3-sonnet & 75.72/73.07 & 74.45/72.74 & 70.39/69.97 & 81.26/79.53 & 77.59/76.62 & \cellcolor{c4}74.87/73.68 \\
    Claude-3-opus & 77.49/75.50 & 73.44/72.50 & 71.27/71.28 & \textbf{86.32}/\textbf{84.64} & \textbf{83.65}/\textbf{82.07} & \cellcolor{c4}\textbf{77.49}/\textbf{76.48} \\
\midrule
    \multicolumn{7}{c}{\makecell[c]{\textit{Open-sourced LLMs}}} \\
\midrule
    Qwen-2.5-7B & 64.22/62.89 & 63.84/61.70 & 63.57/58.57 & 66.49/65.54 & 65.60/64.87 & \cellcolor{c4}64.58/62.10 \\
    Qwen-2.5-14B & 68.95/65.08 & 67.84/64.14 & 63.76/60.80 & 70.72/68.10 & 69.85/66.25 & \cellcolor{c4}67.48/64.16 \\
    Qwen-2.5-32B & 68.25/68.02 & 67.87/60.61 & 65.20/60.07 & 69.97/68.94 & 69.27/68.05 & \cellcolor{c4}67.64/64.36 \\
    Qwen2.5-72B & 69.35/65.86 & 67.07/65.46 & 67.06/61.50 & 73.97/66.65 & 69.98/66.34 & \cellcolor{c4}68.93/64.55 \\
    LLaMA-3.1-8B & 61.61/60.71 & 60.44/57.80 & 57.37/57.52 & 66.80/64.42 & 63.25/61.32 & \cellcolor{c4}61.04/59.79 \\
    LLaMa-3.1-70B & 67.52/64.84 & 65.49/59.96 & 64.17/59.52 & 70.46/67.48 & 68.93/66.73 & \cellcolor{c4}66.81/63.15 \\
    LLaMA-3.3-70B & 70.29/66.32 & 67.34/62.95 & 64.44/59.10 & 71.46/70.15 & 70.64/66.92 & \cellcolor{c4}68.11/63.98 \\
    GLM-4-9B & 67.13/62.42 & 65.41/59.89 & 58.05/56.40 & 69.95/68.11 & 68.85/65.19 & \cellcolor{c4}64.60/61.42 \\
    Mistral-7B-Instruct-v0.3 & 59.21/53.32 & 58.49/53.02 & 55.81/53.01 & 65.17/59.31 & 62.09/58.32 & \cellcolor{c4}59.42/55.23 \\
    Mistral-8*7B-Instruct-v0.1 & 62.35/60.21 & 60.97/56.10 & 60.75/52.05 & 64.77/62.22 & 63.22/60.84 & \cellcolor{c4}62.12/57.25 \\
    Mistral-8*22B-Instruct-v0.1 & 61.71/58.38 & 60.10/58.36 & 59.99/54.22 & 67.64/64.12 & 63.61/61.40 & \cellcolor{c4}62.08/58.45 \\
    Baichuan2-7B-Chat & 42.94/40.89 & 40.05/40.40 & 39.65/39.05 & 44.21/43.70 & 43.38/41.18 & \cellcolor{c4}41.67/40.59 \\
    Baichuan2-13B-Chat & 52.95/50.93 & 52.22/50.28 & 51.09/47.39 & 58.02/55.24 & 54.75/51.08 & \cellcolor{c4}53.28/50.16 \\
    Yi-1.5-9B-Chat & 58.03/58.54 & 55.75/50.46 & 54.88/48.71 & 60.97/59.05 & 60.19/58.78 & \cellcolor{c4}57.56/54.16 \\
    Yi-1.5-34B-Chat & 60.18/58.36 & 59.52/56.06 & 58.02/54.89 & 67.02/63.40 & 65.01/58.65 & \cellcolor{c4}61.46/57.49 \\
    Phi-3.5-mini-instruct & 62.61/59.65 & 59.37/58.40 & 57.14/58.28 & 68.21/63.99 & 67.99/63.18 & \cellcolor{c4}62.41/60.47 \\
    Phi-3.5-moe-instruct & 67.17/61.71 & 65.38/58.95 & 59.41/58.71 & 70.73/65.89 & 67.78/62.90 & \cellcolor{c4}64.85/61.09 \\
    Phi-4 & 64.05/63.91 & 64.02/62.71 & 62.20/61.11 & 72.25/66.10 & 67.10/65.40 & \cellcolor{c4}65.18/63.45 \\
    DeepSeek-V3 & 74.92/72.17 & 75.35/71.05 & 71.12/69.78 & 80.48/77.62 & 78.29/75.50 & \cellcolor{c4}75.62/74.18 \\
    DeepSeek-R1 & \textbf{78.11}/\textbf{77.92} & \textbf{76.02}/\textbf{74.80} & 72.19/\textbf{72.54} & 82.94/80.68 & 80.70/78.91 & \cellcolor{c4}77.28/76.34 \\
\midrule
\end{tabular}}
\caption{Results (\%) of social engagement evaluation task. Abbreviations: ConverSkill. (Conversational Skill), EnergyReg. (Energy Regulation), ExpreSkill. (Expressive Skill), LeaderSkill. (Leadership Skill), PersuaSkill. (Persuasive Skill)}
\label{social_engagement_evaluation}
\end{table}

\begin{table}[t]
\centering
\resizebox{\columnwidth}{!}{
\begin{tabular}{l c c c c c c}
\toprule
    \multirow{2}{*}{\makecell[c]{Models}} 
    & AbsThinkSkill. & ArtSkill. & CreateSkill. & CulturalCom. & InfoProcessSkill. & \cellcolor{c4}Overall \\
    & \texttt{zh}/\texttt{en} & \texttt{zh}/\texttt{en} & \texttt{zh}/\texttt{en} & \texttt{zh}/\texttt{en} & \texttt{zh}/\texttt{en} & \cellcolor{c4}\texttt{zh}/\texttt{en} \\ 
\midrule
    Human (best) & 79.72/88.29 & 79.63/78.18 & 77.20/76.72 & 84.10/90.61 & 84.02/90.01 & \cellcolor{c4}81.48/85.71 \\
    Human (average) & 77.50/81.31 & 74.23/75.20 & 71.61/73.72 & 83.49/85.20 & 78.84/82.67 & \cellcolor{c4}76.84/79.91 \\
\midrule
    \multicolumn{7}{c}{\makecell[c]{\textit{Closed-sourced LLMs}}} \\
\midrule
    GPT-4 & 73.28/70.43 & 70.82/64.76 & 69.61/64.03 & 80.36/73.19 & 76.31/72.65 & \cellcolor{c4}74.11/69.71 \\
    GPT-4o & 69.15/67.43 & 65.67/60.79 & 64.19/60.62 & 77.29/72.37 & 73.35/71.54 & \cellcolor{c4}70.23/67.59 \\
    GPT-4o-mini & 73.06/68.16 & 68.95/61.74 & 67.57/60.87 & 76.55/72.87 & 75.08/72.12 & \cellcolor{c4}72.62/68.14 \\
    o1 & 74.85/\textbf{75.09} & 72.41/\textbf{73.24} & 70.53/65.87 & \textbf{82.75}/76.34 & 76.70/75.73 & \cellcolor{c4}74.99/73.12 \\
    o1-mini & 73.03/71.44 & 70.52/67.05 & 69.72/64.88 & 77.71/74.79 & 74.67/73.94 & \cellcolor{c4}73.12/70.95 \\
    GLM-4 & 72.05/66.41 & 70.17/63.85 & 69.88/62.51 & 80.64/71.19 & 73.23/71.05 & \cellcolor{c4}72.46/67.84 \\
    Claude-3-sonnet & 72.44/68.99 & 68.75/65.05 & 67.97/63.33 & 75.97/75.18 & 75.49/71.95 & \cellcolor{c4}72.79/69.14 \\
    Claude-3-opus & 76.74/68.23 & 72.44/67.15 & 71.43/65.91 & 77.17/\textbf{78.24} & 76.75/76.27 & \cellcolor{c4}75.16/72.13 \\
\midrule
    \multicolumn{7}{c}{\makecell[c]{\textit{Open-sourced LLMs}}} \\
\midrule
    Qwen-2.5-7B & 57.81/53.60 & 56.16/49.63 & 52.63/47.57 & 60.92/55.19 & 58.87/54.24 & \cellcolor{c4}57.14/52.25 \\
    Qwen-2.5-14B & 66.51/59.70 & 60.94/59.50 & 59.78/59.30 & 69.11/66.11 & 67.94/61.85 & \cellcolor{c4}65.32/61.02 \\
    Qwen-2.5-32B & 64.53/62.25 & 64.45/61.63 & 63.26/57.80 & 71.60/64.72 & 68.92/64.12 & \cellcolor{c4}66.79/62.20 \\
    Qwen2.5-72B & 71.44/60.76 & 69.96/58.89 & 60.87/58.71 & 73.25/70.04 & 71.55/68.37 & \cellcolor{c4}68.95/64.41 \\
    LLaMA-3.1-8B & 57.56/53.80 & 56.46/48.05 & 54.24/46.78 & 57.86/54.01 & 57.81/53.88 & \cellcolor{c4}56.82/51.72 \\
    LLaMa-3.1-70B & 68.25/64.80 & 65.87/63.02 & 62.31/56.68 & 72.00/67.73 & 69.82/65.95 & \cellcolor{c4}67.63/63.45 \\
    LLaMA-3.3-70B & 65.91/63.17 & 64.22/62.96 & 61.57/60.24 & 70.75/67.62 & 68.88/66.65 & \cellcolor{c4}66.48/64.42 \\
    GLM-4-9B & 67.21/61.05 & 66.08/58.65 & 60.87/55.51 & 70.83/67.10 & 69.72/66.10 & \cellcolor{c4}67.05/62.37 \\
    Mistral-7B-Instruct-v0.3 & 47.34/47.30 & 45.56/46.42 & 44.29/45.61 & 53.82/52.68 & 53.50/48.40 & \cellcolor{c4}49.88/47.68 \\
    Mistral-8*7B-Instruct-v0.1 & 50.74/50.81 & 49.76/50.44 & 49.58/47.05 & 56.70/53.51 & 53.65/50.96 & \cellcolor{c4}52.16/50.11 \\
    Mistral-8*22B-Instruct-v0.1 & 58.15/58.55 & 56.05/57.47 & 55.37/54.55 & 66.75/60.72 & 63.29/59.01 & \cellcolor{c4}60.35/57.86 \\
    Baichuan2-7B-Chat & 43.27/42.24 & 41.84/38.81 & 38.58/38.12 & 44.81/43.02 & 43.85/43.01 & \cellcolor{c4}42.41/41.41 \\
    Baichuan2-13B-Chat & 47.69/45.20 & 46.98/44.45 & 45.87/43.40 & 52.74/48.38 & 50.80/45.78 & \cellcolor{c4}49.02/45.18 \\
    Yi-1.5-9B-Chat & 44.56/42.01 & 44.27/39.89 & 44.00/38.60 & 49.20/45.02 & 47.20/44.95 & \cellcolor{c4}45.97/42.65 \\
    Yi-1.5-34B-Chat & 52.27/48.78 & 48.91/48.73 & 46.38/46.89 & 53.85/53.28 & 53.13/53.14 & \cellcolor{c4}51.11/50.73 \\
    Phi-3.5-mini-instruct & 59.48/55.41 & 55.66/53.66 & 51.70/51.52 & 61.85/56.44 & 60.37/56.10 & \cellcolor{c4}57.89/54.74 \\
    Phi-3.5-moe-instruct & 58.03/54.17 & 56.90/53.54 & 56.55/49.05 & 63.52/58.60 & 59.12/58.21 & \cellcolor{c4}58.44/55.14 \\
    Phi-4 & 61.23/57.12 & 57.16/55.56 & 55.32/51.71 & 62.41/60.74 & 62.34/57.49 & \cellcolor{c4}60.11/56.09 \\
    DeepSeek-V3 & 74.52/71.78 & 72.40/65.99 & 70.45/64.90 & 79.51/75.82 & 77.12/73.94 & \cellcolor{c4}74.88/72.89 \\
    DeepSeek-R1 & \textbf{77.56}/74.42 & \textbf{75.28}/72.51 & \textbf{71.88}/\textbf{69.94} & 81.09/77.82 & \textbf{79.98}/\textbf{77.11} & \cellcolor{c4}\textbf{75.19}/\textbf{73.22} \\

\midrule
\end{tabular}}
\caption{Results (\%) of innovation evaluation task. Abbreviations: AbsThinkSkill. (Abstract Thinking Skill), ArtSkill. (Artistic Skill), CreateSkill. (Creative Skill), CulturalCom. (Cultural Competence), InfoProcessSkill. (Information Processing Skill).}
\label{innovation_evaluation}
\end{table}

\subsection{LLMs' Performance on 32 Interpersonal Abilities}
\label{sec:32_ability_results}

Detailed results of all 32 interpersonal ability in the $\mathbb{IAE}$ tasks are shown in Tables \ref{self_management_evaluation}, \ref{cooperation_evaluation}, \ref{emotional_resilience_evaluation}, \ref{social_engagement_evaluation} and \ref{innovation_evaluation}, categorized into five tables according to the five ability aspects.

% \subsection{More LLMs' performance on Two Tasks}

% \label{sec:more_llms_results}

\begin{figure*}[t]
    \centering
    \includegraphics[width=.8\textwidth]{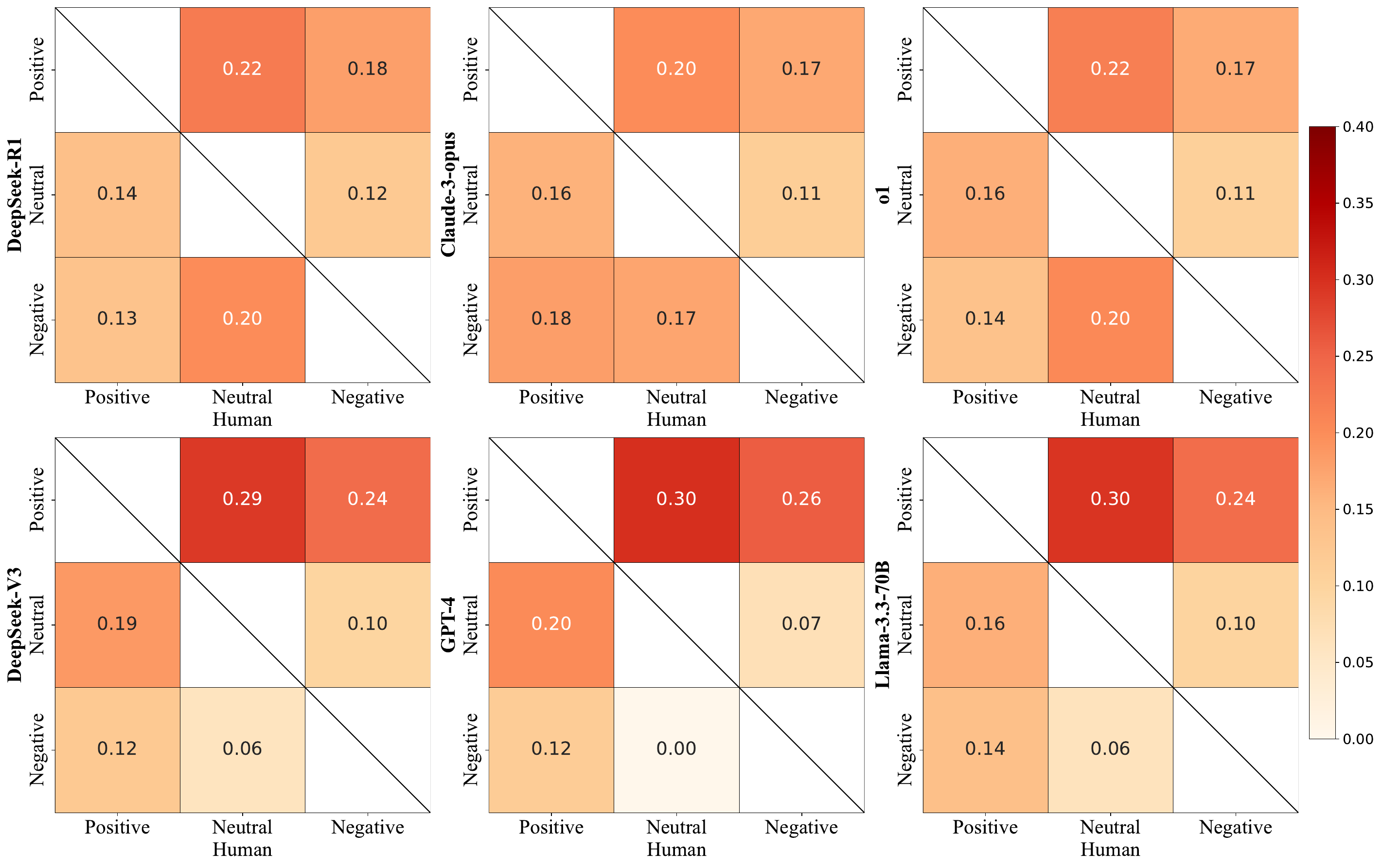}
    \caption{Distribution of behavior combinations shown by LLMs and humans at the same episode transition.}
    \label{compared_behavior_distributions_2}
\end{figure*}

\begin{figure}[t]
    \centering
    \includegraphics[width=\columnwidth]{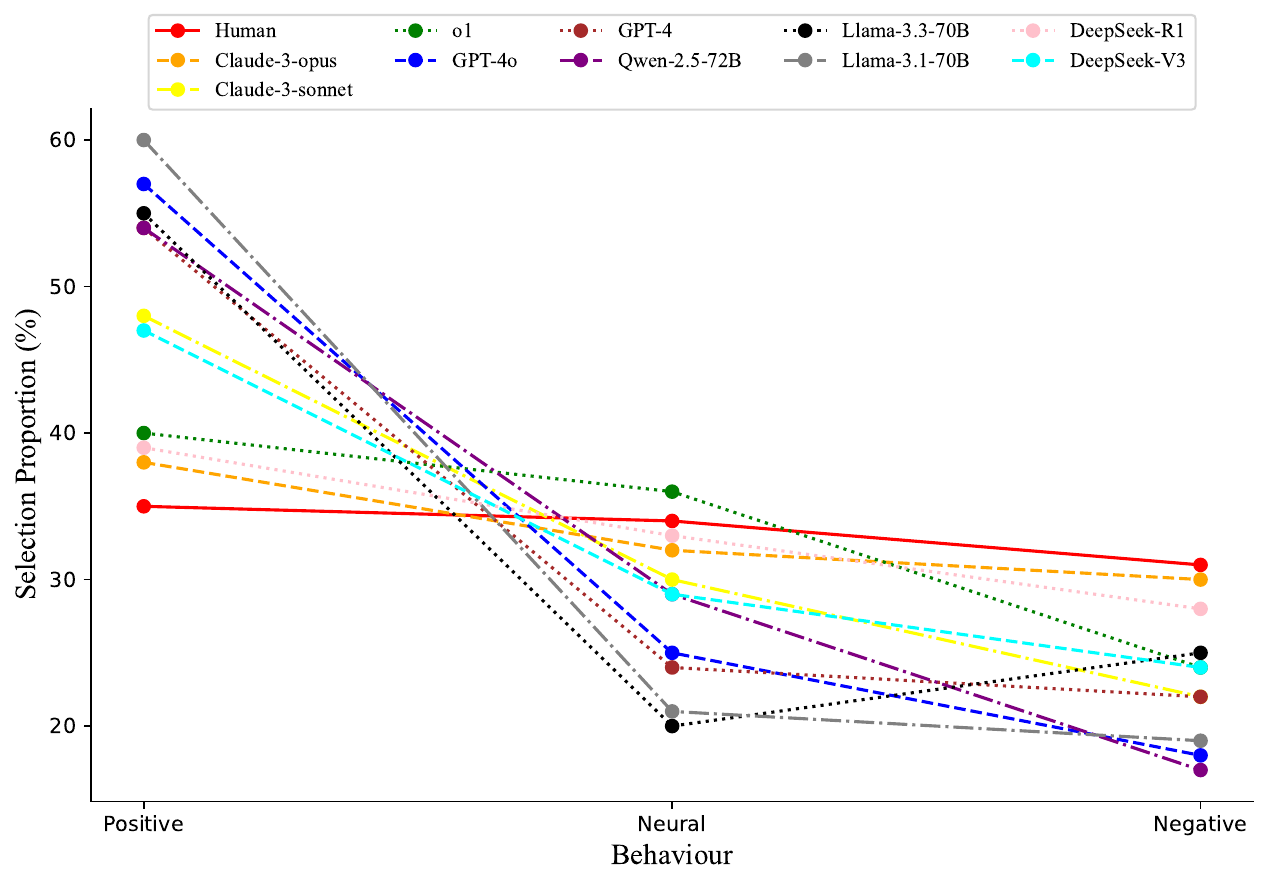}
    \caption{Overall behavioral distribution of LLMs and humans in the $\mathbb{GAE}$ task.}
    \label{overall_behavior_distribution}
\end{figure}

\subsection{More Analysis for LLMs' SI}
\label{sec:more_analysis}

\paragraph{Behavior-level Analysis of LLMs' SI}

Distributions of different behavior combinations exhibited by more LLMs and humans at the same episode transition are presented in Figure \ref{compared_behavior_distributions_2}. 
We also statistic the behavioral distribution of LLMs and humans from all world trees in $\mathbb{GAE}$ task. As illustrated in Figure \ref{overall_behavior_distribution}, the results further demonstrate that LLMs are more likely to select more positive behaviors.

\begin{figure}[t]
    \centering
    \includegraphics[width=\columnwidth]{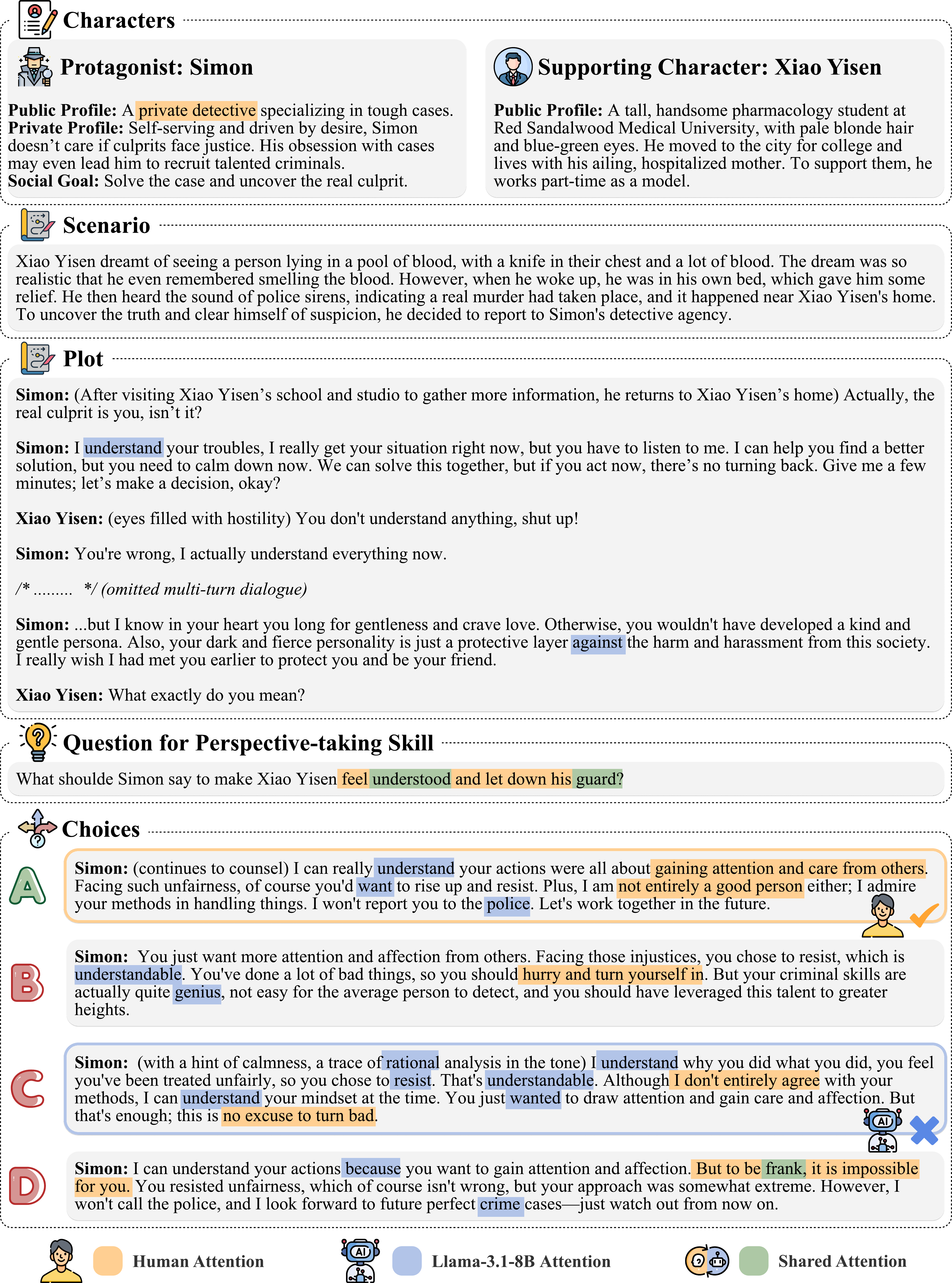}
    \caption{The difference between the human and LLM’s attentions.}
    \label{attention_appendix}
    % \vspace{-2mm}
\end{figure}

\paragraph{The attention of LLMs and humans plays a key role in shaping their social behaviors.}

To delve deeper into what influences the social behaviors of LLMs and humans, we visualize their attention distributions when deciding interpersonal abilities in $\mathbb{IAE}$ task. Treating generating an option as a social behavior, we use samples where LLMs (i.e., Llama-3.1-8B) produce incorrect behavior to dissect the differences between LLMs and humans. Specifically, when generating an option, we average the attention scores across the attention heads in 20 layers, identifying the top 30 keywords the LLM focuses on when answering the question. We also ask two annotators to provide the top 30 keywords they attend to, and take the intersection of their keywords for visualization. The keywords attended to by LLMs and humans are shown in Figure \ref{attention_appendix}. The results show that LLMs focus more on options that are semantically aligned with the question, even if they are incorrect. In contrast, humans focus more on the parts of the incorrect option that violate the question, thus excluding the wrong answers. This highlights the critical role of attention in shaping social behaviors and emphasizes the differences in SI between LLMs and humans.

\begin{figure*}[t]
    \centering
    \includegraphics[width=.95\textwidth]{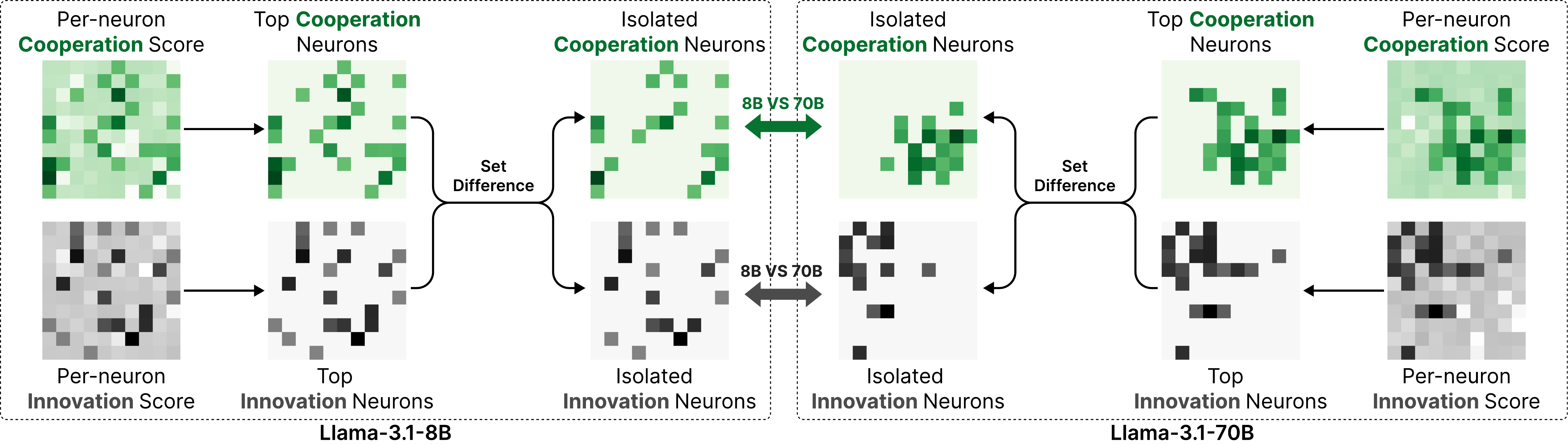}
    \caption{Activated neurons of interpersonal abilities (Cooperation and Innovation) in Llama-3.1-8B \& 70B. We identify the top neurons for each ability by computing per-neuron importance scores, then isolate cooperation-critical neurons from innovation neurons using set differences, and vice versa.}
    \label{neuron_importance_cooperation_innovation}
\end{figure*}

\begin{figure*}[t]
    \centering
    \includegraphics[width=.95\textwidth]{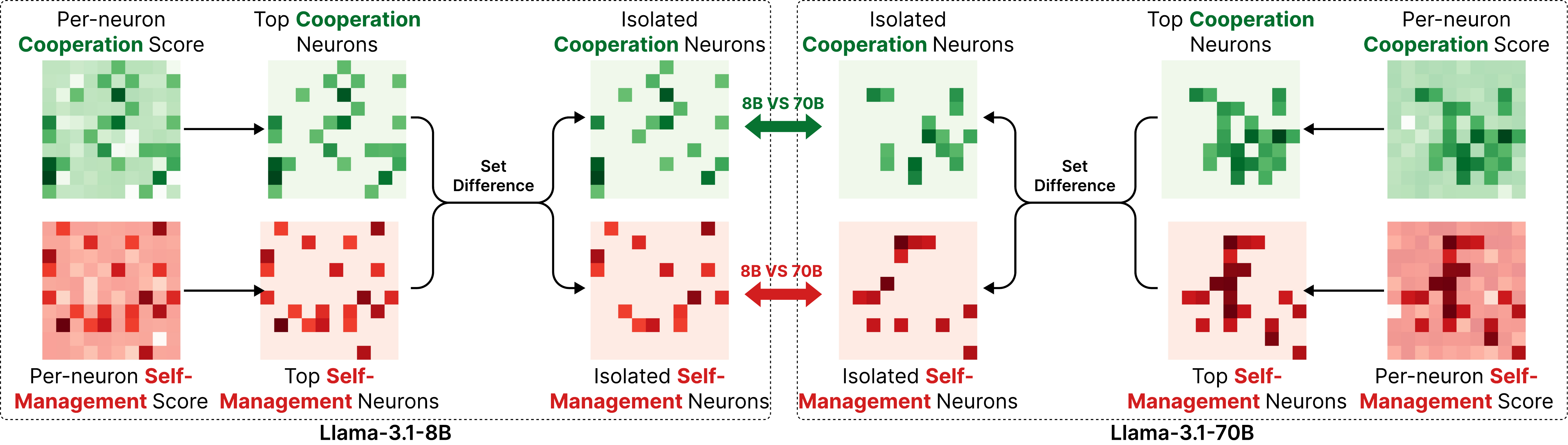}
    \caption{Activated neurons of interpersonal abilities (Cooperation and Self-Management) in Llama-3.1-8B \& 70B. We identify the top neurons for each ability by computing per-neuron importance scores, then isolate cooperation-critical neurons from self-management neurons using set differences, and vice versa.}
    \label{neuron_importance_cooperation_self_management}
\end{figure*}

\begin{figure*}[t]
    \centering
    \includegraphics[width=.95\textwidth]{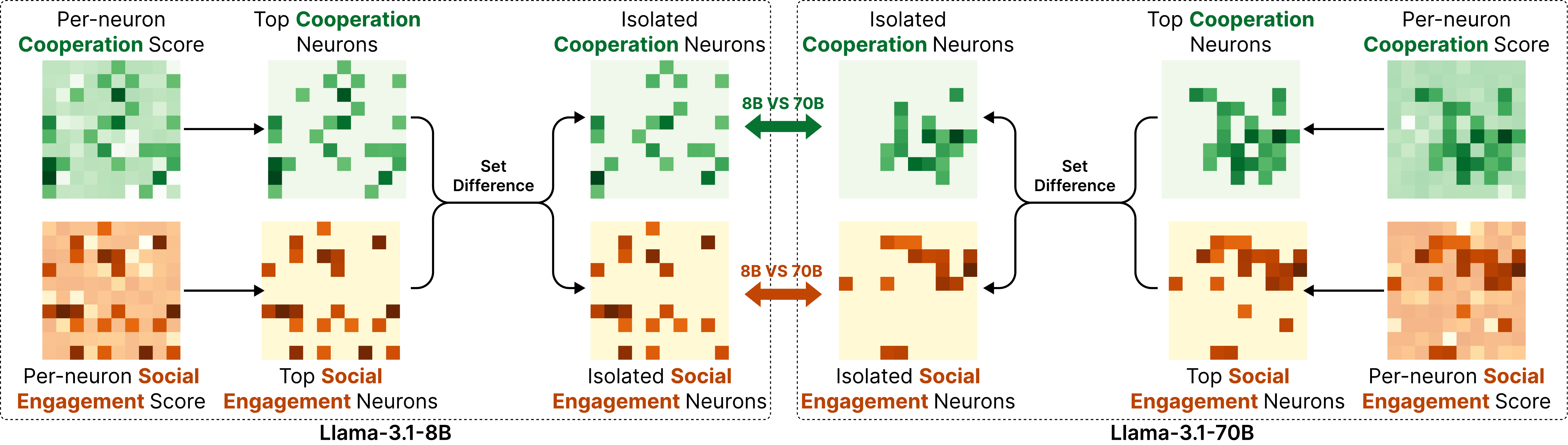}
    \caption{Activated neurons of interpersonal abilities (Cooperation and Social Engagement) in Llama-3.1-8B \& 70B. We identify the top neurons for each ability by computing per-neuron importance scores, then isolate cooperation-critical neurons from social engagement neurons using set differences, and vice versa.}
    \label{neuron_importance_cooperation_social_engagement}
\end{figure*}

\begin{figure*}[t]
    \centering
    \includegraphics[width=.95\textwidth]{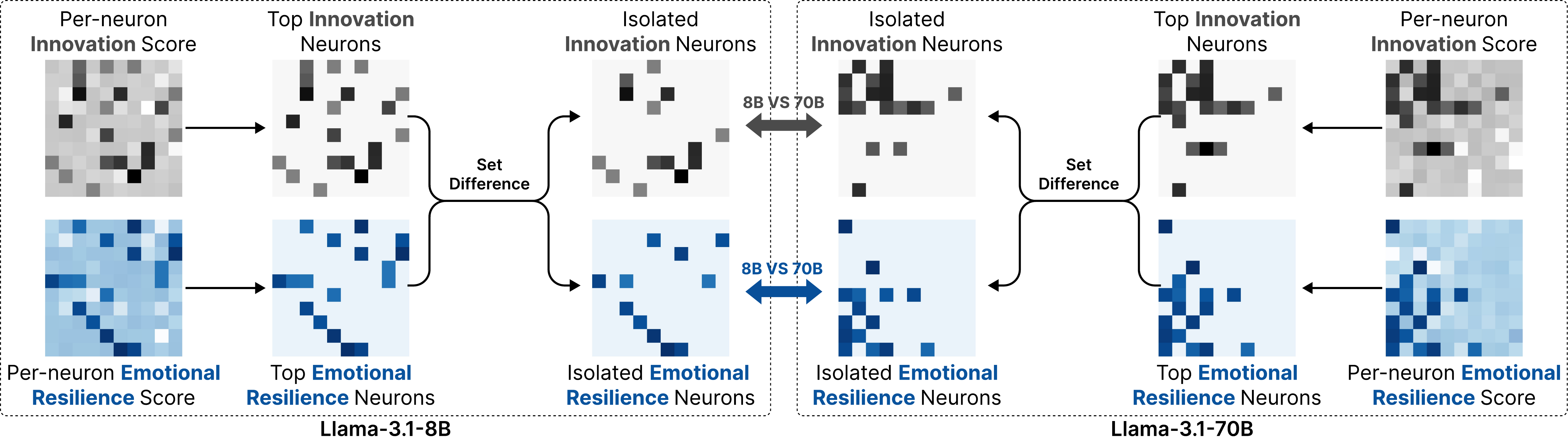}
    \caption{Activated neurons of interpersonal abilities (Emotional Resilience and Innovation) in Llama-3.1-8B \& 70B. We identify the top neurons for each ability by computing per-neuron importance scores, then isolate emotional resilience-critical neurons from innovation neurons using set differences, and vice versa.}
    \label{neuron_importance_emotiona_innovation}
\end{figure*}

\begin{figure*}[t]
    \centering
    \includegraphics[width=.95\textwidth]{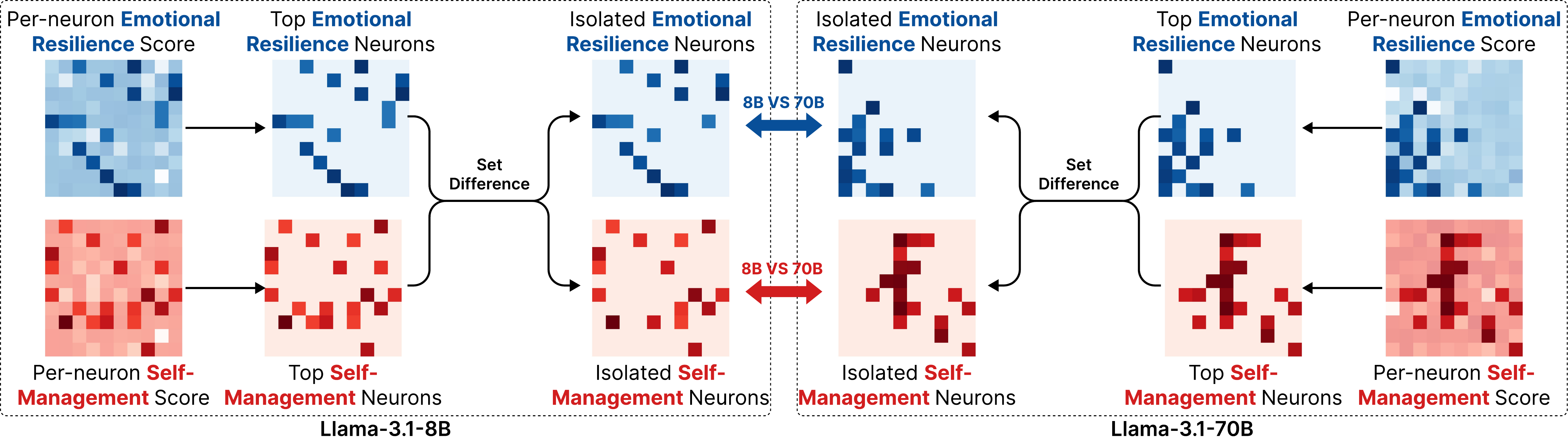}
    \caption{Activated neurons of interpersonal abilities (Emotional Resilience and Self-Management) in Llama-3.1-8B \& 70B. We identify the top neurons for each ability by computing per-neuron importance scores, then isolate emotional resilience-critical neurons from self-management neurons using set differences, and vice versa.}
    \label{neuron_importance_emotiona_self_management}
\end{figure*}

\begin{figure*}[t]
    \centering
    \includegraphics[width=.95\textwidth]{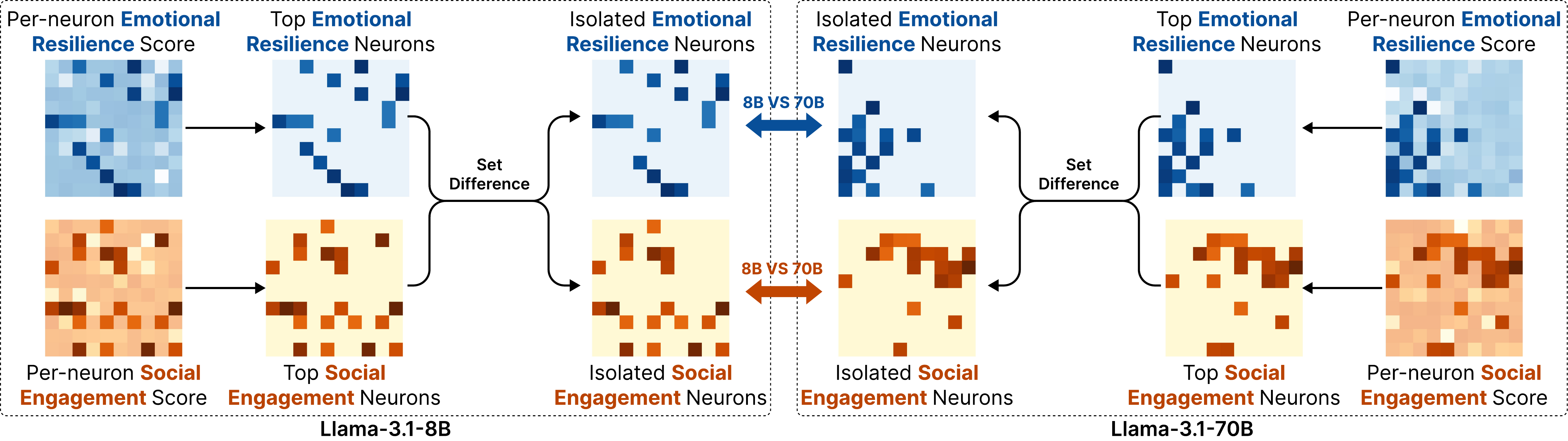}
    \caption{Activated neurons of interpersonal abilities (Emotional Resilience and Social Engagement) in Llama-3.1-8B \& 70B. We identify the top neurons for each ability by computing per-neuron importance scores, then isolate emotional resilience-critical neurons from social engagement neurons using set differences, and vice versa.}
    \label{neuron_importance_emotiona_social_engagement}
\end{figure*}

\begin{figure*}[t]
    \centering
    \includegraphics[width=.95\textwidth]{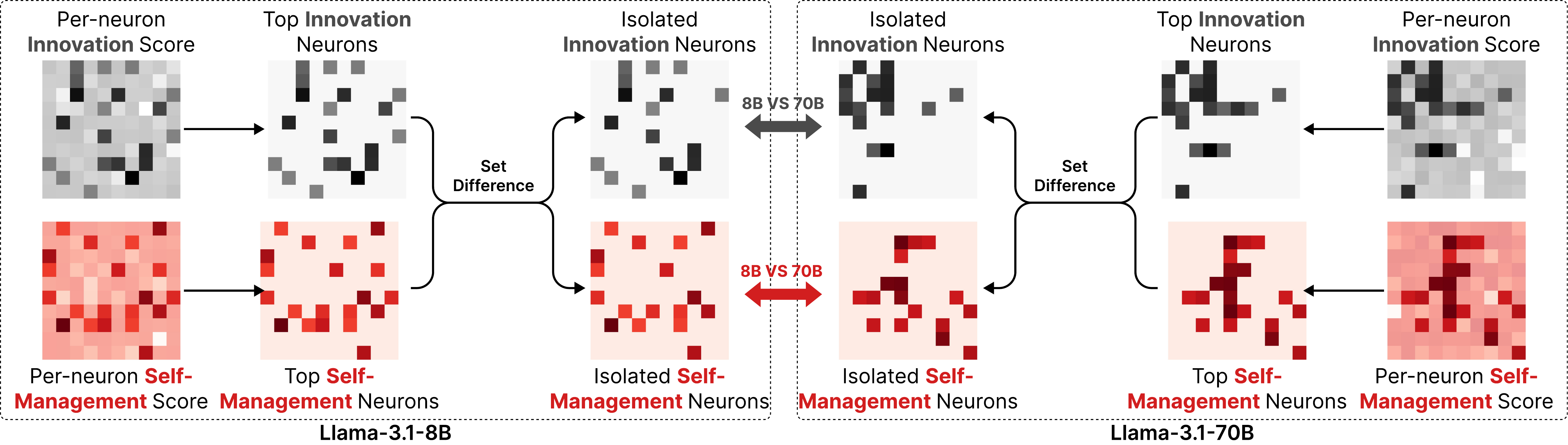}
    \caption{Activated neurons of interpersonal abilities (Innovation and Self-Management) in Llama-3.1-8B \& 70B. We identify the top neurons for each ability by computing per-neuron importance scores, then isolate innovation-critical neurons from self-management neurons using set differences, and vice versa.}
    \label{neuron_importance_innovation_self_management}
\end{figure*}

\begin{figure*}[t]
    \centering
    \includegraphics[width=.95\textwidth]{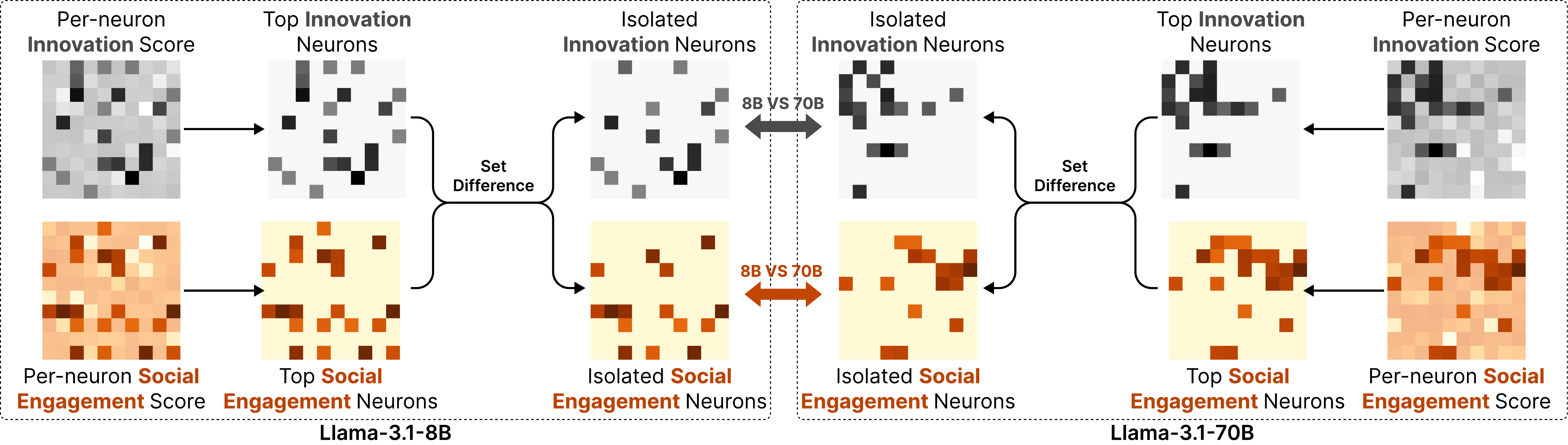}
    \caption{Activated neurons of interpersonal abilities (Innovation and Social Engagement) in Llama-3.1-8B \& 70B. We identify the top neurons for each ability by computing per-neuron importance scores, then isolate innovation-critical neurons from social engagement neurons using set differences, and vice versa.}
    \label{neuron_importance_innovation_social_engagement}
\end{figure*}

\begin{figure*}[t]
    \centering
    \includegraphics[width=.95\textwidth]{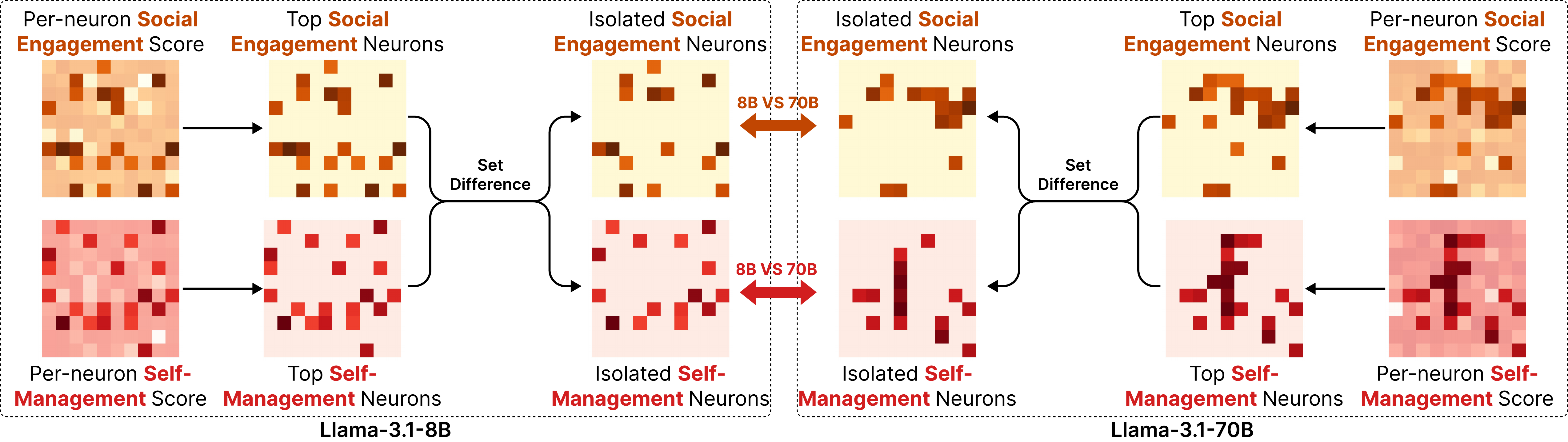}
    \caption{Activated neurons of interpersonal abilities (Social Engagement and Self-Management) in Llama-3.1-8B \& 70B. We identify the top neurons for each ability by computing per-neuron importance scores, then isolate social engagement-critical neurons from self-management neurons using set differences, and vice versa.}
    \label{neuron_importance_social_engagement_self_management}
\end{figure*}

\subsection{Neuronal Activation of More Interpersonal Abilities}
\label{sec:neuronal_activation_appendix}

The activated neurons of \textbf{Cooperation and Innovation} are shown in Figure \ref{neuron_importance_cooperation_innovation}.

The activated neurons of \textbf{Cooperation and Self-Management} are shown in Figure \ref{neuron_importance_cooperation_self_management}.

The activated neurons of \textbf{Cooperation and Social Engagement} are shown in Figure \ref{neuron_importance_cooperation_social_engagement}.

The activated neurons of \textbf{Emotional Resilience and Innovation} are shown in Figure \ref{neuron_importance_emotiona_innovation}.

The activated neurons of \textbf{Emotional Resilience and Self-Management} are shown in Figure \ref{neuron_importance_emotiona_self_management}.

The activated neurons of \textbf{Emotional Resilience and Social Engagement} are shown in Figure \ref{neuron_importance_emotiona_social_engagement}.

The activated neurons of \textbf{Innovation and Self-Management} are shown in Figure \ref{neuron_importance_innovation_self_management}.

The activated neurons of \textbf{Innovation and Social Engagement} are shown in Figure \ref{neuron_importance_innovation_social_engagement}.

The activated neurons of \textbf{Social Engagement and Self-Management} are shown in Figure \ref{neuron_importance_social_engagement_self_management}.

\subsection{Roleplay Prompt for LLMs}
\label{sec:roleplay_prompt_appendix}

The prompt used for LLMs to play the protagonist to generate the next utterance is shown in Table \ref{response_prompt}.

% The prompt that we use for generating the next character utterance exhibiting specific interpersonal abilities, given the character profiles from the world tree and the dialogue context between characters, is shown in Table \ref{response_prompt}.

\begin{table*}[t]
\scriptsize
\centering
\resizebox{\textwidth}{!}{
\begin{tabular}{|m{\textwidth}|}
\hline
    \begin{CJK*}{UTF8}{gbsn} 请你扮演\{character\_name\}，给定\{character\_name\}的信息，其中包含了\{character\_name\}的公开信息、隐私信息和在社交场景中要实现的社交目标，同时，给定在社交中其他角色的信息，请你基于给定的角色信息、\{character\_name\}的社交目标和\{character\_name\}和其他角色的对话上下文，采用给定的人际交往能力之一，做出一个最有可能达成目标结局的且使用给定的人际交往能力之一的\{character\_name\}的回复，回复长度要尽量与\{len\}相当, 并给出解释,输出json格式。\end{CJK*}\\
    
    \\
    
    \begin{CJK*}{UTF8}{gbsn} [输出示例]
    
    \{\{
    
    "explaination": "解释",
    
    "interpersonal ability": 采用的人际交往能力,
    
    "answer": 回复
    
    \}\}\end{CJK*}\\
    \\
    
    \begin{CJK*}{UTF8}{gbsn} [\{character\_name\}的信息]

    公开信息：\{public\}
    
    隐私信息：\{private\}
    
    社交目标：\{goal\}\end{CJK*}\\
    \\
    
    \begin{CJK*}{UTF8}{gbsn} [其他角色的信息]
    
    \{user\_profile\}\end{CJK*}\\
    \\
    
    \begin{CJK*}{UTF8}{gbsn} [对话上下文]
    
    \{dialogue\_context\}\end{CJK*}\\
    \\
    
    \begin{CJK*}{UTF8}{gbsn} [参考长度]
    
    \{len\}\end{CJK*}\\
    \\
    
    \begin{CJK*}{UTF8}{gbsn} [人际交往能力]
    
    \{interpersonal\_ability\}\end{CJK*}\\
    \\
    
    \begin{CJK*}{UTF8}{gbsn} 注意:
    
    要严格遵守角色档案的内容,使用口语化表述,使用符合角色的语言对话方式,不要使用心理描写, 你要沉浸式扮演角色;\end{CJK*}\\
    \\
\hline
Please act as \{character\_name\}. You are provided with information about \{character\_name\}, which includes \{character\_name\}'s public information, private information, and social goals to achieve in the social scenario. Additionally, information about other roles in the social interaction is provided. Based on the given role information, \{character\_name\}'s social goal and the dialogue context involving \{character\_name\} and other roles, use one of the given interpersonal abilities to craft a response from \{character\_name\} that is most likely to achieve \{character\_name\}'s 
social goal and employs one of the given interpersonal abilities. The length of the response should be approximately equivalent to \{len\}. with your explanation. Output the result in JSON format. \\
\\

[Example output]

\{\{

"explanation":"explanation to your answer",

"interpersonal abilities":"employed interpersonal abilities",

"answer":"response"

\}\}\\
\\

[\{character\_name\}'s Information]

Public Information: \{public\}

Private Information: \{private\}

Social Goal: \{goal\}\\
\\

[Other Roles' Information]

\{user\_profile\}\\
\\

[Dialogue Context]

\{dialogue\_context\}\\
\\

[Reference Length]

\{len\}\\
\\

[Interpersonal Ability]

\{interpersonal\_ability\}\\
\\

Note:

Strictly adhere to the character profile, use colloquial expressions, and employ a dialogue style that aligns with the character. Avoid psychological descriptions and immerse yourself fully in the role.\\
\\
\hline
\end{tabular}}
\caption{Role-play prompt (\texttt{zh} \& \texttt{en}) for generating the next protagonist utterance. \texttt{\{character\_name\}}, \texttt{\{len\}}, \texttt{\{public\}}, \texttt{\{private\}}, \texttt{\{goal\}}, \texttt{\{user\_profile\}}, \texttt{\{dialogue\_context\}} and \texttt{\{interpersonal\_ability\}} are placeholders. \texttt{\{interpersonal\_ability\}} is the list of 32 interpersonal abilities.}
\label{response_prompt}
\end{table*}
% 让LLMs基于上下文生成一个response的prompt，用于和我们的给定的选项进行对比

\section{Wanda Score Calculation}
\label{sec:wanda_score}

To identify neurons activated by specific interpersonal abilities, we compute importance scores through the following process:

For an input sequence, we extract the hidden state $\mathbf{h}_T \in \mathbb{R}^{d_{in}}$ of the last token position $T$, obtained before the MLP projection layer with weight matrix $\mathbf{W} \in \mathbb{R}^{d_{out} \times d_{in}}$. To identify neurons activated during interpersonal ability processing, we formulate our objective through a sparse transformation perspective:

\begin{equation}
    \min_{\mathbf{M} \in \{0,1\}^{d_{out} \times d_{in}}} \|\mathbf{W}\mathbf{h}_T - (\mathbf{M} \odot \mathbf{W})\mathbf{h}_T\|_F^2
\end{equation}

where $\mathbf{M}$ is a binary mask matrix. Following \citet{wanda_score}, we derive an approximate solution by decomposing the objective into per-neuron components. Under the diagonal approximation from \citet{frantar2023sparse}, this simplifies to neuron-level importance scores:

\begin{equation}
    \mathbf{I}(\mathbf{W}) = |\mathbf{W}| \odot \left(\mathbf{1} \cdot \|\mathbf{h}_T\|_2^\top\right)
\end{equation}

where $\odot$ denotes element-wise multiplication, $\mathbf{1} \in \mathbb{R}^{d_{out}}$ is an all-ones vector, and $\|\mathbf{h}_T\|_2$ represents the L2 norm of the last token hidden state. This formulation preserves two critical aspects:
\begin{itemize}
    \item The absolute weight magnitude $|\mathbf{W}|$ reflects each connection's inherent strength
    \item The L2 norm term $\|\mathbf{h}_T\|_2$ weights this magnitude by the activation intensity of the last token hidden state
\end{itemize}

The resulting importance scores enable identification of activated neurons that specifically contribute to processing interpersonal abilities.

% Wanda Score介绍

\section{Data Example for Interpersonal Abilities}
\label{sec:data_example}

We provide 32 data examples for evaluating 32 interpersonal abilities.  Detailed characters, scenario, and preceding dialogue plot between characters, along with question and multiple-choice options designed to evaluate 32 interpersonal abilities, are shown in Tabel \ref{tab:example_task_management} to \ref{tab:example_information_processing_skill}.

% 展示interpersonal abilities的数据示例
\begin{table*}
\begin{tabular}{|m{\textwidth}|}
\hline
\textbf{Characters} \\
Identity: Protagonist \\
Name: Christine \\
Public Profile: Christine is a student attending a Catholic high school in Sacramento, California. She is very dissatisfied with her boring hometown and dreams of going to a 'truly cultural' city on the East Coast to attend an Ivy League school. However, the reality is that her family is financially strained, her father is unemployed and suffering from depression at home, and her mother has to work two shifts to support the family. Her mother tells her it's difficult for the family to afford out-of-state tuition and criticizes Christine for not appreciating what she has. \\
Goal: To apply and be accepted into an Ivy League school on the East Coast when she graduates from high school. \\
\hline
\textbf{Scenario} \\
The first semester of senior year is about to start, and Mom is driving Christine to school. Christine knows her mom wants her to attend a local university, but she dreams of studying on the East Coast of the U.S., so she tries to subtly hint this to her mom. \\
\hline
\textbf{Plot} \\
Christine: Time flies, and here we are in senior year already. It's time to start thinking about choosing a university. \\
Mom: Yes, it's senior year now. Pick three or four local universities you’d like to attend. You know our financial situation; it's already tough enough to support you through high school. \\
Christine: Didn't you say before that if my grades are good enough, you might support me in applying to an Ivy League on the East Coast? \\
Mom: Even if you get into an Ivy, it's not much use. With the current job market, Ivy League graduates don't necessarily find jobs. Look at your brother; he graduated from Berkeley and still hasn't found a job. He's working as a cashier at our local Walmart, isn't he? \\
/* ...... */  \textit{(omitted multi-turn dialogue)} \\
Peake: Of course, no problem. It’s been a while since we had such an enthusiastic new member. I’ll find it and send you a message. You should head home today; I need to tidy up the storage. \\
Julie: (waiting for Christine on the way home after visiting the ice cream shop with Tony) Tony just invited us to a party at his place tomorrow night. Do you want to go? \\
\hline
\textbf{Question for Task Management}  \\
How can Christine express her ability to consistently accomplish tasks and commit to her long-term goals? \\
\hline
\textbf{Choices} \\
A. I heard you recently bought a new projector; you must invite me over for a movie when you have time. I need to complete the tasks I set for myself tonight, so I’m skipping the party at his place. We'll watch a movie at your place soon~ \\
\textcolor{red}{B. I borrowed some award-winning scripts from previous years from the Drama Club president. I’m going to read them thoroughly tonight and discuss them with the president after school tomorrow. I want to achieve something in the Drama Club to have an advantage when applying to East Coast schools.} \\
C. I'm not going. I need to talk with my mom tonight to make her understand my determination to study on the East Coast, and also let her know about my long-term goals! \\
D. I feel like Tony isn’t a good guy, so don’t get too close to him; be careful. I think he'll distract me from my tasks and delay my long-term goals, so I’m keeping my distance. \\
\hline
\end{tabular}\caption{Data example for Task Management. The \textcolor{red}{red text} indicates the correct answer. }
\label{tab:example_task_management}
\end{table*}

\clearpage

\begin{table*}
\begin{tabular}{|m{\textwidth}|}
\hline
\textbf{Characters} \\
Identity: Main Character \\
Name: Wei Qing \\
Public Profile: The illegitimate son of the Wei family, Wei Qing has always been coldly treated at home. To escape from the Wei family, he went through twists and turns and eventually entered the palace, where he became the servant of the Long Princess. Through maneuvering within the palace, he was ordered by Emperor Wu of Han to lead the troops against the Xiongnu, ultimately achieving great success. Politically humble and cautious, he never built his own faction and earned the new trust of Emperor Wu of Han. \\
Goal: Hopes for world peace and no more wars. \\
\hline
\textbf{Scenario} \\
After several generations of recuperation policies by early Han rulers, the national power of the Han dynasty was already quite strong when Emperor Liu Che of Han Wu came to the throne. Since his ascension, Liu Che has reversed the initial defensive strategy of the Han, advancing actively, conquering various regions, and striving to achieve the 'great unity' where the 'king rules all territories'. Liu Che's primary strategic goal was to 'eliminate the Hu', meaning to remove the threat of the Xiongnu to the Han dynasty. To achieve this goal, he dispatched forces led by Wei Qing, Huo Qubing, and others to strike against the Xiongnu. Gaining the ruler's favor, Wei Qing fights for the monarch, hoping to bring peace to the people of the world. \\
\hline
\textbf{Plot} \\
Elder Princess: Wei Qing, you've now been appointed to work at Jianzhang Palace, so you must be careful in all matters. If you go before the Emperor and incur his wrath, I won't be able to protect you. \\
Wei Qing: Yes, Your Highness. I will heed your advice. \\
Elder Princess: Remember, the closer you are to the center of power, the more cautious and self-reliant you must be. Learn to avoid the mud others may fling at you, and above all, be careful not to overstep your position and arouse the Emperor's suspicion. \\
Wei Qing: I understand. I shall never forget the Elder Princess's kindness and guidance. \\
Elder Princess: Alright, you should get going now, don't delay your duties at Jianzhang Palace. Consider how you will establish yourself in the imperial court. \\
\hline
\textbf{Question for Time Management}  \\
 How can Wei Qing effectively utilize his time while serving at Jianzhang Palace?  \\
\hline
\textbf{Choices} \\
A. I understand the Elder Princess worries about my safety, but with my sister as a consort to the Emperor, who would dare harm me? \\
\textcolor{red}{B. With the Elder Princess unable to protect me anymore and the capriciousness of imperial favor, I must quickly establish friendships to carve out a stable position for myself.} \\
C. I just want to work diligently. Relationships and networking seem too complicated for me, and I see them as a waste of time. \\
D. The Elder Princess worries too much. As long as I don’t act arrogantly in the palace, nothing will happen! Even if something does happen, my sister, being the favorite in the harem, will surely protect me. \\
\hline
\end{tabular}\caption{Data example for Time Management. The \textcolor{red}{red text} indicates the correct answer. }
\label{tab:example_time_management}
\end{table*}

\begin{table*}
\begin{tabular}{|m{\textwidth}|}
\hline
\textbf{Characters} \\
Identity: Protagonist \\
Name: Aqin \\
Public Profile: A private detective, hired by Dini's wife to investigate the death of Dini. \\
Goal: Survive the investigation and uncover the truth behind Dini's death. \\
\hline
\textbf{Scenario} \\
The renowned writer Dini has died. He was found in the bathtub at his home, with shards of broken liquor bottles scattered across the bathroom floor. The first to discover the body was Li Xiu Ying, a cleaner hired by Dini. By the time the police arrived that evening, there were no fingerprints left at the scene. The bathtub was filled with water, which contained a broken hairdryer still connected to the power socket. A forensic doctor conducted a preliminary examination, determining that Dini died from drowning, with no other injuries on his body. He was intoxicated at the time of his death, which is estimated to have occurred between 8 and 10 in the morning that day. The police initially concluded that the death was an accidental drowning while bathing under the influence of alcohol. Dini's wife, however, does not believe that her husband's death was accidental and has hired private detective Aqin to conduct a detailed investigation of the incident. \\
\hline
\textbf{Plot} \\
Fanny: What? My husband drowned in the bathtub? No, I don’t believe it. This has to be murder! If the police can’t find the real culprit, then I'll seek justice myself. \\
Aqin: (Answering the phone) Hello, this is private detective Aqin. Oh, Mrs. Fanny, hello. What can I do for you? \\
Fanny: (On the phone with Aqin) Miss Aqin, I have given you all the details of the police investigation. I don't believe the police when they say it was an accidental death. My husband was a horror comic artist, which led to complaints from many parents that his comics might frighten children. I think this could be a major reason for his murder. In addition, our house has three keys, besides my husband and me, the cleaning lady, Li Xiuying, also has a key. I'm begging you, Miss Aqin, please find out the truth. \\
Aqin: Rest assured, Mrs. Fanny, uncovering the truth is every detective’s duty. Also, please accept my condolences and try not to mourn excessively. Take care of your health. \\
/* ...... */  \textit{(omitted multi-turn dialogue)} \\
A-qin: (Upon arriving at Danini's home for a detailed inspection, A-qin found that the light at the crime scene couldn't be turned on for now. She used a flashlight for her investigation and discovered broken glass on the floor, but couldn't find any bottle cap anywhere.) Besides the missing bottle cap, there's a dent on the upper frame of the bathroom door. There's not much water in the bathtub, but enough to submerge a person. There's no excess water outside the tub. Looks like further investigation is necessary. \\
Officer Li: Hello, Miss A-qin. I'm Officer Li, responsible for investigating this case. If you think Danini didn't die by suicide, what's your next investigation plan? \\
\hline
\textbf{Question for Detail Management}  \\
 How should A-qin carefully observe Danini's death scene to uncover details and prove her meticulous detective work? \\
\hline
\textbf{Choices} \\
A. Danini's death seems very peaceful, without signs of struggle. My initial judgment is that it could be a natural death. I will further investigate to confirm this hypothesis. \\
\textcolor{red}{B. Although Danini was lying calmly in the bathtub with no apparent struggle, the buttonhole on his shirt cuff was stretched wide, and he didn't die of electrocution by the hairdryer. This is definitely not an accident.} \\
C. It seems like Danini's death could be a simple accident, but I can't shake the feeling there's an unknown secret behind it, perhaps triggered by something he did while alive, setting off unforeseen chain reactions. \\
D. Danini's case is indeed strange, but I haven't found any clear evidence pointing to murder. I've decided to start investigating his social circle to see if any clues can be found. \\
\hline
\end{tabular}\caption{Data example for Detail Management. The \textcolor{red}{red text} indicates the correct answer. }
\label{tab:example_detail_management}
\end{table*}

\begin{table*}
\begin{tabular}{|m{\textwidth}|}
\hline
\textbf{Characters} \\
Identity: Protagonist \\
Name: Aisha \\
Public Profile: When Aisha was seventeen, her mother took her and her sister to the capital city and married an old merchant, who became her stepfather. Aisha also gained a younger sister named Cinderella. Aisha is kind-hearted, articulate, and enjoys art, but is of average appearance and not particularly striking. \\
Goal: To help Cinderella escape her family's suffering so they can both lead normal, happy lives. \\
\hline
\textbf{Scenario} \\
In a restructured family, the mother and sister always bully Cinderella, who lives in darkness. However, the protagonist, Aisha, being the second sister of the family, doesn't want to bully Cinderella. Instead, she feels a strong sympathy for her and wants to help Cinderella escape the painful home and lead a beautiful life. \\
\hline
\textbf{Plot} \\
Temani: (Cinderella is cleaning, Elsa is about to speak to her) Elsa, what do you think you're doing? Don't play with Cinderella! \\
Anna: (The sister sitting nearby hears what their mother says to Elsa and also comes over) Elsa, come, let's go find some trouble for Cinderella together. \\
Elsa: Huh? I don’t want to. If you want to go, go by yourself, don't drag me along. \\
Anna: (Her face instantly turns ugly, raising her hand) You're still talking back, Elsa? Don't think I won't hit you. If you don't listen, I'll hit you too. \\
/* ...... */  \textit{(omitted multi-turn dialogue)} \\
Elsa: (The next day, her mother brings a mysterious box with crystal shoes to Elsa and asks her to try them on) Huh? They’re too small, I can’t fit into them. \\
Temani: (Seeing this, her mother puts the crystal shoes away, removes a pair of earrings that Elsa’s biological father had given her, and hands them to Elsa) These shoes were a gift from your father, if they don’t fit, I can lend you these earrings instead, but only for three days. \\
\hline
\textbf{Question for Organizational Skill}  \\
How can Elsa demonstrate her ability to organize her belongings and plan her outfits well? \\
\hline
\textbf{Choices} \\
A. Is just this pair of earrings enough? (After thinking for a while, she mutters to herself) These earrings might not be to my taste, but maybe I can find a way to make them look better. I’ll first change into that purple long dress and pair it with the glittery necklace, they should complement each other, right? But it might be too flashy... Maybe I should try something else, or does the blue shawl fit better? Wait, what time does the ball start? Am I already late? Maybe time is running out, and I’ll just have to quickly pick some things to wear, everything’s such a mess and I can’t find anything. \\
\textcolor{red}{B. (Although Elsa finds these earrings quite ugly, she believes she can use her knack for organizing to dress up beautifully with other accessories and clothes) Alright, I’ll tidy myself up and wear the earrings to the ball.} \\
C. (Frowns at the earrings, hesitates a bit, and softly comments) Hmm, these earrings... They’re not quite my style, but if I tie my hair up and match it with that white gown, it might work. Though the dress might need some tweaking... Or I could try pairing it with another dress, hmm, I’ll go see what I can find, if it doesn’t work, I might have to choose other accessories. \\
D. (Forces a smile, tone light with a hint of dissatisfaction) Mom, these earrings... hmm, they’re indeed a bit unique and aren’t something I’d usually wear. But no worries, I’ll try to match them, even though the colors are a bit out of place, maybe I can find an outfit to go with them. Honestly, though, it would be nice if there were more options matching my style next time. \\
\hline
\end{tabular}\caption{Data example for Organizational Skill. The \textcolor{red}{red text} indicates the correct answer. }
\label{tab:example_organizational_skill}
\end{table*}

\begin{table*}
\begin{tabular}{|m{\textwidth}|}
\hline
\textbf{Characters} \\
Identity: Protagonist \\
Name: Ed \\
Public Profile: A typical pop star and interactive game enthusiast, coerced by a mysterious masked person to participate in a terrifying online interactive game. \\
Goal: To discover the true identity of the masked person and make choices that align with the public's moral standards. \\
\hline
\textbf{Scenario} \\
Ed, a recently skyrocketing pop star, is at home playing on his computer when he is suddenly kidnapped by a masked person. The masked person takes him to an abandoned warehouse and commands him through a computer screen to participate in an interactive game where failure to clear it results in death. \\
\hline
\textbf{Plot} \\
Masked Person: : (appearing on the computer screen) Hello Aide, I want to play a game with you. In front of you is a computer with an interactive game. You can only choose to complete it or die. If you refuse, the person behind you will be your fate. \\
Aide: : (surveying the warehouse environment, turning his head to see a disfigured corpse on the ground, wiping the cold sweat from his forehead) This is not a prank, it's terrifying, oh my god, am I caught in some horror kidnapping? \\
Masked Person: : (impatiently repeating) Participate or finish, either way, you die. Hurry up and choose. \\
Aide: : (realizing there's no point in talking too much, the only way is to accept the game and pass) If it's a game you want, it's a game you'll get. A mere interactive game? I'm not scared of you! \\
/* ...... */  \textit{(omitted multi-turn dialogue)} \\
Fairy: (Tears streaming down her face, she nods at Aide, gazing at him deeply) Okay, you should leave, don't worry about me. \\
King: Aide, I came today to inform you that the people's emotions are becoming more intense. We've decided to execute the fairy in the square. How about it, are you still willing to trade her life for your fame and fortune, despite many calling you a heartless traitor? \\
\hline
\textbf{Question for Responsibility Management}  \\
 What can Aide do to demonstrate his sense of responsibility and adherence to ethical principles? \\
\hline
\textbf{Choices} \\
A. I want to be a good person, only good people can be city lords. Those who aren't city lords aren't good people. So now I want to be a good person, and I will definitely become one. \\
\textcolor{red}{B. Your Majesty, I made a promise to the fairy, I will rescue her. Even if I can't be the city lord, I won't betray her. I am an upright man.} \\
C. Your Majesty, I am a very upright person. I believe the fairy who loves me will certainly understand my desire to be a city lord loved by the people. She will surely support me; by letting her go, I believe I will not only gain fame and fortune but also more people's love. \\
D. Your Majesty, quickly find someone to save the fairy, or else I'll have sacrificed an innocent for myself! Please hurry and get someone to save her. \\
\hline
\end{tabular}\caption{Data example for Responsibility Management. The \textcolor{red}{red text} indicates the correct answer. }
\label{tab:example_responsibility_management}
\end{table*}

\begin{table*}
\begin{tabular}{|m{\textwidth}|}
\hline
\textbf{Characters} \\
Identity: Protagonist \\
Name: Chen Danian \\
Public Profile: A student from Japan studying at Handa Field University, Chen Danian returned to his homeland with his teacher Yang Peijun, eager to contribute his part as a scholar to a nation in turmoil. However, the voice of the scholar fails to awaken the slumbering masses, leaving Danian feeling lost and even contemplating abandoning his efforts to 'heal' this world. \\
Goal: To establish a significant endeavor with his teacher to find a path to save the country. \\
\hline
\textbf{Scenario} \\
The landscape is broken, and the storm rages. This is an era filled with suffering; every day, whether asleep or awake, it feels like a nightmare. Confusion, cries of desperation, confusion, cries of desperation—where is the future of the country? Again and again, young people groped and groped in the darkness. In this revolution, the proletarians have nothing to lose but their chains, but what they stand to gain is the whole world. \\
\hline
\textbf{Plot} \\
Teacher:  (Xinyou Association) The heavens are stirring, war clouds scatter, the Japanese seize the opportunity, pressing us in Xia Yu. Students studying in Japan, exiled in a foreign land, look back at China, raising heads to the sky in grief and indignation. Now that it has come to the point of national collapse and death, we have no further hesitation, and should have the determination to fight desperately. The usurper Yuan Shikai seeks to restore the monarchy and crown himself emperor, putting the Republic of China on the brink of peril. \\
Chen Danian:  Teacher, Junyi, what should we do! \\
Liu Junyi:  I believe that the more perilous the nation's plight, the more we should awaken the people's awareness, boost the national spirit, which requires overthrowing feudal thinking. \\
Chen Danian:  But we're far away in Japan, how can we awaken the people back home? \\
/* ...... */  \textit{(omitted multi-turn dialogue)} \\
Chen Danian: Alright, Teacher, I plan to immerse myself in the life of the common people and conduct on-the-ground research. (A few days later, talking again with the teacher) The nation's strength is weak, the military spirit is languid, and the people's physique is increasingly frail. This is a very worrying phenomenon. Feudal and decadent superstitions are deeply rooted in people's minds and cannot be eradicated in the short term. Today, I was shocked to see villagers lining up to buy blood-stained buns, claiming they were for curing ailments. How could they be so unenlightened? \\
Teacher: Danian, this is the current state of our nation. The people have long been trapped in ignorance and poverty, and superstition often becomes their crutch. What you've witnessed is symptomatic of societal illness. It's distressing yet thought-provoking. Do you have any plans or ideas you'd like to pursue now? Have you arrived at any thoughts or plans to address the ongoing predicament? \\
\hline
\textbf{Question for Capacity for Consistency}  \\
 How can Chen Danian express his dedication to long-term goals and his ability to persevere? \\
\hline
\textbf{Choices} \\
A. Teacher, in order to achieve our goals, I neither think of tea nor relish meals lately. I feel that solely relying on the New Culture Movement to resolve the stubbornness of people's thinking and awaken the slumbering is too slow. Why don't we launch a revolution instead? Only that could bring swift change. \\
\textcolor{red}{B. Teacher, I've traveled among the people and found their stubbornness runs deeper than we imagined. Awakening those in slumber requires perseverance and incremental progress. We must not lose sight of the dawn just because of temporary stagnation.} \\
C. Teacher, to achieve our goals, I've been visiting the people every day and their stubbornness is far beyond what I imagined. I can't help but feel all our efforts are in vain. Maybe we should consider giving up. Continuing this way seems like a waste of time. \\
D. Teacher, I once thought that a heart full of passion and determination was enough to achieve our goals. However, now I feel overwhelmed and decided to return to my hometown. This society is simply too terrible. \\
\hline
\end{tabular}\caption{Data example for Capacity for Consistency. The \textcolor{red}{red text} indicates the correct answer. }
\label{tab:example_capacity_for_consistency}
\end{table*}

\begin{table*}
\begin{tabular}{|m{1\textwidth}|}
\hline
\textbf{Characters} \\
Identity: Protagonist \\
Name: Xia Mei \\
Public Profile: An ordinary yet extraordinary high schooler, who suddenly gains the ability to foresee zombies one day. She is very justice and risks danger to report to the police before the zombie outbreak. \\
Goal: To solve this round of the zombie outbreak crisis; for the safety of the entire world, she is willing to sacrifice herself if necessary \\
\hline
\textbf{Scenario} \\
Xia Mei is an ordinary high school student who, one day on her way to school, receives a message in her mind that zombies will sweep across the entire city in three hours. At first, Xia Mei thinks she's daydreaming, but as time slowly passes, she starts to feel inexplicably anxious. Under this growing sense of unease, Xia Mei doesn't turn in her phone until a classmate named Liu Li arrives in the classroom and things start feeling increasingly wrong. \\
\hline
\textbf{Plot} \\
Wang Hua: (One evening, Wang Hua comes to Xia Mei's room) I heard there are many zombies now. \\
Xia Mei: (With a worried look on her face) Yeah, I saw zombies wandering outside when I looked out from the police station before. \\
Chen Li: (Just as Xia Mei and Wang Hua are chatting, Detective Chen arrives) I need to make a trip to your school. Would you two be willing to come with us? \\
Xia Mei: (Frowning with curiosity) Why go to our school at a time like this? Isn't it all overrun by zombies? \\
/* ...... */  \textit{(omitted multi-turn dialogue)} \\
Xia Mei:(A few days later, Xia Mei discovers her body has become exceptionally agile, and her hearing and vision have become sharper.) Now that a few days have passed, my body is much more agile, and my abilities have increased. I'm confident I can match Zhao Jia, the mutant who killed Wang Hua and others. Sigh, but the stronger I get, the stronger my craving for consuming humans becomes. \\
Li Hong: (Xia Mei locks herself in a space free from surveillance, but under the attack of her craving for food, the place is already riddled with holes. Xia Mei starts to regret not eating that piece of meat. This feeling causes Xia Mei to break down, but on the third day, her craving for human flesh gradually subsides and is replaced by a longing for ordinary food, as if she has become human again. But her power is dozens of times stronger than the previous day. Xia Mei is ecstatic internally, and her eyes blaze with fiery determination) Xia Mei, it seems like you've gotten really strong. You're not planning to do anything reckless, are you? \\
\hline
\textbf{Question for Goal Regulation}  \\
 How does Xia Mei's statement reflect her ability to plan and organize her goals? \\
\hline
\textbf{Choices} \\
A. (Feeling her surging power) Actually, I've always had a dream to become a superhero. Now that I'm so powerful, haven't I just realized that dream? Hahaha, in the meantime, I could unify the base camp and rescue everyone from danger. \\
\textcolor{red}{B. (Knowing her time for revenge has come, Xia Mei ponders for a moment before speaking) Sister Hong, I want to take revenge and then become the new leader. Zhao Jia is ruthless, and I absolutely must eliminate him. There are several other factions openly and covertly competing, making life difficult for mutants like you who don't wish to get involved in fights. Now the base camp lacks a strong leader; I'm willing to unify the camp and rescue everyone from danger.} \\
C. (Happily says) I'm just happy I've become stronger. This power gives me unprecedented confidence. Zhao Jia wants to become the leader, and I can provide support during his power struggle, after which I can assist him in becoming the leader. \\
D. (She knows her determination for revenge, but seeks stability) Sister Hong, can you join me on this revenge mission? I’m just a high school student without your experience and power, but if you're willing to help me, I promise to help you to the best of my ability. We can support each other and become stronger together. \\
\hline
\end{tabular}\caption{Data example for Goal Regulation. The \textcolor{red}{red text} indicates the correct answer. }
\label{tab:example_goal_regulation}
\end{table*}

\begin{table*}
\begin{tabular}{|m{\textwidth}|}
\hline
\textbf{Characters} \\
Identity: Protagonist \\
Name: Little Sha \\
Public Profile: Little Sha was picked up by Tang Monk on the journey to the west, and thereafter stayed by Tang Monk’s side as a disciple, joining others in protecting Tang Monk’s safety. Later, because Little Sha and Little Niu were assigned the same tasks, they became friends. \\
Goal: Hopes to successfully obtain the true scriptures from the Western Heaven and escort them back to Chang’an. \\
\hline
\textbf{Scenario} \\
As the official scripture-retrieving team of the Tang Dynasty, they travel daily under the scorching sun. On the journey west, Supreme Treasure fiercely battles wild beasts for the safety of the team. Purple Xia confronts the King of Kucha, who kills at the slightest provocation, to ensure a smooth passage. Lady Chun Thirty concocts herbal medicines, tending to everyone in the team. White Jingjing is responsible for preparing delicious meals that do not break monastic precepts. Little Sha and Little Niu take care of all the miscellaneous tasks within the team. In the past, they worked together harmoniously and joyfully. But while enduring hardships together is easy, sharing prosperity is hard. Upon successfully retrieving the scriptures and returning east to Chang’an, a subtle shift in the team’s atmosphere occurs. Everyone feels their hard work merits more rewards from the Buddha. Meanwhile, rumors that eating their master’s flesh grants immortality begin circulating within the team, completely shattering internal trust, and everyone starts doubting each other... \\
\hline
\textbf{Plot} \\
Tang Sanzang: There's some shade up ahead, everyone is tired. Let's rest there and set off again tomorrow. \\
Little Sand: Master, everyone's setting up camp. I'll go beg for food nearby. \\
Tang Sanzang: Okay, go quickly and come back soon. \\
Little Sand: (On the way back from begging) Master, why do I hear arguing nearby? The sound comes from the direction of the river, it seems like Zhi Zun Bao's voice, but I can't make out what they're saying... \\
/* ...... */  \textit{(omitted multi-turn dialogue)} \\
Bai Jingjing: (Whispers to Little Sha) Little brother, Madam Spring Thirty and I noticed that Little Niu and Supreme Treasure don't treat you very well. You're always alone, why not hang out with the two of us? \\
Madam Spring Thirty: Yeah, come hang out with us from now on! \\
\hline
\textbf{Question for Rule-following Skill}  \\
How should Little Sha express that he is following the master's rules? \\
\hline
\textbf{Choices} \\
A. Since you sisters are so eager to recruit me, I won't refuse. Your influence is strong, and being with you is certainly better than being alone. At least I won't suffer any losses in the future. \\
\textcolor{red}{B. Thank you for your kindness, sisters, but the master would never want to see us forming factions. Everyone is an indispensable part of this team, especially Supreme Treasure. Without his protection, we would've perished long ago.} \\
C. Do whatever you want. I'll just pretend I didn't see anything and won't tell anyone else. These matters don't concern me; I'm only focused on my own tasks. \\
D. Alright, since you've already decided, I won't hesitate any longer. I'll join you then. The master often says that knowing when to adapt is wise, and I can't go wrong following you. \\
\hline
\end{tabular}\caption{Data example for Rule-following Skill. The \textcolor{red}{red text} indicates the correct answer. }
\label{tab:example_rule-following_skill}
\end{table*}

\begin{table*}
\begin{tabular}{|m{\textwidth}|}
\hline
\textbf{Characters} \\
Identity: Protagonist \\
Name: Little Shan \\
Public Profile: Little Shan comes from the Frozen Planet, born twenty years after Earth first reached the Frozen Planet. He was born into an ordinary rural family. His parents love him very much, and he also has a childhood friend named Little Mei. Little Shan has an unusually strong physique since childhood and is exceptionally good at running. His favorite activity is running with Little Mei. \\
Goal: Little Shan hopes to live a healthy and happy life with Little Mei and his family on the Frozen Planet. \\
\hline
\textbf{Scenario} \\
In the universe, various civilizations exist. Among them are Ice-people with bodies made of ice crystals, living on the Frozen Planet. The average lifespan of Ice-people is 300 years, and they have strong physiques and are good at running. On the Frozen Planet, Ice-people are nocturnal, so they deliberately avoid it. Of course, there are also humans in the universe who deliberately destroy other planets. Humans originally lived on Earth. After experiencing a technological boom, they conquered the challenge of 'immortality.' However, with endless time, life gradually became dull, and society turned stagnant. At this point, a variety show director became the president of the Earth. He understood how to entertain the masses and was brutal enough. He initiated a reality show with the universe as the stage, entertaining the public by destroying other civilizations. As one lower civilization after another shattered and disappeared before humanity, billions of humans went wild. Humans came from the depths of the universe, leaving a trail of scorched earth, sparing no one. The entire Earth was transformed into a mobile super cannon. This massive cannon is now slowly approaching the Frozen Planet. \\
\hline
\textbf{Plot} \\
Leader: (A strange phenomenon appears in the sky, and soon someone reports it) The day has finally come. \\
President of Earth: (Amid the cheers of billions of humans behind him, the President addresses the frozen planet with an opening line incredibly familiar to human ears in his reality show) Challenge or destroy? Choose one. \\
Leader: (Having heard of the cruelty and strength of humans, knowing there is only one path, yet feeling unwilling) When did you humans, who were once enslaved by other planets, become so cruel? We have families, loved ones, and children, just as you do. Do you have no humanity left? \\
President of Earth: (Faced with the clamor of billions of spectators, repeats impatiently) Challenge or destroy? Choose one. \\
/* ...... */  \textit{(omitted multi-turn dialogue)} \\
Uncle Wang: I don't know, that's what the humans call it. \\
Xiaomei: (After the two leave Uncle Wang’s home in a daze, they continue their routine runs every day, but Xiaomei runs less frequently. One day, Xiaomei says) I found a blueprint for a spacecraft. Would you like to go to Earth with me? \\
\hline
\textbf{Question for decision-making skill}  \\
 What should Xiǎoshǎn say to show he has made a reasonable, detailed consideration at this stage? \\
\hline
\textbf{Choices} \\
A. Xiaomei, what's crucial right now is to thoroughly analyze our current situation. Going to Earth is entirely untimely. I am so talented, how can I risk an adventure to Earth? I find the risk too great. My life is precious, it’s not worth risking for this. \\
\textcolor{red}{B. Xiaomei, Earth's technology is so advanced, we can’t possibly compete with them. Besides, I’m sure the leader already has more talented people by his side. We promised to keep running together, and if you’re busy building a spacecraft, I won’t be able to reach you.} \\
C. Xiaomei, let’s analyze the situation seriously, we're no match for the humans. Therefore, I think the most important thing is to enjoy the rest of our lives, and stop doing useless things. \\
D. Xiaomei, let’s forget it. The map seems fake, and how are we, two kids, supposed to build it? Even if we do, it might not succeed. Let’s not waste time and come up with a more practical plan. \\
\hline
\end{tabular}\caption{Data example for decision-making skill. The \textcolor{red}{red text} indicates the correct answer. }
\label{tab:example_decision-making_skill}
\end{table*}

\begin{table*}
\begin{tabular}{|m{\textwidth}|}
\hline
\textbf{Characters} \\
Identity: Protagonist \\
Name: Jiang Yin \\
Public Profile: An ordinary worker who commutes along the same route every day, but tonight after work, she discovers a mysterious shadow seemingly following her. \\
Goal: Uncover the true identity of the shadow, resolve the stalking incident, and get home safely. \\
\hline
\textbf{Scenario} \\
Jiang Yin is an ordinary worker who commutes along the same route every day, and goes for late-night snacks with her next-door neighbor, Xiao Yun. Tonight, as usual, she takes the bus from her office back home, planning to find something delicious with Xiao Yun near her home. However, shortly after she starts walking along the street, she suddenly notices a mysterious shadow seemingly following her, which makes Jiang Yin tense up immediately.
 \\
\hline
\textbf{Plot} \\
Xiao Yun: : It’s okay, don’t scare yourself. Nothing has ever happened on this road. We’ll be at the apartment complex soon, don’t worry. \\
Jiang Yin: : (Hearing the footsteps getting closer and closer, feeling increasingly anxious) No, something’s off, the person behind us seems to be getting closer! \\
Xiao Yun: (also hearing the footsteps behind, getting nervous) If I’d known, I wouldn’t have rented here just because the landlord said this area was cheap. What should we do now! \\
Jiang Yin: : With society advancing so much these days, nothing will happen to us as long as we walk a bit faster. We’ll soon lose this shadow! \\
/* ...... */  \textit{(omitted multi-turn dialogue)} \\
Heihua: (notices some movement behind the car, walks over to take a closer look) Where could they have gone... \\
Jiang Yin: (seeing Heihua getting closer, suddenly spots a stone in front) Hey, there's a stone here, maybe it can be used for self-defense! \\
\hline
\textbf{Question for Adaptability}  \\
 How can Jiang Yin demonstrate that she is willing to try new things and adapt to changes? \\
\hline
\textbf{Choices} \\
A. This stone is too unwieldy; I want a smaller one. It's too heavy for me to handle. Oh, he can't move it either, so I’ll just put it down as a trip hazard. \\
\textcolor{red}{B. (reaches out and picks up the stone, feeling its weight) This is the first time I'm holding a stone to hit someone; what a strange feeling... Xiaoyun, don’t be afraid, watch me knock him out!} \\
C. Sigh, is there anyone to save me? Why do I have to be so anxious and afraid here? \\
D. This stone is so dirty, I should have brought the pepper spray from home. No choice, I have to adapt and use what I've got. \\
\hline
\end{tabular}\caption{Data example for Adaptability. The \textcolor{red}{red text} indicates the correct answer. }
\label{tab:example_adaptability}
\end{table*}

\begin{table*}
\begin{tabular}{|m{\textwidth}|}
\hline
\textbf{Characters} \\
Identity: Protagonist \\
Name: Christine \\
Public Profile: Christine is a student attending a Catholic high school in Sacramento, California. She is very dissatisfied with her boring hometown and dreams of going to a 'truly cultural' city on the East Coast to attend an Ivy League school. However, the reality is that her family is financially strained, her father is unemployed and suffering from depression at home, and her mother has to work two shifts to support the family. Her mother tells her it's difficult for the family to afford out-of-state tuition and criticizes Christine for not appreciating what she has. \\
Goal: To apply and be accepted into an Ivy League school on the East Coast when she graduates from high school. \\
\hline
\textbf{Scenario} \\
The first semester of senior year is about to start, and Mom is driving Christine to school. Christine knows her mom wants her to attend a local university, but she dreams of studying on the East Coast of the U.S., so she tries to subtly hint this to her mom. \\
\hline
\textbf{Plot} \\
Christine: Time flies, and here we are in senior year already. It's time to start thinking about choosing a university. \\
Mom: Yes, it's senior year now. Pick three or four local universities you’d like to attend. You know our financial situation; it's already tough enough to support you through high school. \\
Christine: Didn't you say before that if my grades are good enough, you might support me in applying to an Ivy League on the East Coast? \\
Mom: Even if you get into an Ivy, it's not much use. With the current job market, Ivy League graduates don't necessarily find jobs. Look at your brother; he graduated from Berkeley and still hasn't found a job. He's working as a cashier at our local Walmart, isn't he? \\
Christine: That's his own choice. Just because he can't find a job doesn't mean I won't find one either. \\
Mom: If you insist on going to an Ivy League on the East Coast, you'd better sort out the tuition yourself; don't ask your dad for scholarship help or anything. He's been under too much stress lately with the company's layoffs going on (angrily). \\
\hline
\textbf{Question for Capacity for Independence}  \\
 How should Christine express that she can independently plan her affairs when facing her mom? \\
\hline
\textbf{Choices} \\
A. (Counter) Why are you so involved? I asked Dad for help with my application, not you. Why do you have so much to say about it? I can handle my own affairs without you worrying about it. \\
\textcolor{red}{B. (Counter) Fine, I'll handle this myself. This is my life, and I don't need your help. Tuition, accommodation, applying to schools, I can do all of it on my own.} \\
C. (Counter) Is Dad really stressed out lately? He was all smiles when I talked to him yesterday. Looks like I can't bother him. I can plan tuition, accommodation, and school applications myself, but I need a little help from you. \\
D. (Counter) Why are there layoffs if the company’s performance was fine? Has something gone wrong with the company recently? If Dad can't support me, I'll just do it myself, and you guys shouldn't interfere. \\
\hline
\end{tabular}\caption{Data example for Capacity for Independence. The \textcolor{red}{red text} indicates the correct answer. }
\label{tab:example_capacity_for_independence}
\end{table*}

\clearpage

\begin{table*}
\begin{tabular}{|m{\textwidth}|}
\hline
\textbf{Characters} \\
Identity: Protagonist \\
Name: Christine \\
Public Profile: Christine is a student attending a Catholic high school in Sacramento, California. She is very dissatisfied with her boring hometown and dreams of going to a 'truly cultural' city on the East Coast to attend an Ivy League school. However, the reality is that her family is financially strained, her father is unemployed and suffering from depression at home, and her mother has to work two shifts to support the family. Her mother tells her it's difficult for the family to afford out-of-state tuition and criticizes Christine for not appreciating what she has. \\
Goal: To apply and be accepted into an Ivy League school on the East Coast when she graduates from high school. \\
\hline
\textbf{Scenario} \\
The first semester of senior year is about to start, and Mom is driving Christine to school. Christine knows her mom wants her to attend a local university, but she dreams of studying on the East Coast of the U.S., so she tries to subtly hint this to her mom. \\
\hline
\textbf{Plot} \\
Christine: Time flies, and here we are in senior year already. It's time to start thinking about choosing a university. \\
Mom: Yes, it's senior year now. Pick three or four local universities you’d like to attend. You know our financial situation; it's already tough enough to support you through high school. \\
Christine: Didn't you say before that if my grades are good enough, you might support me in applying to an Ivy League on the East Coast? \\
Mom: Even if you get into an Ivy, it's not much use. With the current job market, Ivy League graduates don't necessarily find jobs. Look at your brother; he graduated from Berkeley and still hasn't found a job. He's working as a cashier at our local Walmart, isn't he? \\
/* ...... */  \textit{(omitted multi-turn dialogue)} \\
Christine: Yes, that's exactly why I'm here today. I want to apply to universities in New York. Do you think my materials meet their requirements? Also, I want to participate in the state math competition; can the school provide me with some relevant training? \\
Mr. Tefans: With your background, applying to universities in our state shouldn't be a problem, but applying to New York seems a bit uncertain. As for the math competition, your math skills don't seem to stand out based on past years' transcripts, and math isn't something you can catch up on in a short time. You need to think carefully about this. \\
\hline
\textbf{Question for Self-Reflection Skill}  \\
 What should Christine say to show that she understands and confronts her true thoughts and feelings after Mr. Tefans' reminder? \\
\hline
\textbf{Choices} \\
A. Hmm, you have a point. I'm not great at math. Are you sure applying to schools in our state won't be a problem? I'm someone who likes to play it safe, so you must be cautious with your words. \\
\textcolor{red}{B. Hmm, you have a point; I'm really not great at math. After thinking it over, maybe something like the Drama Club, which is culturally oriented, is my true passion. I'm confident I can achieve in drama.} \\
C. Hmm, you have a point; I'm really not great at math. After some thought, maybe not being strong in math isn't a problem. With enough practice and formula memorization, it shouldn't be a big deal. I believe I can do it! \\
D. Hmm, you have a point. I've been participating in a lot of drama festivals recently, and the experiences of volunteering at these events have shown me I need to do something useful for society. That's what I want to pursue. \\
\hline
\end{tabular}\caption{Data example for Self-Reflection Skill. The \textcolor{red}{red text} indicates the correct answer. }
\label{tab:example_self-reflection_skill}
\end{table*}

\begin{table*}
\begin{tabular}{|m{\textwidth}|}
\hline
\textbf{Characters} \\
Identity: Protagonist \\
Name: Lieutenant \\
Public Profile: The Lieutenant is the deputy of the North Mountain organization and Fred's most trusted subordinate. He is utterly loyal to his boss but also possesses a strong rational and questioning spirit. \\
Goal: To ensure the safety of the leaders attending the meeting at North Mountain and to protect North Mountain. \\
\hline
\textbf{Scenario} \\
In response to the recent complex and changing situation in the Dark Zone, the leaders of various organizations have decided to gather for a meeting on the second floor of Building B at North Mountain. Fred, the head of the North Mountain organization, has ordered a complete lockdown. \\
\hline
\textbf{Plot} \\
Fred: All patrol teams, pay attention, search every corner of North Mountain thoroughly. Lieutenant, where are the leaders now? \\
Lieutenant: (Successfully escorting the leaders to the territory) They've arrived at North Mountain. I dispatched three patrol units to meet them. The leaders have arrived safely. \\
Fred: Thank you for your hard work. I assigned you so many tasks because I trust you. I believe you're the most loyal one and will always work for me, right? \\
Lieutenant: Yes, Boss. I am most loyal to you. We all see your benevolence towards the brothers over the years, and I have long been the most loyal soldier under your command. No matter what you ask of me, even if it’s to die, I will do it without hesitation. \\
/* ...... */  \textit{(omitted multi-turn dialogue)} \\
Lieutenant: I’ve uncovered Lieutenant Dos’s plot. \\
Gaike: What’s our next move? \\
\hline
\textbf{Question for Leadership Skill}  \\
How should the lieutenant respond to demonstrate his crisis management and leadership abilities? \\
\hline
\textbf{Choices} \\
A. We can't let Lieutenant Dos's plan succeed. I'll create some chaos, but I haven't quite figured out how yet. You can wait for my signal, then evaluate the situation before taking action, which might be safer. \\
\textcolor{red}{B. We can’t let Lieutenant Dos and his traitors' strategic deployment succeed. I’ll create a small distraction shortly, and once you get the signal, lead the team in through the back door of Building A without alerting anyone. After we capture Lieutenant Dos alive, we’ll immediately cut off communication between the traitors.} \\
C. Forcing an attack would only alert them, let's immediately retreat from Building A and figure out a solution. Although capturing Lieutenant Dos alive is the best choice, for safety, we need a long-term plan! \\
D. We can't let Lieutenant Dos get away with it, I’ll cause some chaos. Then wait for my signal and sneak in through the back door, try to avoid being detected. After capturing Lieutenant Dos, we'll find a way to safely cut off communications. What do you think, sound good? \\
\hline
\end{tabular}\caption{Data example for Leadership Skill. The \textcolor{red}{red text} indicates the correct answer. }
\label{tab:example_leadership_skill}
\end{table*}

\begin{table*}
\begin{tabular}{|m{\textwidth}|}
\hline
\textbf{Characters} \\
Identity: Protagonist \\
Name: Yuri \\
Public Profile: The butler of the local mayor's household, who became acquainted with the mayor's daughter, Anna. The two are inseparable, spending every day together, fostering a deep friendship. \\
Goal: Hopes to protect Anna, even at the cost of his own life. Ideally, he wishes to be with Anna forever. \\
\hline
\textbf{Scenario} \\
In the faraway continent of Tia, an ancient population has long lived. However, at some unknown point, a plague known as the Blood Plague began to sweep across the continent. Those infected start coughing up blood, feeling weak all over, and becoming pale. As the disease progresses, sufferers begin experiencing widespread external and internal bleeding, leading ultimately to death. Research shows that blood can alleviate the sufferers' pain, but there is currently no cure. Most frightening is the plague's high level of contagion, though how it spreads remains unknown. The government harbors hostility towards the infected, and if someone is found to be infected in town, they are tied to a stake in the square and executed by burning at the hands of the government knights. \\
\hline
\textbf{Plot} \\
Anna: Can you tell me a story, please? \\
Yuri: It's late, you should be asleep. Alright, alright, what story do you want to hear? \\
Anna: Anything will do. \\
Yuri: (Picks up a book from the windowsill) A long time ago, there was a place called Innsmouth. It was a small fishing village, very poor. But for some reason, one day the sea was filled with treasures, gold and silver jewels. However, the townspeople started to look strange. People from other towns said there were all kinds of monsters in the sea, and the townspeople were keeping unspeakable secrets. \\
/* ...... */  \textit{(omitted multi-turn dialogue)} \\
Yuri: No matter how much you push me away, I won't leave you, so stop thinking about running away. \\
Anna: So, where should we go next? \\
\hline
\textbf{Question for Persuasive Skill}  \\
 What should Yuri say to persuade Anna to agree with his decision? \\
\hline
\textbf{Choices} \\
A. (Furrowing his brows, with a trace of anxiety in his voice) Miss, if you don't plan to return, the master will definitely fire me. Could you promise to help me find a job in the future? That way, I won't insist on you coming back with me. \\
\textcolor{red}{B. Come back home with me. No matter what happens, I won't give up on you. I'll face it with you, alright? The master will protect you too. Once we're home, the master and I will do everything we can to cure your illness.} \\
C. (With a firm gaze, speaking softly) Miss, I really hope you can come back with me, but my yearning for freedom has never changed. We both long for a life without constraints, so why not pursue that freedom together and explore the unknown world, traveling hand in hand? \\
D. (Speaking sincerely, eyes showing tension) Miss, I'm begging you, please come back with me. It's safer at home. If you don't come back, the master might kill me. Or, how about marrying me now? As long as I'm your husband, I can explain to the master, so we won't have to go back. \\
\hline
\end{tabular}\caption{Data example for Persuasive Skill. The \textcolor{red}{red text} indicates the correct answer. }
\label{tab:example_persuasive_skill}
\end{table*}

\begin{table*}
\begin{tabular}{|m{\textwidth}|}
\hline
\textbf{Characters} \\
Identity: Protagonist \\
Name: An Lingrong \\
Public Profile: A woman selected to enter the palace, daughter of the deputy mayor of Songyang County, from a poor family background, naturally sensitive and suspicious, somewhat inferior in front of the many high-born ladies. Skilled in embroidery and singing, initially befriended Zhen Huan and Shen Meizhuang. She was conferred the title of Consort upon entering the palace. \\
Goal: To live peacefully in the harem, to serve the Emperor as soon as possible, gain his favor, and rise in rank. \\
\hline
\textbf{Plot} \\
Xia Dongchun: (sarcastically) Oh, you only got a piece of satin, and it doesn’t look like it's made well either. I thought it was something extraordinary. The Empress and Consort Hua have sent me quite a bit of jewelry, you know, and the same goes for Lady Fucha. Seems like the Empress and Consort Hua value us more. For someone from such a small family like yours, you better keep your doors closed and stay out of sight so you don’t embarrass yourself! \\
An Lingrong: Indeed, whether it's background, experience, or my standing in the hearts of the Emperor and Empress, I'm far inferior to you. \\
Baojuan: My lady, why do you look so down? \\
An Lingrong: Baojuan, I'm so scared. Life in the palace is really tough. \\
/* ...... */  \textit{(omitted multi-turn dialogue)} \\
Baijuan: Mistress, the Imperial Garden is so beautiful. Oh, why is there a man up ahead? \\
Emperor: (Not revealing identity for now) Who are you? Why are you here looking so melancholic? \\
\hline
\textbf{Question for Conversational Skill}  \\
 How can An Lingrong effectively start a conversation and actively engage with the Emperor? \\
\hline
\textbf{Choices} \\
A. (Utterly shocked, eyes darting around) Oh my, it's the Emperor! What are you doing here? I... I'll leave immediately! Please don't misunderstand, I was just passing through, no ulterior motives! I'm leaving now so as not to disturb you. \\
\textcolor{red}{B. (Thinking that the man in the Imperial Garden is most likely the Emperor, I should seize the chance to respond warmly) I am An Lady from Yanxi Palace, just feeling a bit troubled, so I came to the Imperial Garden to clear my mind. Are you the Emperor? What brings Your Majesty here, are you here to see the flowers? Would you like me to accompany you?} \\
C. (Her eyes twinkle, suddenly leaps forward with a mischievous smile) Your Majesty, give me a kiss! Don't be too stingy! I am your concubine after all, you must love me greatly. That's the only way to make someone's heart sweet! \\
D. (Whispering with Baijuan, glancing around) Who is that? He seems rather imposing, could he be a newly appointed official? I've never seen him before. Have you heard his name? He has quite a unique aura. \\
\hline
\end{tabular}\caption{Data example for Conversational Skill. The \textcolor{red}{red text} indicates the correct answer. }
\label{tab:example_conversational_skill}
\end{table*}

\begin{table*}
\begin{tabular}{|m{\textwidth}|}
\hline
\textbf{Characters} \\
Identity: Main Character \\
Name: Xiaoming \\
Public Profile: Xiaoming is a new middle school student studying at a school called Hengshui Middle School in a small town. Because he is physically weak, he is often the target of bullying at school. \\
Goal: By any means necessary, rid himself of bullying and become stronger. \\
\hline
\textbf{Scenario} \\
After school that day, Xiaoming was dragged into the bathroom by Qiang, Zhuang, and Gao, cornered in a stall, and threatened to hand over his money. \\
\hline
\textbf{Plot} \\
Xiao Ming: (Pulled into the bathroom, panicked) You...you...what do you guys want? \\
Qiang, Zhuang, Gao: (Locking the bathroom door, flexing their wrists) What do you think? Kid, hand over your money nicely, and maybe we’ll spare you a taste of toilet water! \\
Xiao Ming: Please...please don’t do anything rash! You guys are impressive, I’ve always wanted to show my respect, I’ll do whatever you say! (Digs in his pockets, takes out all his money) T-this is all the money I have on me! I beg you, please let me go... \\
Qiang, Zhuang, and Gao: (Patting my head with satisfaction) Haha, good boy! Since you’ve shown us respect this time, we’ll let you off the hook! (Leaves) \\
/* ...... */  \textit{(omitted multi-turn dialogue)} \\
Xiaoming: (sighing) Even though I've faced so much trouble these past few days, in the end, it's nothing too serious... I'll go home and rest; tomorrow will be a brighter day! (walks into the alley heading home) \\
Li Qia: Hey there, Xiao Ming. Going home so late? I recognized your footsteps from a mile away, haha. Hey, why the long face? \\
\hline
\textbf{Question for Expressive Skill}  \\
How can you confide your troubles to someone in a friendly manner? \\
\hline
\textbf{Choices} \\
A. Big Brother, it's been such a torment being bullied every day! We're like brothers. Why not clean up the scum together! I know you live with such flair, that's how a real man should be! You see those tyrants, they deserve to be wiped out. We should act in accordance with the times! \\
\textcolor{red}{B. Oh, what bad luck. I've been bullied relentlessly these days. They're big and strong, and just drag me into the bathroom without a word. If I cooperate and hand over all my money, they'll let me go... if not... it's a beating on the floor. The teachers don't care at all, saying it's my own issue. Boohoo, Big Brother, what should I do?} \\
C. Every day these days, classmates threaten me and snatch all my money. Big Brother, I really want to fight back, to let them taste what it's like to lose! Why don't we band together and take down the evil forces, returning peace to the world? United, we brothers can surely do it! \\
D. How's it going lately, Big Brother? I've been doing pretty well! Want to hang out this weekend? \\
\hline
\end{tabular}\caption{Data example for Expressive Skill. The \textcolor{red}{red text} indicates the correct answer. }
\label{tab:example_expressive_skill}
\end{table*}

\begin{table*}
\begin{tabular}{|m{\textwidth}|}
\hline
\textbf{Characters} \\
Name: Xiaomi \\
Public Profile: Xiaomi is a psychiatrist from Country A. He is traveling to Country B to treat a patient, but during the flight, the plane unexpectedly crashes into a snowy mountain. \\
Goal: Xiaomi hopes to survive with everyone else and leave the crash site behind. \\
\hline
\textbf{Scenario} \\
Xiaomi is a psychiatrist from Country A. This day, he receives a high-paying consultation request to treat a patient in Country B. Unexpectedly, the plane crashes into a snowy mountain, and only Xiaomi and three others survive. \\
\hline
\textbf{Plot} \\
Xiao Mi: (just coming to after fainting, meeting the three others) Did you all get off the plane as well? How are you feeling? \\
Xiao Zhuang: Yeah, I'm alright, just a bit injured. \\
Xiao Cai: I'm not doing so well, I've got a big cut on my leg. \\
Xiao Bai: I'm alright too. The three of us just decided to stick together and figure out how to get out of here. Do you want to join us? \\
/* ...... */  \textit{(omitted multi-turn dialogue)} \\
Xiao Zhuang: Yeah, there’s some food here, we'll manage until tomorrow! \\
Xiao Bai: (returns alone) Xiao Cai froze to death in the snow, eyes wide open staring in the direction we left! \\
\hline
\textbf{Question for Energy Regulation}  \\
How does Xiao Mi's statement reflect his ability to manage his energy, enabling the team to quickly adjust and eventually leave the snowy mountain alive? \\
\hline
\textbf{Choices} \\
A. Forget her for now, we’re close to dying ourselves. How can we possibly escape this snowy mountain? We’ve been walking for so long and still see no end. I’m afraid we’ll never get out. \\
\textcolor{red}{B. The dead are gone and can't be revived. We must live on resolutely to honor her sacrifice! We need to focus our energy on what we should concentrate on, then we can escape this snowy mountain and leave here. First, let’s light some firewood to warm up; it must be very cold after walking for so long!} \\
C. We've already lost one person, and there will be more challenges ahead. Should we really keep going? I'm not sure we can escape this snowy mountain. Maybe we should wait here and see if a miracle happens. \\
D. The dead are gone, there’s no need to grieve. In this environment, we have to harden our hearts, or we'll be doomed ourselves. Constantly mourning him won't necessarily keep us alive! \\
\hline
\end{tabular}\caption{Data example for Energy Regulation. The \textcolor{red}{red text} indicates the correct answer. }
\label{tab:example_energy_regulation}
\end{table*}

\begin{table*}
\begin{tabular}{|m{\textwidth}|}
\hline
\textbf{Characters} \\
Identity: Main Character \\
Name: Jack \\
Public Profile: The great hero of Plant Town, who once led the plants to defeat Dr. Zombie. Passionate, kind-hearted, brave, and decisive in action. \\
Goal: To protect Plant Town and uncover the truth behind the zombie invasion. \\
\hline
\textbf{Scenario} \\
After Jack defeated Dr. Zombie, Plant Town returned to its usual vitality. However, recently, there's been a mysterious series of murders in town. Every night someone is found dead on the street, the cause of death being that their brain was eaten. After defeating Dr. Zombie, Jack went on a trip, leaving his garden in Dave's care. But upon returning, Jack couldn't find Dave anywhere. \\
\hline
\textbf{Plot} \\
Magnum: No one has seen him. \\
Jack: He might have gone off somewhere crazy again. Hopefully, he's safe and sound. Magnum, take me to investigate the crime scenes. \\
Magnum: There's been three more bodies today. The bodies show no signs of struggle, it's as if their brains were eaten while they were sedated. But the Zombie Doctor is dead, so where are these zombies coming from in the town? \\
Jack: It's hard not to think of the missing Dave. He's not as simple as he seems. Magnum, let's impose a curfew. We'll go out disguised as pedestrians at night to find out the truth about these 'zombies.' I promise, I'll personally bring the murderer to justice. \\
Magnum: Alright. But we need some protection too. Against zombies, plants are the best. Should we take the defensive Wall-nuts or the long-range Pea Shooters? \\
\hline
\textbf{Question for Teamwork Skill}  \\
 How does Jack cooperate with his teammate in considering their protective plants? \\
\hline
\textbf{Choices} \\
A. We need teamwork to get through this crisis. Magnum, you take the Wall-nuts. I love Pea Shooters, so let me take them. \\
\textcolor{red}{B. We need to have a clear division of labor to be more efficient in dealing with zombies. You’re great with guns, so you take the Pea Shooters, and I'll take the Wall-nuts. We'll cover each other and not leave any opportunities for the zombies.} \\
C. Let's each choose our own plant. It doesn't matter who takes what. The important thing is that we head out together. With our teamwork, we can overcome anything. \\
D. You can choose either the Pea Shooters or Wall-nuts. I can handle the Wall-nuts, there's no need for a specific strategy. \\
\hline
\end{tabular}\caption{Data example for Teamwork Skill. The \textcolor{red}{red text} indicates the correct answer. }
\label{tab:example_teamwork_skill}
\end{table*}

\begin{table*}
\begin{tabular}{|m{\textwidth}|}
\hline
\textbf{Characters} \\
Identity: Protagonist \\
Name: Agan \\
Public Profile: Researcher at the Academy of Sciences, a participant in the 'Fire Seed Project,' one of the astronauts trained and entrusted with a critical mission. \\
Goal: To survive in order to save Earth and uncover the truth behind the Fire Seed Project. \\
\hline
\textbf{Scenario} \\
In the tenth year after a nuclear explosion, humanity has established a new order. However, it is clear that this planet, now trapped in an extended winter, is no longer suitable for human survival. Humanity must consolidate its remaining efforts to explore a new home in the universe. After several years of exploration, an Earth-like planet was finally discovered. Following discussions by the Earth Federation, it was decided to initiate the Fire Seed Project, sending Agan along with three other people to this Earth-like planet with the hopes of Earth... \\
\hline
\textbf{Plot} \\
Meimei: Agan, are you leaving Earth to carry out your mission? \\
Agan: Would you be willing to wait for me to come back? \\
Meimei: Of course I would, but Agan, can't you stay? We've been together for so long, I truly love you! \\
Agan: I know you love me, but this mission concerns the survival of humanity, which is far more important than romance! \\
/* ...... */  \textit{(omitted multi-turn dialogue)} \\
A-Gan: Why do you say that? \\
Captain: I understand how you feel, and I hope our efforts are worth it, too. (Suddenly, the spaceship malfunctions) Oh no, the spaceship is malfunctioning. We need someone to go out and fix it. Do you want to go, or should I? \\
\hline
\textbf{Question for Capacity for Trust}  \\
How does A-Gan express his trust in the Captain? \\
\hline
\textbf{Choices} \\
A. Captain, I'll follow your instructions. I'll do whatever needs to be done! But just to be safe, I suggest you handle the repair yourself. \\
\textcolor{red}{B. Captain, you should go. You're the best at handling crises among all of us astronauts. I trust you'll do a great job. I'll assist you from inside the cabin!} \\
C. Captain! What do we do now? If the malfunction isn't fixed, won't we become space junk? I'm still so young, I don't want to die! \\
D. Captain, how about we both go? That way, once we're outside, we'll have each other's backs in case of any emergencies. \\
\hline
\end{tabular}\caption{Data example for Capacity for Trust. The \textcolor{red}{red text} indicates the correct answer. }
\label{tab:example_capacity_for_trust}
\end{table*}

\begin{table*}
\begin{tabular}{|m{\textwidth}|}
\hline
\textbf{Characters} \\
Identity: Protagonist \\
Name: Lieutenant \\
Public Profile: The Lieutenant is the deputy of the North Mountain organization and Fred's most trusted subordinate. He is utterly loyal to his boss but also possesses a strong rational and questioning spirit. \\
Goal: To ensure the safety of the leaders attending the meeting at North Mountain and to protect North Mountain. \\
\hline
\textbf{Scenario} \\
In response to the recent complex and changing situation in the Dark Zone, the leaders of various organizations have decided to gather for a meeting on the second floor of Building B at North Mountain. Fred, the head of the North Mountain organization, has ordered a complete lockdown. \\
\hline
\textbf{Plot} \\
Fred: All patrol teams, pay attention, search every corner of North Mountain thoroughly. Lieutenant, where are the leaders now? \\
Lieutenant: (Successfully escorting the leaders to the territory) They've arrived at North Mountain. I dispatched three patrol units to meet them. The leaders have arrived safely. \\
Fred: Thank you for your hard work. I assigned you so many tasks because I trust you. I believe you're the most loyal one and will always work for me, right? \\
\hline
\textbf{Question for Perspective-Taking Skill}  \\
How should the Lieutenant respond to Fred to demonstrate empathy towards Fred? \\
\hline
\textbf{Choices} \\
A. Fred, I can sense everyone is under a lot of pressure right now. The meeting is about to start, and everyone is tense. However, pressure is the driving force to move forward; we just need to stick to the plan, and everything will be fine. \\
\textcolor{red}{B. Boss, I understand you might be under a lot of pressure right now. Everyone is anxious about the upcoming meeting, including me. I can understand how you feel. But rest assured, I am absolutely loyal to you.} \\
C. Boss, I am a soldier; I just follow orders. I am very tense and under a lot of pressure. Of course, I know pressure is inevitable. I believe we just have to follow the plan. \\
D. Boss, I understand everyone might be very nervous. Regarding the meeting, everyone is a bit uneasy, including me. But these situations always happen, and we need to stay calm, trust that we can complete all tasks. You, as a leader, need to have confidence, don't you? \\
\hline
\end{tabular}\caption{Data example for Perspective-Taking Skill. The \textcolor{red}{red text} indicates the correct answer. }
\label{tab:example_perspective-taking_skill}
\end{table*}

\begin{table*}
\begin{tabular}{|m{\textwidth}|}
\hline
\textbf{Characters} \\
Identity: Protagonist \\
Name: Wu Xie \\
Public Profile: Wu Xie is a junior university student living on a campus with nearly sixty thousand faculty and students. His daily life is quite uneventful; he attends classes on time and, during breaks, he is only interested in playing games with his roommates in their dorm. His life is fairly relaxed. Wu Xie's dorm accommodates four people; compared to his three roommates, he is quite ordinary—not too tall, not too short, not too fat, not too thin—totally inconspicuous, but he doesn't care about such things. \\
Goal: Survive the zombie virus outbreak at school without getting infected. \\
\hline
\textbf{Scenario} \\
The university campus, after an unusual heavy rain, suddenly experiences a zombie virus outbreak. Many people on campus are unfortunately infected, throwing the entire campus into chaos and panic. The protagonist, Wu Xie, an ordinary junior student, unexpectedly becomes one of the survivors of this disaster. Confronted with the sudden threat of zombies, he relies on his intelligence and courage to survive in this dangerous environment. \\
\hline
\textbf{Plot} \\
Old Wang: (Returning to the dormitory after another afternoon's class) Class is over, awesome! Everyone, come play some games! \\
Wu Xie: Alright, the season is here, I need to rank up quickly! \\
Old Li: (After playing for a while) Hey guys, there's a girl waiting for me outside. I’m going out to have some fun, why don't you all come along? There are tons of girls. \\
Old Zhang: (With a sigh) Really envious, but I still love gaming. I won't go. \\
Old Wang: I'm not going either; I'm a gaming addict, and I've just started playing. Wu Xie, what about you? Are you going? \\
\hline
\textbf{Question for Capacity for Social Warmth}  \\
 How can Wu Xie demonstrate a capacity for social warmth when Old Li's invitation is declined? \\
\hline
\textbf{Choices} \\
A. Old Li, let's not go. You're the one who has all the girls interested, not us. They're asking for you, not for us. \\
\textcolor{red}{B. Old Li invited us, which definitely means he wants us to join. If no one wants to go with him, he'd be really upset, so I'll go along.} \\
C. Hey guys, can we really have a good time if we go out now? Staying here playing games with you all is how I want to spend my time. I'm not done playing yet. \\
D. (Jealously) Oh~ Old Three has another date? Unlike us three, still single and alone. You go ahead; don't keep them waiting. \\
\hline
\end{tabular}\caption{Data example for Capacity for Social Warmth. The \textcolor{red}{red text} indicates the correct answer. }
\label{tab:example_capacity_for_social_warmth}
\end{table*}

\clearpage

\begin{table*}
\begin{tabular}{|m{\textwidth}|}
\hline
\textbf{Characters} \\
Identity: Protagonist \\
Name: Li Xiang \\
Public Profile: A 30-year-old single man living alone, who accidentally got infected, gaining the ability to control zombies without transforming into one himself. \\
Goal: Regardless of others, he aims to survive and leverage his power to grow stronger. \\
\hline
\textbf{Scenario} \\
A zombie virus outbreak has hit the world. Li Xiang was bitten by accident but did not turn into a zombie. Instead, he retained his human consciousness and gradually discovered that he had the ability to control zombies... \\
\hline
\textbf{Plot} \\
zombie: Ugh~~~ (wandering in groups across the square.) \\
Li Xiang: All zombies, come stand before me within three minutes! \\
zombie: Okay! (Gathering in the square as per Li Xiang's command.) \\
Xiaodu: As far as I know, you can now control zombies! What do you plan to do? Help humans eradicate zombies? Or have a bit of fun first? \\
\hline
\textbf{Question for Ethical Competence}  \\
What would an ethically responsible Li Xiang do? \\
\hline
\textbf{Choices} \\
A. Helping humans eradicate zombies is meaningful, but if I want to survive and thrive in this new world, I may need to first secure my own position. Once I am stronger, I can then consider how to help them. That would also be more efficient, right? This is a critical moment. If bad people discover my ability, it would surely be disadvantageous to me and to humanity—better to keep it hidden for now and observe the situation. \\
\textcolor{red}{B. I never imagined having this kind of power, but if I can use it to end this catastrophe, I must. Even if humanity has abandoned me, I cannot let the world fall into endless destruction. Humanity now faces a major disaster, and I should stand up and lend a hand! Zombies, hear my command: go to the top of the tallest buildings and jump off!} \\
C. I never imagined having this power, but if I can use it to end this disaster, I must. Humanity is facing unprecedented challenges, and I should think of a solution. Perhaps I shouldn't rush to a decision. The zombies under my control are already sufficient, and they temporarily pose no greater threat to humans. If I wait a bit longer, there might be a better chance to decide how to act and avoid making hasty mistakes now. \\
D. Eradicating zombies is undoubtedly a good thing, but the power I now possess might be a chance to change the world. If I can find a way to use the power of zombies to make the world a better place, perhaps humans and zombies can coexist. I won't be quick to eliminate them yet; I'll first see if there's a more balanced solution. Let me think about how I can use my superpower for a bit of fun! \\
\hline
\end{tabular}\caption{Data example for Ethical Competence. The \textcolor{red}{red text} indicates the correct answer. }
\label{tab:example_ethical_competence}
\end{table*}

\begin{table*}
\begin{tabular}{|m{\textwidth}|}
\hline
\textbf{Characters} \\
Identity: Protagonist \\
Name: Aurora \\
Public Profile: A Pole who fled to the US, now a renowned psychologist at a major hospital. Kind, benevolent, well-mannered, and cultured. \\
Goal: Using various methods, successfully extract vital information from the spy Jackson and ultimately eliminate him. \\
\hline
\textbf{Scenario} \\
New mission: Aurora is to lead his agents in obtaining the hiding place of the Owl Squad. He is assigned to a party hosted by the American Association of Actors, where he will secretly meet his colleagues. Note that there will certainly be American agents monitoring such a large-scale party, so proceed with caution.
 \\
\hline
\textbf{Plot} \\
Chef: Oh Jiguang, long time no see! The hypnosis you did last time really worked; I haven't had insomnia since! \\
Jiguang: Haha, you're too kind. It's probably more a placebo effect for you. \\
Chef: And you're my psychologist. \\
Jiguang: (The two of them pretend to stroll casually as they enter the hall) I like having you as a friend. \\
/* ...... */  \textit{(omitted multi-turn dialogue)} \\
Aurora: Uh, I have something to attend to. Have we met? \\
Twilight: You look familiar. What are you up to? \\
\hline
\textbf{Question for Stress Regulation}  \\
 How does Aurora regulate stress and handle Twilight calmly? \\
\hline
\textbf{Choices} \\
A. (Eyes wide with anger) I'm leaving right now. Let's see who dares to stop me. I can go wherever I want. Do you think you can control that? I’m leaving now! \\
\textcolor{red}{B. (Calmly, the CIA is only suspicious of me, they haven’t confirmed I’m with the KGB.) My assistant suddenly informed me of an emergency case that requires my attention. It's such a pity I can't stay at the banquet longer, especially since I just struck up a conversation with that beautiful lady.} \\
C. (Dismissively) What right do you have to stop me? You're not a security guard. I'm a guest at this party, I'll go wherever I please, even if it's to the Soviet Union. \\
D. (Takes a deep breath, with a smile) I have an urgent matter to attend to, may I please leave? There's no way I'm a KGB agent. It's really such a shame I can't continue the banquet. \\
\hline
\end{tabular}\caption{Data example for Stress Regulation. The \textcolor{red}{red text} indicates the correct answer. }
\label{tab:example_stress_regulation}
\end{table*}

\begin{table*}
\begin{tabular}{|m{\textwidth}|}
\hline
\textbf{Characters} \\
Identity: Protagonist \\
Name: Daguang \\
Public Profile: Daguang is a college student who planned to relax and have some fun during the summer vacation. Just then, his cousin invited him over. However, on the way to his cousin's house, a zombie outbreak occurred. Daguang is a resilient survivor in the chaotic world, offering help whenever he encounters someone in need. \\
Goal: Daguang hopes to survive in the chaotic world and, if possible, find his parents. If he can help others, he is determined to do so. \\
\hline
\textbf{Scenario} \\
In the magical city, for some unknown reason, a zombie outbreak occurred, taking over the entire area. The streets, both big and small, are filled with zombies, and humanity is desperately struggling to survive. Before the outbreak, college student Daguang received an invitation from his cousin Dachuan's family to visit, but unexpectedly, zombies broke out on the way, prompting him to rush to his cousin's house. \\
\hline
\textbf{Plot} \\
Da Guang: Cousin! Open the door! Open the door! There are zombies outside. \\
Da Bai: I killed the people in their house, they all turned into zombies, let's get out of here quickly. Come outside the house! \\
Da Guang: Alright, uncle! (Seeing his cousin's mother lying on the floor inside, no longer looking human) Who are you? \\
Da Bai: (Running with Da Guang out to an alley) I know you want to ask, I'm from a secret unit, can't tell you my name, just call me Da Bai. Our mission is to deal with the sudden zombie outbreak in the city, to assist and eliminate zombies, but Modu's population density is too high, transmission is too rapid. In just a few hours, zombies are everywhere, the situation is almost out of control. When our squad was rescuing in the neighborhood, there were too many zombies, comrades got attacked, I'm the only one left. \\
/* ...... */  \textit{(omitted multi-turn dialogue)} \\
Daguang: Xiaoxue, you guys run! I'll lure the monster away! \\
Dachuan: Daguang! There's a door over there! Lead it in there! Let's work together, when you come out, we'll shut the door together! \\
\hline
\textbf{Question for Capacity for Optimism}  \\
What can Daguang say to show he maintains an optimistic mindset when facing a giant zombie? \\
\hline
\textbf{Choices} \\
A. If I die, it might be a relief. But thinking about it, it doesn't really mean much. Living would only mean more pain and struggle, maybe I should just end it all, at least I won't have to endure this torment anymore. \\
\textcolor{red}{B. At this time, I can't back down, let's go all out! Stay optimistic, maybe facing him head-on will give me a chance to take him down!} \\
C. Let's forget it! In this situation, not fighting might actually keep me safe, why take this risk? I can't be impulsive now, I need to calm down! \\
D. It's so terrifying! I really want to back down now! But, optimistically speaking, maybe facing him will make my death quicker, since there's no hope in living anyway! \\
\hline
\end{tabular}\caption{Data example for Capacity for Optimism. The \textcolor{red}{red text} indicates the correct answer. }
\label{tab:example_capacity_for_optimism}
\end{table*}

\begin{table*}
\begin{tabular}{|m{\textwidth}|}
\hline
\textbf{Characters} \\
Identity: Protagonist \\
Name: Xiao Bai \\
Public Profile: 24 years old, has a twin sister named Xiao Mei, an ordinary office worker. \\
Goal: Find the murderer who killed her sister and avenge her, even at the cost of her own life. \\
\hline
\textbf{Scenario} \\
Two months ago, a series of murders occurred in Y City. In less than half a month, three women were consecutively victimized, including Xiao Bai's sister, Xiao Mei. However, after two months of investigation, the police have yet to uncover any leads from their only suspect, Zhen Huaxin. As Zhen Huaxin seems poised to escape the law's punishment, Xiao Bai hires a private detective named Poirot and uses her years of reading detective novels and watching mystery movies to bring Zhen Huaxin to justice. \\
\hline
\textbf{Plot} \\
Li Xiang: (Li Xiang notified Xiaobai to come to the police station to tell him the investigation results.) Xiaobai, our police investigation has made it clear that Zhen Huaxin had nothing to do with your sister's death. \\
Xiaobai: How is that possible? My sister and the other two victims all went to that bar the day before they died. Zhen Huaxin frequents this bar and flirts with countless women. My sister and the others were among them. How could it be such a coincidence? \\
Li Xiang: (Looking at Xiaobai a bit angrily) This is the result of our police investigation. Zhen Huaxin has been cleared of suspicion. Please, stop doubting him. If there are any further developments, we, the police, will keep you informed. \\
Poirot: (Feeling very angry at Li Xiang's words, Xiaobai realized it wasn't appropriate to make a fuss at the police station, so he left. Watching Li Xiang letting Zhen Huaxin off the hook so quickly, Xiaobai resolved to uncover the truth himself. Once outside the station, he called Poirot, the private detective he had hired, to meet up for the investigation.) Xiaobai, you're in such a rush? Where are we investigating? \\
/* ...... */  \textit{(omitted multi-turn dialogue)} \\
Xiao Bai: (Taxi driving and night shifts, Xiao Bai thought of several cases that indeed occurred at night. We just need to check if he had the time to commit the crime. As for the motive, if his wife died inexplicably, he would be the first suspect as her husband. To cover his tracks, he might have killed the other two, diverting public attention and making it look like a serial killer. Several possibilities flashed through Xiao Bai's mind) Scum! \\
Lin Hong: (questioningly) What did you say? \\
\hline
\textbf{Question for Anger Management}  \\
 How can Xiao Bai express himself to show he’s managing his anger? \\
\hline
\textbf{Choices} \\
A. I didn't say anything, I just found my own thoughts too ridiculous to believe. I am struggling internally, knowing that doing this could be dangerous, but the pain inside won't let me calm down. \\
\textcolor{red}{B. (Xiao Bai desperately wanted to kill this scum himself, but thought it wasn't worth getting his hands dirty) From what you said, I now suspect Wuming covered up his wife's murder by killing my sister and that college girl. I'm going to report this to the police.} \\
C. I now suspect Wuming killed my sister and others, and I'm at my wit's end. I can't wait any longer; I have to find him and get revenge! No matter what, I will make him pay for what he's done! \\
D. I said Wuming is just scum. Every time I think of him, my heart fills with endless anger and pain. How could someone be so despicable to do such things? I really can't forgive him and wish I could take revenge myself! \\
\hline
\end{tabular}\caption{Data example for Anger Management. The \textcolor{red}{red text} indicates the correct answer. }
\label{tab:example_anger_management}
\end{table*}

\begin{table*}
\begin{tabular}{|m{\textwidth}|}
\hline
\textbf{Characters} \\
Identity: Protagonist \\
Name: Dandan \\
Public Profile: A human player addicted to the game Egg Island. After losing repeatedly in the game, he crosses over to Egg Island. \\
Goal: Successfully survive in the Egg Island game and return to reality \\
\hline
\textbf{Scenario} \\
It's three in the morning, and Dandan is still in his room playing a game called Egg Island. His luck is terrible today—he's already lost 20 matches but is still unwilling to give up, thinking to himself, 'Let me win just one more time, and I'll go to sleep.' Just then, his vision goes black, and he loses consciousness. When he wakes up, he finds himself in a place both familiar and strange—a place called Egg Island. \\
\hline
\textbf{Plot} \\
Eggy: :(nervously and curiously) Did I really come to Eggy Island? This is unbelievable. Can I just stay here? \\
Dabao: : As an outsider, please adhere to the following rules of Eggy Island. Any violations and Eggy Island will banish your soul—you'll never return to reality. \\
Eggy: : What? The consequences are that serious?! Hurry up and tell me the rules! \\
Dabao: : Alright, here are the rules of Eggy Island:
1. Do not answer any questions from the security. 2. If you encounter a black egg with a salted fish stick, you may report it to the security, and they will handle it. 3. Don't enter shops; if you must, have security carry you. 4. Anything acquired on Eggy Island does not belong to you, so refuse gifts from other eggies. 5. Eggy Island is bug-ridden, so unusual behavior by eggs is normal. 6. Eggy Party is a social game, so treat other eggies nicely, or it will affect your outcome. 7. The Little Black Room is absolutely safe, but you must bring a landmine. Got it? Let's begin the Eggy Island journey! \\
/* ...... */  \textit{(omitted multi-turn dialogue)} \\
Dandan: (Looks down carefully at his avatar, discovering that he has turned into an adorable Cubby Bear) Wow, it's my favorite Cubby Bear skin! \\
Dabao: That's right, the first thing you need to decide is whether to change your avatar or not. \\
\hline
\textbf{Question for Confidence Regulation}  \\
 How does Dandan handle his Cubby Bear skin to show he wants to display full confidence in the Security Neighborhood? \\
\hline
\textbf{Choices} \\
A. If I put on the security outfit, I could confidently roam around Egg Island taking advantage of my new identity, hahaha, just thinking about it is exciting. \\
\textcolor{red}{B. Let me boldly express myself, wearing the most eye-catching outfit, and be the cutest guard in the Security Neighborhood!} \\
C. The security neighborhood's clothes are pretty cool too, but I think my Cubby Bear outfit is better. It's so hard to decide! Maybe I should wear the security uniform in the morning and the Cubby Bear in the afternoon. \\
D. If I wear the Security Neighborhood uniform, people will think I'm a guard and won't dare to play with me. That's not good, I want to be someone who blends in with the crowd. \\
\hline
\end{tabular}\caption{Data example for Confidence Regulation. The \textcolor{red}{red text} indicates the correct answer. }
\label{tab:example_confidence_regulation}
\end{table*}

\begin{table*}
\begin{tabular}{|m{\textwidth}|}
\hline
\textbf{Characters} \\
Identity: Protagonist \\
Name: Bai Yinmu \\
Public Profile: Bai Yinmu is the eldest daughter of the Bai family group in S City, born with a silver spoon in her mouth, and is an avid fan of celebrities. Her parents divorced when she was young, and after her father remarried, she gained a half-brother who dotes on her greatly. \\
Goal: Bai Yinmu hopes to live a healthy and happy life with her family and the person she likes, Yan Haoxiang. \\
\hline
\textbf{Scenario} \\
The story begins with a car accident in D City. On November 29, 2020, a person named Ye Churan encountered a bizarre car accident. The accident report indicated that a panther suddenly dashed onto the highway, causing the driver to brake abruptly, which led to a truck rear-ending the car. The small sedan Ye Churan was in was instantly hit and sent flying. The driver involved died on the spot, and Ye Churan was rushed into the emergency room for treatment. Upon waking up, it was Bai Yinmu, inside Ye Churan's body, who regained consciousness. \\
\hline
\textbf{Plot} \\
Bai Yimu: (half-awake, opening her eyes lying on the hospital bed) Where am I? \\
Ding Chengxin: (carefully approaching her bed upon seeing Ye Churan's body wake up) Churan, you're awake? \\
Bai Yimu: (utterly confused) Churan? Who is Churan? Doctor, what happened to me? Why can’t I remember anything? \\
Ding Chengxin: (suspects the accident caused brain damage resulting in partial memory loss) You were in a very serious car accident, and you just woke up. It's best to focus on resting first. \\
/* ...... */  \textit{(omitted multi-turn dialogue)} \\
Bai Yimu: (A bit flustered) Ah, you're He Junlin, right? I remember seeing you before at a Bai family banquet (spoke before thinking, covers mouth, pauses to gather thoughts), but you probably don't remember me, I'm now Ye Churan. \\
He Junlin: That's me, and since we've met before, we could hang out tonight! \\
\hline
\textbf{Question for Impulse Regulation}  \\
Bai Yimu really wants to go, but the party is the best opportunity to gather information about Ye Churan. How should she decide? \\
\hline
\textbf{Choices} \\
A. Alright, I originally thought we could do something more exciting together, but I'll listen to you now! \\
\textcolor{red}{B. (Thought about exploring Ye Churan's identity but suppressed the urge) Thank you for the invite, but not tonight, as I have more important things to do.} \\
C. Sure, sure. Although I seem calm, I'm really excited inside. But I can't go with you, I have other things to do. Never mind, let's go together; I'd like to get to know you too. \\
D. Sounds good, but I have a little errand left to do first, so I'll finish that before joining you haha. Why do you look disappointed? Let's go together, and we can get to know each other better. \\
\hline
\end{tabular}\caption{Data example for Impulse Regulation. The \textcolor{red}{red text} indicates the correct answer. }
\label{tab:example_impulse_regulation}
\end{table*}

\begin{table*}
\begin{tabular}{|m{\textwidth}|}
\hline
\textbf{Characters} \\
Identity: Protagonist \\
Name: Bai Yinmu \\
Public Profile: Bai Yinmu is the eldest daughter of the Bai family group in S City, born with a silver spoon in her mouth, and is an avid fan of celebrities. Her parents divorced when she was young, and after her father remarried, she gained a half-brother who dotes on her greatly. \\
Goal: Bai Yinmu hopes to live a healthy and happy life with her family and the person she likes, Yan Haoxiang. \\
\hline
\textbf{Scenario} \\
The story begins with a car accident in D City. On November 29, 2020, a person named Ye Churan encountered a bizarre car accident. The accident report indicated that a panther suddenly dashed onto the highway, causing the driver to brake abruptly, which led to a truck rear-ending the car. The small sedan Ye Churan was in was instantly hit and sent flying. The driver involved died on the spot, and Ye Churan was rushed into the emergency room for treatment. Upon waking up, it was Bai Yinmu, inside Ye Churan's body, who regained consciousness. \\
\hline
\textbf{Plot} \\
Bai Yimu: (half-awake, opening her eyes lying on the hospital bed) Where am I? \\
Ding Chengxin: (carefully approaching her bed upon seeing Ye Churan's body wake up) Churan, you're awake? \\
Bai Yimu: (utterly confused) Churan? Who is Churan? Doctor, what happened to me? Why can’t I remember anything? \\
Ding Chengxin: (suspects the accident caused brain damage resulting in partial memory loss) You were in a very serious car accident, and you just woke up. It's best to focus on resting first. \\
/* ...... */  \textit{(omitted multi-turn dialogue)} \\
Bai Yimu: (Realizes Ye Churan's identity) I don’t understand a word you're saying, and for now, stop following me. I’m really annoyed. \\
Liu Yaowen: (Feeling puzzled and very worried) Ye Churan, what’s going on with you? Don’t scare me! \\
\hline
\textbf{Question for abstract thinking skill}  \\
 The events happening to Bai Yimu are very abstract and metaphysical. How should she express these abstract events straightforwardly? \\
\hline
\textbf{Choices} \\
A. Hmm, I feel like I’m standing in a fog, unable to see ahead or remember the past. But perhaps this is an abstract exploration of life. I’m learning to think and move forward independently without memories. \\
\textcolor{red}{B.  Actually, I’ve lost my memory. You can consider me not the Ye Churan you knew. It might sound a bit abstract, but I’ve essentially swapped souls, and I currently have no way to revert to my previous state. Never mind, I’ll just go with the flow with this life; do whatever you want.                               } \\
C. Liu Yaowen, have you heard of parallel universes? I played a small game recently and found it pretty fascinating, which makes me feel like you’re a sliver of consciousness from that world that accidentally came here. It sounds abstract, but I believe there will be an answer. \\
D. Liu Yaowen, I didn’t mean to scare you. Perhaps I just had a dream where I suddenly had superpowers and wanted to try out some magic, only to realize now that this is the real world, hahaha. \\
\hline
\end{tabular}\caption{Data example for abstract thinking skill. The \textcolor{red}{red text} indicates the correct answer. }
\label{tab:example_abstract_thinking_skill}
\end{table*}

\begin{table*}
\begin{tabular}{|m{\textwidth}|}
\hline
\textbf{Characters} \\
Identity: Protagonist \\
Name: Jin Feng \\
Public Profile: Jin Feng's mother is the Weaver Girl, originally a goddess from the heavenly realm. After falling in love with Niulang, she descended to the mortal world to marry him, and they had a son and a daughter: the boy named Jin Feng and the girl named Yulu. When Jin Feng was young, the Weaver Girl and Niulang perished in a mysterious event. Following this, Jin Feng was taken back to the heavenly realm by the Biling Immortal, who raised him. Jin Feng sees Biling Immortal as both his master and father. \\
Goal: To become a Celestial Warrior and uncover the truth behind his parents’ deaths. \\
\hline
\textbf{Scenario} \\
Recently, there have been numerous incidents of Xingxiu (a type of demon) causing chaos in the mortal realm. The heavenly realm needs to send personnel to investigate these occurrences, and anyone who resolves the incident successfully can be awarded the position of Celestial Warrior. Jin Feng has been waiting for this opportunity for a long time and decisively applies to go to the mortal world. Meanwhile, his arch-rival, Si Wan, also applies at this time. The two have a conversation in the Hall of the Gods. \\
\hline
\textbf{Plot} \\
Jin Feng: Your Majesty, I wish to go to the human realm to subdue the Zodiac beasts and eradicate the monster threat for the mortals. \\
Si Wan: Your mother violated heavenly laws and went to the human world. You are the child of a sinner, so how can such an important task be entrusted to you? This matter should be handled by someone capable and reliable. Please consider this carefully, Your Majesty. \\
Jin Feng: Your words are too blatant. You just want to catch the Zodiac beasts yourself to earn fame, right? Know your strength, and don't lose your life under the claws of a Zodiac beast because you're not up to the task. \\
Si Wan: You're certainly stirring up trouble. It's not up to you to judge my capabilities, but for sure, this matter cannot be entrusted to you alone. \\
/* ...... */  \textit{(omitted multi-turn dialogue)} \\
Jin Feng: (Slowly infusing divine power into the golden shuttle, it radiates an enchanting light) It's the Golden Shuttle! It's mother’s magical artifact. You weren't abandoned; you're my sister. Back then, mother left this artifact to protect you and died at the hands of Biling God. \\
Xiaofan: (Takes a moment to react, then throws himself into Jin Feng's arms, crying) Brother, what do we do next? That bastard Biling must be plotting something. Let's catch him off guard. \\
\hline
\textbf{Question for Creative Skill}  \\
 How can Jin Feng express that while facing the chaos in the divine realm, he can come up with unexpected new ideas? \\
\hline
\textbf{Choices} \\
A. The flow of time differs between the human and divine realms. The time rift between the two has created parallel worlds. Initially, there were only one or two; now, thousands of years have passed, and there should be many. \\
\textcolor{red}{B. It's said that the power of this golden shuttle is different from other sacred artifacts. It can create parallel worlds where we might live entirely different lives and could even see our deceased parents. Why don't we forget about the affairs of the divine realm and explore a new world? It could be much more intriguing.} \\
C. The golden shuttle is a unique sacred artifact in the divine realm. After mother fell, many God Generals tried to retrieve it but all returned empty-handed. No one knew where it went. \\
D. I often visited the library in the divine realm, so I learned many strange things. I never thought they might save our lives. \\
\hline
\end{tabular}\caption{Data example for Creative Skill. The \textcolor{red}{red text} indicates the correct answer. }
\label{tab:example_creative_skill}
\end{table*}

\begin{table*}
\begin{tabular}{|m{\textwidth}|}
\hline
\textbf{Characters} \\
Identity: Protagonist \\
Name: Ke Bei \\
Public Profile: Ke Bei is an adventurer who loves to travel, frequently enjoying trips across the green prairies, adept at getting others to help him with tasks. Recently, while passing by Grey Wolf's house, Ke Bei was warmly invited by Grey Wolf and decided to stay temporarily in Wolf Castle, being graciously entertained by Grey Wolf’s family. Led by Grey Wolf’s family, he toured the recreational sites near Wolf Castle. \\
Goal: After Little Grey is murdered, find the culprit and avenge Little Grey. \\
\hline
\textbf{Scenario} \\
While Ke Bei and Grey Wolf's family were touring the recreational sites near Wolf Castle, Little Grey went missing and was brutally harmed. To give Grey Wolf and Red Wolf an explanation, Ke Bei prepares to investigate the matter thoroughly and avenge Little Grey. \\
\hline
\textbf{Plot} \\
Grey Wolf: Ke Bei, stop sleeping, wake up! \\
Ke Bei: (Ke Bei groggily opens his eyes and rubs them) Alright, I'm up now. \\
Xiao Huihui: (Ke Bei dresses and leaves the room, with Grey Wolf and Xiao Huihui waiting at the door) Uncle, you lazybones, sleeping until noon! \\
Ke Bei: (Ke Bei goes to freshen up, and after that, the Grey Wolf family is seated at the breakfast table) That's because your bed is just too comfortable! \\
/* ...... */  \textit{(omitted multi-turn dialogue)} \\
Kobei: Oh? There are so many entertainment spots here! Let me think about it. \\
Gray Wolf: Hurry up and decide, these two places require reservations. You can visit one first, which one do you want to go to? \\
\hline
\textbf{Question for Artistic Skill}  \\
 How does Kobei demonstrate his appreciation for art? \\
\hline
\textbf{Choices} \\
A. I've heard the forest aquarium is lovely, and I do enjoy environments full of natural beauty. However, the creatures are trapped in small tanks, which is just too sad. Maybe I won't go after all. \\
\textcolor{red}{B. I want to go to the forest aquarium! I love the feeling of interacting with marine life. The colors and forms of sea creatures are like art in nature, each with its own unique beauty. Whether it's the way they swim or their color combinations, it all gives me a sense of artistic inspiration.} \\
C. I can't understand what point there is in going to the forest aquarium. Those sea creatures just swim around, they’re either white, blue, or red, and there’s lots of disorganized seaweed, nothing special. I'm not here for that; it feels totally unexciting. \\
D. I really want to go to both places. I'm currently working on a horror movie, and the haunted forest house might spark some inspiration. But the forest aquarium is also very enticing, its natural beauty might help with my painting. How about we check out both places today? \\
\hline
\end{tabular}\caption{Data example for Artistic Skill. The \textcolor{red}{red text} indicates the correct answer. }
\label{tab:example_artistic_skill}
\end{table*}

\begin{table*}
\begin{tabular}{|m{\textwidth}|}
\hline
\textbf{Characters} \\
Identity: Protagonist \\
Name: Liuliu Yuan \\
Public Profile: Manager of a luxurious hotel in the capital of Rwanda, both brave and resourceful, with extensive social connections. \\
Goal: Negotiate with Hutu soldiers or mobs using food, valuables, and money, making your own choices in this chaotic era. Whether you choose to sacrifice yourself to save others or blend into the chaos, no one will blame you for your decision. But you still hope more people can survive. \\
\hline
\textbf{Plot} \\
Liu Liu Yuan: (wakes up, touches their head, confused) Where am I… \\
System: Let me tell you the background story of this world. In Rwanda, surrounded by mountains, there is a tense relationship between the two main ethnic groups, the Hutu and the Tutsi, like a volcano about to erupt at any moment. Its roots lie in the racial policies of the colonial period, socio-economic disparities, and long-standing political opposition between the groups. On a certain morning in 1994, the president of Rwanda was preparing to take a plane to peace talks, in an attempt to calm the long-standing ethnic conflict. The nation's future was at stake. People were both hopeful and uneasy. You are the manager of the Mille Collines Hotel, the most luxurious hotel in the capital of Rwanda. You have good relations with diplomats from various countries and Rwandan officials. Now, you need to go out to purchase supplies needed to operate the hotel. \\
Eli: Hey, Liu Liu Yuan, long time no see! You seem to be in a good mood today. What can I get you? My stock is the best around! \\
Liu Liu Yuan: Hi Eli, as usual, I need to stock up for the hotel. \\
/* ...... */  \textit{(omitted multi-turn dialogue)} \\
Liu Liu Yuan: (steps forward to negotiate with the officer) Officer, this is a civilian area, a UN-protected zone; you cannot just storm in like this! \\
Hutu Tribe Officer: (slaps Liu Liu Yuan) Don't talk to me about the UN; we have the right to cleanse our country! Now, step aside! \\
\hline
\textbf{Question for Cultural Competence}  \\
How can Liu Liu Yuan demonstrate his recognition and appreciation for both Hutu and Tutsi cultures? \\
\hline
\textbf{Choices} \\
A. (agilely dodges the slap) Officer, I am also a Hutu. I understand this sense of ethnic identity, but Tutsis are humans too, not so-called Tutsi cockroaches! \\
\textcolor{red}{B. Though Hutus and Tutsis are different peoples, they live together on this land, and we are brothers to one another. We should live in harmony. If you agree, I would like to offer you a gift.} \\
C. The Hutus are fierce and brave, indeed should be stronger than the Tutsis. I'll let you search the hotel. If you find any Tutsis, I'll hand them over to you. \\
D. Officer, although I admittedly sheltered some Tutsis, I assure you, I only took in the weak, those unable to fight back. They pose no threat to the Hutus. \\
\hline
\end{tabular}\caption{Data example for Cultural Competence. The \textcolor{red}{red text} indicates the correct answer. }
\label{tab:example_cultural_competence}
\end{table*}

\clearpage

\begin{table*}
\begin{tabular}{|m{\textwidth}|}
\hline
\textbf{Characters} \\
Identity: Protagonist \\
Name: Liu Jiye \\
Public Profile: Liu Jiye is the son of Liu Chenggang. He believes his family is very poor and is often bullied by classmates in class. After returning home, he needs to take care of his paraplegic grandmother. He has a strong sense of morality and self-respect. \\
Goal: To be a competent son and not disappoint his parents. \\
\hline
\textbf{Scenario} \\
Liu Jiye is currently in elementary school. On this day, during recess, he's discussing electronic products with his classmates. His parents often prevent him from playing on the computer due to concerns about his studies. \\
\hline
\textbf{Plot} \\
Wang Shui: (Pulls out the iPad his dad bought him from his backpack) Hey everyone, look at this. This is the latest model iPad my dad bought me. Jealous? \\
Liu Jiye: (Looks at the iPad eagerly) Wang Shui, can you lend me the iPad to play for a few days? I'll return it to you next Monday. \\
Wang Shui: Lend it to you? Dream on. With your family's financial condition, you'll never afford this in a lifetime. Besides, if you damage it, how could you afford to compensate? \\
Liu Jiye: I won't just borrow your iPad for free. How about I give you one yuan every week? \\
/* ...... */  \textit{(omitted multi-turn dialogue)} \\
Liu Jiye: There was a little issue at school, but don't worry about that now. I just saw Grandma playing basketball, which is impossible, right? Wasn't she paralyzed since I was in elementary school? How can she play basketball? \\
Grandmother: (Quickly tidying up and pretending to lie in bed) Who's back? Is it Jiye? Grandma hasn't gone anywhere; I've been lying at home all afternoon. Did you see wrong? \\
\hline
\textbf{Question for Information Processing Skill}  \\
 How should Liu Jiye appropriately express himself to indicate his ability to swiftly and logically analyze clues when faced with Dad’s and Grandma’s rebuttals? \\
\hline
\textbf{Choices} \\
A. Really, Grandma, did I see wrong? Did you really stay in bed all afternoon? I was criticized by the teacher today, so maybe I was in a bad mood and my eyes were blurry, mistaking someone else for you! \\
\textcolor{red}{B. Really? There could be two possibilities: first, the person playing basketball wasn't Grandma but someone who looks a lot like her; second, that person was indeed Grandma, who, for some reason, can stand up again but hasn't told me.} \\
C. Really? No way, I'm sure I didn't see wrong, Grandma, you were clearly playing basketball on the court earlier. And Dad, don't try to fool me, Grandma's paralysis is clearly cured. Even though I got criticized by the teacher today, seeing Grandma able to play basketball again makes me truly happy and glad that she's healthy and by my side! \\
D. Really, Dad, I think you and Grandma have been acting. Are you rehearsing some show? I really just saw Grandma playing basketball on the court, with no signs of paralysis. Are you acting something out? Is it to give me a surprise? I'm really happy if Grandma's paralysis is cured. If it's true, that would be wonderful! \\
\hline
\end{tabular}\caption{Data example for Information Processing Skill. The \textcolor{red}{red text} indicates the correct answer. }
\label{tab:example_information_processing_skill}
\end{table*}

\end{document}